\documentclass[11pt]{article}
\usepackage[margin=1in]{geometry}

\usepackage{times}
\usepackage{xcolor,hyperref}
\providecommand{\notationchange}[1]{{\color{black}#1}}

\usepackage[utf8]{inputenc}
\usepackage[T1]{fontenc}
\usepackage{booktabs}
\usepackage{amsfonts}
\usepackage{nicefrac,bbm}
\usepackage{microtype}
\usepackage{wrapfig}
\usepackage{mathtools}
\usepackage{bm}
\usepackage{enumitem}
\usepackage{tikz}
\usetikzlibrary{positioning, arrows.meta, fit, backgrounds, calc, shapes.geometric}
\usepackage{pgfplots}
\pgfplotsset{compat=1.18}
\usepackage{array}
\usepackage{makecell}
\usepackage{rotating}
\usepackage{xspace}
\newcommand{\R}{\mathbb{R}}
\newcommand{\E}{\mathbb{E}}
\newcommand{\sigmoid}{\sigma}
\DeclareMathOperator{\ReLU}{ReLU}
\DeclareMathOperator{\SiLU}{SiLU}
\DeclareMathOperator{\GELU}{GELU}

\DeclareMathOperator{\RMSNorm}{RMSNorm}

\newcommand{\cdtname}{\texttt{FoundCause}\xspace}

\newcommand{\yes}{\textcolor{green!55!black}{\checkmark}}
\newcommand{\no}{\textcolor{red!70!black}{x}}
\newcommand{\partiall}{\textcolor{orange!80!black}{$\sim$}}

\definecolor{cInput}{HTML}{E8F1F8}
\definecolor{cEncoder}{HTML}{DDEEFF}
\definecolor{cStats}{HTML}{FDECD3}
\definecolor{cEdge}{HTML}{D9EAD3}
\definecolor{cTri}{HTML}{FFE5E5}
\definecolor{cConf}{HTML}{E6D9F2}
\definecolor{cOutput}{HTML}{F5F5F5}
\definecolor{cBorder}{HTML}{333333}
\definecolor{cArrow}{HTML}{555555}
\tikzset{
  block/.style={draw=cBorder, rounded corners=2pt, minimum width=2.3cm, minimum height=0.75cm,
                align=center, font=\small, line width=0.5pt, thick},
  subblock/.style={draw=cBorder, rounded corners=1.5pt, minimum width=1.9cm, minimum height=0.6cm,
                   align=center, font=\footnotesize, line width=0.4pt},
  tensor/.style={draw=cBorder, rectangle, rounded corners=1pt, minimum height=0.55cm,
                 align=center, font=\scriptsize\ttfamily, line width=0.4pt, fill=cInput},
  mod/.style={thick, -{Stealth[length=2mm]}, color=cArrow},
  lab/.style={font=\scriptsize, midway, above, sloped},
  grp/.style={draw=cBorder, thick, dashed, rounded corners=3pt, inner sep=6pt, line width=0.6pt},
}
\usepackage{fontawesome5}
\title{\texttt{FoundCause}: A Foundation Model for Causal Discovery with Latent Confounders from Observational Data}

\author{Patrick Bl{\"o}baum$^{1}$, Krishnakumar Balasubramanian$^{1,2}$, Shiva Prasad Kasiviswanathan$^{1}$\\
$^1$Amazon Web Services\\
$^2$Department of Statistics, University of California, Davis\\
}
\date{}
\usepackage[square,numbers,sort&compress]{natbib}

\begin{document}

\maketitle

\begin{abstract}

Causal discovery from observational data requires recovering directed relations and latent confounding without interventions. We propose \cdtname, an {\em amortized causal discovery foundation model} trained only on synthetic structural causal models that maps datasets to causal graphs in a single forward pass. A permutation-equivariant transformer alternates attention over samples and variables to model cross-variable dependence and per-variable distributions. Pairwise dependence and asymmetry statistics guide its attention, edge decoder and confounder module. The decoder separates edge existence from direction, while triangular refinement incorporates evidence from chains and colliders. Learnable latent tokens predict which variable pairs share a hidden common cause, and a masked input representation handles missing data. To our knowledge, \cdtname is the first amortized causal discovery approach for i.i.d.\ observational data to predict latent confounding as a separate, explicit output. On 14 real-world and semi-synthetic datasets, \cdtname leads 11 classical and 7 amortized baselines in average AUROC, $F_1$, skeleton $F_1$, and normalized structural Hamming distance ($\mathrm{SHD}/d$): clearly in $\mathrm{SHD}/d$ ($0.875$ vs.\ $0.956$), narrowly in $F_1$ ($0.531$ vs.\ $0.516$). It also has the highest accuracy among amortized methods on bivariate cause--effect pairs.

\begin{center}
\faGithub~\textbf{Model}: \href{https://github.com/amazon-science/foundcause}{https://github.com/amazon-science/foundcause}
\end{center}
\end{abstract}

% Compact the main-paper float gaps; restore the class spacing before the appendices.
%\newlength{\mainpapertextfloatsep}
%\setlength{\mainpapertextfloatsep}{\textfloatsep}
%\setlength{\textfloatsep}{8pt plus 2pt minus 2pt}

\section{Introduction}

Causal discovery aims to recover directed relationships and latent confounding from observational data, without controlled interventions. Given i.i.d.\ samples, the goal is to infer the causal graph of the data-generating process, typically formalized as a structural causal model (SCM).

Constraint-based methods (e.g.,~\cite{spirtes2000causation,zhang2008causality}) and score-based methods (e.g.,~\cite{chickering2002optimal}) rely on conditional independence tests or combinatorial search, while continuous optimization approaches (e.g.,~\cite{zheng2018dags,ng2020role}) formulate discovery as a differentiable problem. Other methods use functional assumptions on data-generating mechanisms to achieve identifiability (e.g.,~\cite{shimizu2011directlingam,anm,Zhang_UAI}). These methods require a separate search or optimization for each dataset and can be sensitive to hyperparameters and distributional assumptions. More fundamentally, observational causal discovery is not identifiable without additional assumptions, which real-world data may violate. This motivates inference procedures that combine signals across several identifiability regimes.

Recent work treats causal discovery as \emph{amortized inference} (e.g.,~\cite{lorch2022avici,ke2023learning,montagna2024demystifying,wu2024recipe, Mahajan2024amortized}): a model trained on synthetic datasets learns to map data directly to graph structure. This allows single-pass inference instead of per-dataset optimization. Most neural approaches, however, make limited use of known causal signals, such as cause--effect asymmetries and structural patterns like chains and colliders. SEA and CauScale, which add classical-algorithm estimates or global statistics as inputs, are partial exceptions. Incorporating these signals may help models generalize beyond their training distribution.

We propose \cdtname, which combines amortized inference with explicit statistical inductive biases. \notationchange{We focus on practical inference across diverse data-generating processes.} A permutation-equivariant transformer alternates attention over samples and variables. Here, pairwise causal statistics guide attention and enter the edge decoder and confounder module directly. The decoder separates edge existence from direction, and triangular refinement uses evidence from higher-order causal motifs. A dedicated module with learnable latent tokens predicts a symmetric matrix of hidden-common-cause probabilities. The model thus predicts both directed relations and latent confounding in a single forward pass.

\subsection{Related Work}
Constraint-based and hybrid approaches (e.g., PC, FCI, RFCI, GFCI) use conditional independence tests and orientation rules to estimate Markov equivalence classes, where FCI variants handle latent confounding~\citep{spirtes2000causation,colombo2014order,ogarrio2016gfci}. Score-based methods such as GES search over equivalence classes using decomposable scores~\citep{chickering2002optimal,nazaret2025xges}. Continuous optimization approaches (e.g., NOTEARS, DAG-GNN, GraN-DAG, GOLEM, DAGMA) replace combinatorial search with smooth acyclicity constraints or differentiable DAG penalties~\citep{zheng2018dags,yu2019daggnn,lachapelle2020gradient,ng2020role,bello2022dagma}.

Intervention-aware methods (e.g., DCDI, ENCO) use interventional data to improve identifiability, where ENCO separates edge existence and orientation~\citep{brouillard2020dcdi,lippe2022enco}. Bayesian and generative approaches (e.g., DECI, DiBS, BayesDAG) learn distributions over graphs and mechanisms, but require dataset-specific posterior inference~\citep{geffner2022deci,lorch2021dibs,annadani2023bayesdag}. Bivariate methods (e.g., ANM, IGCI, PNL) use asymmetries such as residual independence or non-Gaussianity to infer direction from pairs~\citep{anm,daniusis2010igci,Zhang_UAI,mooij2016tuebingen}. Latent-variable methods represent hidden confounding with mixed graphs (e.g., PAGs, MAGs, ADMGs)~\citep{bhattacharya2021differentiable,ashman2023neural,ma2024scalable}. We incorporate asymmetry and confounding signals into an amortized architecture.

\textbf{Identifiability Assumptions.}
Under Markov and faithfulness, the DAG is identifiable only up to its Markov equivalence class~\citep{spirtes2000causation}. Full-DAG identifiability requires additional structure, including linear non-Gaussian noise (LiNGAM)~\citep{shimizu2006linear}, nonlinear additive noise in the bivariate case (ANMs)~\citep{anm}, post-nonlinear models with invertible transformations~\citep{Zhang_UAI}, nonlinear additive models yielding a topological order (CAM)~\citep{buhlmann2014cam}, equal or known error variances in linear Gaussian systems~\citep{peters2014identifiability}, or score-Jacobian asymmetries in additive-Gaussian DAGs~\citep{rolland2022score}. With latent confounding or selection bias, identification weakens to PAGs representing equivalence classes of MAGs~\citep{zhang2008causality,colombo2014order,ogarrio2016gfci}, while counterfactuals are identified under, e.g., monotonicity~\citep{Chao2023ModelingCM} or bijectivity~\citep{nasr2023counterfactual}. These asymptotic guarantees rely on hard-to-verify assumptions: faithfulness is often nearly violated~\citep{uhler2013geometry}, and real mechanisms may violate linearity, additive noise, or equal variances~\citep{glymour2019review}. Amortized discovery has the same limits; its training data acts as a prior on the test model~\citep{montagna2024demystifying}. Therefore, \cdtname learns empirical direction signals across synthetic SCMs spanning many of these settings.

\textbf{Amortized Causal Discovery.}
Amortized causal discovery trains a model once on synthetic SCMs and performs inference in a single forward pass. 
AVICI uses alternating attention over variables and samples with a simple edge decoder~\citep{lorch2022avici}, while CSIvA uses an encoder--decoder with variable-identity embeddings and autoregressive graph generation~\citep{ke2023learning}. Hybrid methods such as SEA aggregate graphs from classical algorithms on variable subsets via a neural aggregator~\citep{wu2024recipe}. Recent models include TabCausal and CDFM,\footnote{CDFM appeared after our initial draft.} pretrained on diverse SCM families~\citep{li2026tabcausal,qiao2026cdfm}; Arrow, acyclic by construction via a skeleton and a node order~\citep{thompson2026arrow}; and CauScale, which scales to large graphs with an inverse-covariance prior~\citep{peng2026causcale}. CDFM also marks latent confounding as bidirected edges and accepts missing values. Other approaches score structures by likelihood with prior-fitted networks~\citep{cdpfn2025} or use weak prior knowledge as input~\citep{kipt2026}. 

DAG-FM (Table~\ref{tab:fm-comparison}) predicts leaf nodes and their parents in turn, routing through a mixture of mechanism-specific experts~\citep{chen2026dagfm}. Other pretrained models address related problems. BCNP meta-learns a posterior over causal graphs from which structures can be sampled~\citep{dhir2025a}; ADAG pretrains an attention model across many linear SEMs~\citep{yin2025adag}; DCD-PFN uses a prior-fitted network to identify Markov boundaries and assembles the graph from local discoveries~\citep{guan2026dcdpfn}; targeted cause discovery infers the causes of a single target variable with a subsampled ensemble~\citep{kim2025largescale}; and TabPFN-CFM predicts both causal structure and outcomes~\citep{zhu2026tabpfncfm}. Learning causal direction by supervised classification predates these models: RCC classifies kernel mean embeddings of labelled cause--effect samples~\citep{lopezpaz2015towards}. CausalArena evaluates classical and pretrained methods under a common protocol on synthetic, semantic, and formula-grounded SCMs and on public real datasets, and identifies overlap between pretraining and test generators as a central difficulty in evaluating pretrained models~\citep{li2026causalarena}.

Amortization has also been applied beyond structure discovery, including ATE/CATE and intervention-effect estimation~\citep{nilforoshan2023zero,robertson2025pfn,balazadeh2026causalpfn,reuter2026use}. \cdtname follows this paradigm but differs in four ways: (i) it injects pairwise causal statistics into attention and decoding, (ii) it factorizes edge existence and direction, (iii) it refines pairwise representations via triangular graph-level message passing, and (iv) it predicts latent confounding as an explicit output, using dedicated tokens. 
Table~\ref{tab:fm-comparison} compares these methods, including (i), (iii), and (iv).

\begin{table}[t]
\centering
\small
\setlength{\tabcolsep}{3pt}
\renewcommand{\arraystretch}{1.15}
\resizebox{\textwidth}{!}{%
\begin{tabular}{@{}lccccccccc@{}}
\toprule
\textbf{Method}
 & \makecell{Latent\\confounders}
 & \makecell{Missing\\data}
 & \makecell{Variable-count\\agnostic}
 & \makecell{Nonlinear\\mechanisms}
 & \makecell{V-structure /\\structural reas.}
 & \makecell{Pairwise\\stat. features}
 & \makecell{Explicit\\DAG}
 & \makecell{Heterogeneous\\populations}
 & \makecell{Real-world\\benchmarks} \\
\midrule
AVICI \citep{lorch2022avici}
 & \no      & \no      & \yes      & \yes      & \no      & \no      & \no      & \no      & \partiall \\
CSIvA \citep{ke2023learning}
 & \no      & \no      & \partiall & \yes      & \no      & \no      & \no      & \no      & \partiall \\
\citet{montagna2024demystifying}
 & \no      & \no      & \yes      & \yes      & \no      & \no      & \no      & \no      & \no      \\
Cond\_FIP \citep{Mahajan2024amortized}
 & \no      & \no      & \yes      & \yes      & \no      & \no      & \yes     & \no      & \partiall \\
SEA \citep{wu2024recipe}
 & \no      & \no      & \yes      & \yes      & \partiall & \yes     & \no      & \no      & \partiall \\
Causal Pretraining (TS) \citep{stein2024causal} & \no      & \no      & \partiall  & \yes      & \no      & \no      & \no      & \no      & \partiall \\
%CD-PFN \citep{cdpfn2025}& \no      & \no      & \yes      & \yes      & \no      & \no      & \no      & \no      & \partiall \\
%Knowledge-Informed PT \citep{kipt2026} & \no      & \no      & \yes      & \yes      & \no      & \no      & \no      & \no      & \partiall \\
TabCausal \citep{li2026tabcausal}
 & \no      & \no      & \yes      & \yes      & \no      & \no      & \partiall & \no      & \yes      \\
Arrow \citep{thompson2026arrow}
 & \no      & \no      & \yes      & \yes      & \no      & \no      & \yes     & \no      & \partiall \\
CauScale \citep{peng2026causcale}
 & \no      & \no      & \yes      & \yes      & \partiall & \yes     & \partiall & \no      & \no      \\
CDFM \citep{qiao2026cdfm}
 & \yes     & \yes     & \yes      & \yes      & \no      & \no      & \partiall& \no      & \yes      \\
DAG-FM \citep{chen2026dagfm}
 & \no      & \no      & \yes      & \yes      & \no      & \no      & \yes     & \no      & \yes      \\
\addlinespace[3pt]
\textbf{\cdtname (ours)}
 & \textbf{\yes} & \textbf{\yes} & \textbf{\yes} & \textbf{\yes} & \textbf{\yes} & \textbf{\yes} & \textbf{\partiall} & \textbf{\partiall} & \textbf{\yes} \\
\bottomrule
\end{tabular}%
}
\vspace{-6pt}
\caption{Feature comparison with existing foundation-model and amortized causal discovery methods for observational data. 
\yes{} denotes full support, \no{} denotes lack of support, and $\sim$ indicates partial support or reliance on post-hoc extensions. 
Only \cdtname and CDFM predict latent confounding and handle missing data natively; CDFM encodes confounding as bidirected edges in its single adjacency output (a direct edge takes priority), whereas \cdtname predicts a separate confounding matrix and also uses pairwise statistical features and structural reasoning.}
\label{tab:fm-comparison}
\end{table}

\section{Problem Setup}
\label{sec:problem_setup}

Given $\bm{X}\in\mathbb{R}^{N\times D}$ with $N$ i.i.d.\ samples over $D$ observed variables, we aim to infer directed causal relations and latent confounding. We assume an acyclic SCM over the observed and any unobserved variables, $X_v = f_v(\bm{X}_{\mathrm{Pa}(v)},\epsilon_v)$, with parents $\mathrm{Pa}(v)$, possibly nonlinear and non-additive $f_v$, and jointly independent noise $\epsilon_v$.
We target a directed acyclic graph (DAG) $\bm{D}^*\in\{0,1\}^{D\times D}$, where $D^*_{ij}=1$ iff a directed path from $i$ to $j$ contains no other observed variable, and a symmetric matrix $\bm{C}^*\in\{0,1\}^{D\times D}$, where $C^*_{ij}=1$ iff an unobserved variable has such a path to both $i$ and $j$.

Following the AVICI framework~\citep{lorch2022avici}, we train on synthetic datasets with known structure, regenerated every epoch (Appendix~\ref{sec:training-data}). No per-dataset optimization is needed: a single forward pass gives edge and confounding probabilities, which we average over $K=10$ passes (Appendix~\ref{app:IP}). Section~\ref{sec:exp} tests whether the learned statistical patterns transfer to real-world data.

The model takes a data matrix $\bm{X}$ and an optional binary observation mask $\bm{M}\in\{0,1\}^{N\times D}$ as input, where $M_{ni}=1$ denotes an observed entry and $M_{ni}=0$ denotes a missing entry.
During training, we sample problems with $D\in[2,50]$ variables and $N\in[100,599]$ observations. At inference time, the same architecture can be applied to larger numbers of variables and samples.
The output consists of an edge-probability matrix $\hat{\bm{D}}\in[0,1]^{D\times D}$ and a symmetric confounding-probability matrix $\hat{\bm{C}}\in[0,1]^{D\times D}$.
For directed edges, $\hat D_{ij}=\sigma(\ell_{ij})$ is the predicted probability of the causal relation $i\to j$, where $\ell_{ij}$ is the corresponding edge logit.
For latent confounding, $\hat C_{ij}$ is the predicted probability that variables $i$ and $j$ share a hidden common cause.

\section{Model Architecture and Loss Function}
\label{sec:architecture}

\begin{figure}[!t]
\centering
\begin{tikzpicture}[
  mainbox/.style={block, minimum width=8cm, inner sep=2.5pt},
  sidebox/.style={block, inner ysep=2.5pt, inner xsep=4pt},
  io/.style={tensor, font=\footnotesize\ttfamily, minimum height=0.6cm},
  outbox/.style={tensor, fill=cOutput, font=\small, minimum width=1.1cm, minimum height=0.6cm},
  aux/.style={mod, dashed},
  link/.style={thick, color=cArrow},
  auxlink/.style={thick, dashed, color=cArrow},
  elab/.style={font=\scriptsize, inner sep=1.5pt},
  jn/.style={circle, fill=cArrow, inner sep=0pt, minimum size=3.2pt},
]

% ---------------- inputs ----------------
\node[io, minimum width=3.3cm] (X) at (4.1,0) {Data $\bm{X}\in\mathbb{R}^{N\times D}$};
\node[io, minimum width=3.3cm] (M) at (7.9,0) {Mask $\bm{M}\in\{0,1\}^{N\times D}$};
\coordinate (ib) at (8.9,-0.72);   % input-bus split: right of both joins

\node[sidebox, fill=cStats, minimum width=4.2cm, anchor=west] (stats) at (10.5,-0.72)
  {\textbf{Pairwise Statistics}\\
   {\footnotesize 45 raw-data features}\\
   {\footnotesize sym/asym/V-structure}\\
   {\footnotesize $+$ reliability metadata}};

% ---------------- main column ----------------
\node[mainbox, fill=cInput, anchor=north] (norm) at (6,-1.1)
  {\textbf{Per-variable Normalization} (fp32)\\
   {\footnotesize 2-channel input $[\bar{x}_{ni}m_{ni},\;m_{ni}]$}};

\node[mainbox, fill=cEncoder, below=0.4cm of norm] (enc)
  {\textbf{Axis-Factorized Encoder}\\
   {\footnotesize alternating variable- and sample-attention}\\
   {\footnotesize 16 blocks, $d_h{=}768$, 8 heads, $K_g{=}12$ global tokens}\\
   {\footnotesize stat-conditioned bias in variable-attention blocks}\\
   {\footnotesize learned fusion of intermediate and final block outputs}};   

\node[mainbox, fill=cEncoder, below=0.4cm of enc] (pma)
  {\textbf{PMA Pooling} over samples\\
   {\footnotesize $K_q{=}8$ query tokens, 4 heads, max-pool residual}};

\node[io, fill=cInput, minimum width=8cm, font=\footnotesize, below=0.42cm of pma] (Z)
  {$\bm{Z} \in \mathbb{R}^{D \times d_h}$};

\node[mainbox, fill=cConf, below=0.75cm of Z] (conf)
  {\textbf{Confounder Module}\\
   {\footnotesize $K_c{=}8$ tokens, 2-layer cross-attention}\\
   {\footnotesize noisy-OR loadings $\to \hat{\bm{C}}$}};

\node[mainbox, fill=cEdge, below=0.6cm of conf] (edge)
  {\textbf{Factored Edge Decoder}\\
   {\footnotesize existence head: symmetric features $+$ detached $\hat C_{ij}$}\\
   {\footnotesize direction head: gated antisymmetric features}};

\node[mainbox, fill=cOutput, minimum height=1.4cm, below=0.5cm of edge] (blend)
  {\textbf{Skel/Dir Blend} $+$ parent--child role score\\
   {\footnotesize learned $\sigma(\alpha_e),\sigma(\alpha_d)$; clamp $\pm 15$}\\
   {\footnotesize optional DAG post-processing}};   

% ---------------- right column ----------------
\node[sidebox, fill=cStats, inner xsep=6pt, minimum width=4.2cm, anchor=west] (gate) at ([xshift=0.5cm]Z.east)
  {\textbf{Feature Gate}\\
   {\footnotesize $\mathrm{MLP}(d_h{+}45 \to 256 \to 20)$}\\
   {\footnotesize per-pair sigmoid gate}};

\path (blend.center) ++(4.5,0) node[sidebox, fill=cTri, minimum width=4.2cm, minimum height=1.4cm, anchor=west] (tri)
  {\textbf{Triangular Refinement}\\
   {\footnotesize 3 rounds: outgoing, fork,}\\
   {\footnotesize collider, row-wise attention}};

% ---------------- outputs (left gutter) ----------------
\path (conf.center) ++(-5,0) node[outbox] (Chat) {$\hat{\bm{C}}$};
\node[outbox, below=0.4cm of blend] (Dhat) {$\hat{\bm{D}}$};   
% ---------------- main data path (solid) ----------------
\draw[link] (X.south) -- (X.south |- ib);
\draw[link] (M.south) -- (M.south |- ib);
\draw[mod] (X.south |- ib) -- (stats.west);
\node[jn] at (M.south |- ib) {};
\node[jn] at (ib) {};
\draw[mod] (ib) -- (ib |- norm.north);

\draw[mod] (norm) -- (enc);
\draw[mod] (enc) -- (pma);
\draw[mod] (pma) -- (Z);
\draw[mod] (Z.east) -- (gate.west);
\draw[mod] (Z.south) -- (conf.north);
\draw[mod] (Z.west) -- ++(-1.9,0) |- (edge.west);
\draw[mod] (conf.south) -- node[right, elab] {detached $\hat C_{ij}$} (edge.north);
\draw[mod] (conf.west) -- (Chat.east);
\draw[mod] (edge.south) -- (blend.north);
\draw[mod] ([yshift=-0.42cm]edge.east) -| (tri.north);
\draw[mod] (tri.west) -- (blend.east);
\draw[mod] (blend.south) -- (Dhat.north);
% role score r_ij = (W_V z_i)^T (W_U z_j) is computed from Z and added to the blended logits (cdt_v9.py 3917-3924)
\coordinate (zl) at ([xshift=-1.9cm]Z.west |- edge.west);   % Z lane at the decoder height
\draw[mod] (zl) |- node[pos=0.75, above, elab] {role score} (blend.west);
\node[jn] at (zl) {};

% ---------------- pairwise-statistics bus (dashed) ----------------
\coordinate (bx) at (15.1,0);
\coordinate (gm) at ([yshift=-0.25cm]gate.south);    % gate output -> bus merge height
\draw[auxlink] (stats.east) -| (gm -| bx);
\draw[aux] (gm -| bx) -- ([yshift=2.2pt]conf.east -| bx);   % direction tip: bus flows down after the merge
\draw[auxlink] (conf.east -| bx) -- (edge.east -| bx);
\draw[aux] (enc.east -| bx) -- node[pos=0.7, above, elab] {attention bias} (enc.east);
\draw[aux] (gate.east -| bx) -- (gate.east);
\draw[aux] (gate.south) |- node[pos=0.75, below, elab] {gated asym} ([xshift=-2.2pt]gm -| bx);   % tip stops short of the junction dot
\draw[aux] (conf.east -| bx) -- node[pos=0.7, above, elab] {sym $+$ gated asym} (conf.east);
\draw[aux] (edge.east -| bx) -- node[pos=0.7, above, elab] {sym $+$ gated asym} (edge.east);
\foreach \p in {enc.east, gate.east, gm, conf.east} \node[jn] at (\p -| bx) {};

% ---------------- legend (empty right-column slot beside PMA) ----------------
\coordinate (lg) at ([xshift=0.9cm, yshift=0.12cm]pma.north east);
\draw[mod] (lg) -- ++(0.8,0) node[right, font=\footnotesize, text=black] {main data path};
\draw[aux] ([yshift=-0.45cm]lg) -- ++(0.8,0) node[right, font=\footnotesize, text=black] {auxiliary input};

\end{tikzpicture}
\vspace{-6pt}
\caption{\textbf{Top-level Architecture.}
From data $\bm{X}$ and mask $\bm{M}$, the model computes normalized inputs and raw-data pairwise statistics. An axis-factorized encoder and attention-based pooling produce variable embeddings for the confounder module and factored edge decoder with triangular refinement.
Dashed lines carry pairwise statistics and gated features.}
\label{fig:cdt-arch}
\end{figure}

Figure~\ref{fig:cdt-arch} shows the four components: (i) an axis-factorized encoder, (ii) attention-based pooling, (iii) a factored edge decoder with triangular refinement and (iv) a learnable-token confounder module. The model has approximately $139$M parameters (Appendix~\ref{app:architecture_details}).

We represent each entry $(n,i)$ as  $\bm{x}_{ni}=[\bar{x}_{ni}m_{ni},\,m_{ni}]\in\mathbb{R}^2,$ where $\bar{x}_{ni}$ is the value standardized per variable over its observed entries and $m_{ni}\in\{0,1\}$ indicates whether the entry is observed. A linear projection gives $\bm{h}^{(0)}_{ni}\in\mathbb{R}^{d_h}$ via $\bm{h}^{(0)}_{ni}=\bm{W}_{\mathrm{in}}\bm{x}_{ni}+\bm{b}_{\mathrm{in}}$, $\bm{W}_{\mathrm{in}}\in\mathbb{R}^{d_h\times 2}$. Embeddings of missing entries ($m_{ni}=0$) are set to zero after the projection and again after every encoder block, and sample-wise attention and pooling exclude missing samples. No positional embeddings are used, making the model permutation-equivariant over variables (reordering the input columns reorders the rows and columns of $\hat{\bm{D}}$ and $\hat{\bm{C}}$), which is essential since causal structure should not depend on variable ordering.

\noindent\textbf{\notationchange{(i) Axis-Factorized Encoder.}}
Let $\bm{H}\in\mathbb{R}^{N\times D\times d_h}$ denote the stacked embeddings. 
We apply $2L=16$ attention blocks ($L=8$ layers; Table~\ref{tab:hyperparams}) alternating between variable-wise attention (over $D$) and sample-wise attention (over $N$) (see Figure~\ref{fig:cdt-encoder}, Appendix~\ref{app:architecture_details}). 
This factorization reflects two types of signal that causal discovery relies on: (conditional) dependencies across variables and the distribution of each variable across samples (e.g., skewness or non-Gaussianity).

Variable-attention incorporates:
(a) $K_g\in\mathbb{N}$ learnable global tokens appended along the variable dimension, and  
(b) a stat-conditioned bias  $b_{ij}^{(h)} = \left[\bm{W}_2\,\GELU\!\left(\bm{W}_1\bm{s}_{ij}\right)\right]_h$ (separate weights in each variable-attention layer), where $\bm{s}_{ij}\in\mathbb{R}^{d_s}$ are pairwise statistics and $\GELU$ is the Gaussian Error Linear Unit~\citep{hendrycks2016gaussian}. 
This bias adds classical dependence signals (e.g., correlation and asymmetry), guiding the model towards known causal cues.

Each block uses RMSNorm~\citep{zhang2019rmsnorm}, multi-head scaled dot-product attention (SDPA)~\citep{vaswani2017attention} and a SwiGLU~\citep{shazeer2020glu} feedforward network. 
The encoder outputs contextualized representations $\bm{H}^{\mathrm{enc}}\in\mathbb{R}^{N\times D\times d_h}$.

\noindent\textbf{\notationchange{(ii) PMA Pooling Over Samples.}}
For each variable $i\in\{1,\ldots,D\}$, we aggregate $\{\bm{H}^{\mathrm{enc}}_{ni}\}_{n=1}^N$ using Pooling by Multihead Attention (PMA)~\citep{lee2019set}, producing $\bm{z}_i^{\mathrm{PMA}}\in\mathbb{R}^{d_h}$. 
We also compute $\bm{z}_i^{\max}=\max_{n:\,m_{ni}=1} \bm{H}^{\mathrm{enc}}_{ni}.$ We combine these using a learned gate $g\in\mathbb{R}$: $
\bm{z}_i=\sigma(g)\bm{z}_i^{\mathrm{PMA}}+(1-\sigma(g))\bm{z}_i^{\max},$ where $\sigma(\cdot)$ denotes the sigmoid function.
Stacking over variables gives $\bm{Z}\in\mathbb{R}^{D\times d_h}$. 
This pooling is intended to capture higher-order distributional properties (e.g., multimodality and tail behavior). \notationchange{The pooling choice is fixed across the ablations.}

\noindent\textbf{(iii) Factored Edge Decoder.}
For each ordered pair $(i,j)$, we construct $\bm{f}^{\mathrm{exist}}_{ij}=[\bm{z}_i+\bm{z}_j;\bm{z}_i\odot\bm{z}_j;\mathrm{proj}_{\mathrm{sym}}(\bm{s}^{\mathrm{sym}}_{ij});\hat C_{ij}],$ and $\bm{f}^{\mathrm{dir}}_{ij}=[\bm{z}_i-\bm{z}_j;\mathrm{proj}_{\mathrm{asym}}(\bm{g}_{ij}\odot\bm{s}^{\mathrm{asym}}_{ij})],$ where $\bm{s}^{\mathrm{sym}}_{ij}$ and $\bm{s}^{\mathrm{asym}}_{ij}$ are the symmetric and antisymmetric statistics derived from $\bm{s}_{ij}$, $\bm{g}_{ij}$ is the encoder-conditioned feature gate (Appendix~\ref{app:architecture_details}) and $\hat C_{ij}$ is the output of the confounder module (component (iv) below), which is computed before the decoder and enters it detached from gradients. This separates edge existence (symmetric dependence) from direction (asymmetric signals), reflecting classical causal discovery principles.

Separate existence and direction heads map $\bm{f}^{\mathrm{exist}}_{ij}$ and $\bm{f}^{\mathrm{dir}}_{ij}$ to hidden vectors, which a fusion MLP combines into $\ell^{\mathrm{base}}_{ij}\in\mathbb{R}$. 
We also compute a role score $r_{ij}={(\bm{W}_V\bm{z}_i)^\top(\bm{W}_U\bm{z}_j)}/{\sqrt{d_{\mathrm{pair}}}}$, where $\bm{W}_V,\bm{W}_U\in\mathbb{R}^{d_{\mathrm{pair}}\times d_h}$. We add $r_{ij}$ with a small fixed (non-learned) weight $\omega$ to the edge logits after triangular refinement (below). This provides an explicit inductive bias toward consistent parent--child roles. See also Figure~\ref{fig:cdt-edge}, Appendix~\ref{app:architecture_details}.

We construct pair representations $\bm{P}\in\mathbb{R}^{D\times D\times d_{\mathrm{pair}}}$ and refine them for $R$ rounds using triangle-based updates inspired by AlphaFold2~\citep{jumper2021alphafold}. 
Each update aggregates over intermediate variables $k\in\{1,\ldots,D\}$, allowing reasoning over triples $(i,k,j)$. 
This captures causal motifs such as chains, forks and colliders, whose third-variable evidence helps decide which pairs are adjacent and, through colliders (V-structures), how edges are oriented. Refined logits $\bm{\ell}^{\mathrm{tri}}\in\mathbb{R}^{D\times D}$ are blended with $\bm{\ell}^{\mathrm{base}}$ separately for the skeleton and the direction component, with learned weights $\sigma(\alpha_e)$ and $\sigma(\alpha_d)$ and a learned scale (Figure~\ref{fig:cdt-arch} and Appendix~\ref{app:architecture_details}), giving $\tilde\ell_{ij}$. The final logit is $\ell_{ij}=\tilde\ell_{ij}+\omega r_{ij}$, and final edge probabilities are $\hat D_{ij}=\sigma(\ell_{ij}).$

\noindent\textbf{(iv) Confounder Module.}
Latent confounding is modeled using a learnable matrix $\bm{T}\in\mathbb{R}^{K_c\times d_h}$, where each row $\bm{t}_k\in\mathbb{R}^{d_h}$ represents a candidate hidden common cause and $K_c$ is the number of such confounders (see Figure~\ref{fig:cdt-conf}, Appendix~\ref{app:architecture_details}). The tokens $\bm{t}_k$ first cross-attend to $\bm{Z}$ in two layers, giving data-dependent tokens $\bm{t}'_k$. For each variable $i$ and confounder $k$, we compute $S_{ik}=\sigma(\mathrm{MLP}([\bm{z}_i;\bm{t}'_k]))$, $S_{ik}\in(0,1),$ which denotes the probability that confounder $k$ influences variable~$i$.  We aggregate the loadings using a noisy-OR, $C^{\mathrm{OR}}_{ij}=1-\prod_{k=1}^{K_c}(1-S_{ik}S_{jk})$, and add a scalar logit $\kappa_{ij}$ computed from the symmetric and gated antisymmetric pairwise statistics: $\hat C_{ij}=\sigma\big(\mathrm{logit}(C^{\mathrm{OR}}_{ij})+\kappa_{ij}\big)$, giving $\hat{\bm{C}}\in[0,1]^{D\times D}$. 
\notationchange{This gives the model an explicit output for hidden common causes alongside the directed edges. Section~\ref{sec:train-ablation} reports its effect on edge recovery.}

\begingroup
\setlength{\abovedisplayskip}{6pt plus 2pt minus 2pt}
\setlength{\belowdisplayskip}{6pt plus 2pt minus 2pt}
\setlength{\belowdisplayshortskip}{4pt plus 2pt minus 2pt}
\noindent\textbf{\notationchange{Training Objective.}}
Let $\bm{\ell}\in\mathbb{R}^{D\times D}$ be the directed edge logits, and $\bm{D}^*,\bm{C}^*\in\{0,1\}^{D\times D}$ the ground-truth adjacency and confounding matrices.
The valid-pair mask $V_{ij}=\mathbbm{1}[i\neq j]\,v_i v_j$ excludes diagonal entries and padded variables, with $v_i=1$ for present variables.
Let $|V|=\sum_{ij}V_{ij}$, $n_+=\sum_{ij}D^*_{ij}V_{ij}$, and $n_-=\sum_{ij}(1-D^*_{ij})V_{ij}$. The objective is
\begin{multline}\label{eq:mainloss}
\mathcal{L}=
\mathcal{L}_{\mathrm{dir}}
+\lambda_{\mathrm{asym}}\mathcal{L}_{\mathrm{asym}}
+\lambda_{\mathrm{cal}}\mathcal{L}_{\mathrm{cal}}
+\lambda_{\mathrm{bdir}}\mathcal{L}_{\mathrm{bdir}}
+\lambda_{\mathrm{skel}}\mathcal{L}_{\mathrm{skel}}
+\lambda_{\mathrm{conf}}\mathcal{L}_{\mathrm{conf}} \\
+\mathcal{L}_{\mathrm{fp}}
+\lambda_{\mathrm{den}}\mathcal{L}_{\mathrm{den}}
+\lambda_{\mathrm{pma}}\mathcal{L}_{\mathrm{pma}}
+\lambda_{\mathrm{gate}}\mathcal{L}_{\mathrm{gate}},
\end{multline}
which combines directed edge prediction, direction calibration, skeleton and confounding supervision, and regularization.

\textbf{(a) Directed Edge Loss.} The main term is a class-balanced binary cross-entropy over directed edges. 
We use direction-aware label smoothing
\[
\tilde{D}_{ij}=
\begin{cases}
1-\epsilon_d, & D^*_{ij}=1,\\
\epsilon_d, & D^*_{ji}=1,\\
\epsilon_0, & \text{otherwise},
\end{cases}
\qquad
\epsilon_d=0.05,\quad \epsilon_0=0.005,
\]
which introduces asymmetric supervision between $(i,j)$ and $(j,i)$, encouraging consistent edge orientation.

We define
\[
\mathcal{L}_{\mathrm{dir}}
=
\frac{1}{|V|}
\sum_{ij}
\Bigl(D^*_{ij}w_+ + (1-D^*_{ij})\Bigr)
V_{ij}\,
\mathrm{BCE}(\ell_{ij},\tilde{D}_{ij}),
\]
where $\mathrm{BCE}(\ell,y)=-y\log\sigma(\ell)-(1-y)\log(1-\sigma(\ell)).$ To address class imbalance, the positive-class weight $w_+$ is adapted online based on recall: $\hat r_t = 0.95\hat r_{t-1}+0.05r_t,$ and
\[
w_+
=
\mathrm{clamp}\!\left(
\sqrt{\frac{n_-}{n_+}}
\Bigl[1+2\,\mathrm{clamp}(0.65-\hat r_t,-0.3,0.3)\Bigr],
1,5
\right),
\]
where $r_t$ is the training recall of epoch $t$, computed from true-positive and false-negative counts summed over all GPUs and excluding replayed hard examples. $\hat r_t$, and thus the recall factor in $w_+$, is updated once per epoch, while $n_+$ and $n_-$ are counted on the current batch. 
This stabilizes learning under severe edge sparsity while adapting the recall--precision tradeoff.

\textbf{(b) Auxiliary Losses.} $\mathcal{L}_{\mathrm{asym}}$ suppresses the reverse direction of true edges if its confidence exceeds a threshold.
$\mathcal{L}_{\mathrm{cal}}$ penalizes logit gaps between opposite directions of a true edge above a margin $\Delta_{\max}$ to limit overconfidence.
$\mathcal{L}_{\mathrm{bdir}}$ supervises direction only for edges whose existence has been detected, and $\mathcal{L}_{\mathrm{skel}}$ supervises the undirected skeleton.
$\mathcal{L}_{\mathrm{conf}}$ supervises the confounder module, which receives no edge-loss gradients since $\hat C_{ij}$ enters the decoder detached.

Finally, $\mathcal{L}_{\mathrm{fp}}$ and $\mathcal{L}_{\mathrm{den}}$ penalize overly dense graphs, while $\mathcal{L}_{\mathrm{pma}}$ and $\mathcal{L}_{\mathrm{gate}}$ regularize the pooling and gating mechanisms.
Full definitions and hyperparameters are given in Appendix~\ref{app:loss_table}.
\endgroup
% 
%════════════════════════════════════════════════════════════════
\section{Training Pipeline}
\label{sec:optimisation}
% ════════════════════════════════════════════════════════════════

% \subsection{Optimizer: Schedule-Free AdamW}

% We use Schedule-Free AdamW~\citep{defazio2024road} with the following hyperparameters:

% \begin{table}[h]
% \centering
% \caption{Optimization hyperparameters.}
% \begin{tabular}{ll}
% \toprule
% \textbf{Parameter} & \textbf{Value} \\
% \midrule
% Base learning rate & $4 \times 10^{-4}$ \\
% Minimum learning rate & $1 \times 10^{-5}$ \\
% LR decay schedule & Cosine over 300 epochs, then flat \\
% $\beta_1, \beta_2$ & $0.9, 0.99$ \\
% Weight decay & $10^{-4}$ (2D+ params only; biases and RMSNorm scales excluded) \\
% Gradient clipping & Global norm 1.0 \\
% Batch size & 24 (3 per GPU $\times$ 8 GPUs) \\
% Warmup & ${\sim}1$ epoch (built-in to Schedule-Free) \\
% Training epochs & 1000 \\
% \bottomrule
% \end{tabular}
% \end{table}

% Schedule-Free AdamW maintains two parameter sequences: $\bm{z}$ (gradient iterates used during training) and $\bm{x}$ (evaluation iterates, a running average of $\bm{z}$).  Training uses $\bm{z}$-params; evaluation and checkpointing use $\bm{x}$-params, switched via \texttt{optimizer.train()}/\texttt{optimizer.eval()}.  The $\beta_2 = 0.99$ (up from the default 0.95) provides smoother variance estimates across data regeneration shifts.

We train using Schedule-Free AdamW~\citep{defazio2024road} with base learning rate $4\times 10^{-4}$, cosine decay to $\eta_{\min}=10^{-5}$ over the 300 training epochs, and momentum parameters $\beta_1=0.9$, $\beta_2=0.99$. 
Weight decay of $10^{-4}$ is applied to the weight matrices (except those of the attention-bias network) and the learned tokens, but not to biases or normalization parameters. Gradients are clipped to a global norm of 1.0. 
Training runs for 300 epochs with batch size 24. Schedule-Free AdamW maintains training parameters $\bm{\theta}$ and evaluation parameters $\bar{\bm{\theta}}$, where $\bar{\bm{\theta}}$ is a running average of $\bm{\theta}$. Updates are applied to $\bm{\theta}$, while evaluation and checkpointing use $\bar{\bm{\theta}}$.

% \noindent\textbf{Cosine LR envelope.}
% A manual cosine annealing envelope is applied on top of Schedule-Free:
% \[
%   \eta_t = \eta_{\min} + \frac{\eta_{\max} - \eta_{\min}}{2}\left(1 + \cos\!\left(\frac{\pi \cdot \min(t, T_{\mathrm{decay}})}{T_{\mathrm{decay}}}\right)\right),
% \]
% with $\eta_{\max} = 4 \times 10^{-4}$, $\eta_{\min} = 10^{-5}$, and $T_{\mathrm{decay}} = 300$ epochs.  After epoch 300, the LR stays flat at $\eta_{\min}$.  This combines Schedule-Free's implicit averaging benefits with a macro-level LR decay to stabilise late-stage training.

We train using bfloat16 mixed precision with selective float32 computation for numerically sensitive operations (e.g., normalization, matrix inversions, and loss computation). 
We use distributed data parallelism across 8 H100 GPUs with synchronized gradients and data-parallel sharding. Global statistics for adaptive loss weighting are synchronized across workers.

\notationchange{We continuously regenerate training data through parallel sampling of synthetic SCMs.} 
To prevent shortcut learning, we apply permutation augmentation—randomly permuting variable indices with corresponding relabeling of $\bm{D}^*$ and $\bm{C}^*$—removing any variable-index ordering left by the generator and encouraging reliance on distributional and relational signals. Permutation does not remove var-sortability (marginal variance increasing along the causal order;~\citealp{reisach2021beware}), which is a property of the data values; the encoder inputs and all pairwise statistics but one are scale-free up to numerical floors (Appendix~\ref{app:architecture_details}). We also use hard mining: tasks from recent epochs on which the model was confidently wrong are replayed with weight 2 in $\mathcal{L}_{\mathrm{dir}}$ and $\mathcal{L}_{\mathrm{skel}}$.

\section{Experimental Results}\label{sec:exp}
We train only on synthetic SCMs from a modified DoWhy generator~\citep{sharma2020dowhy,blobaum2024dowhy} and a custom diverse-graph generator, with supervision for directed edges and latent confounding (Appendix~\ref{sec:training-data}).

\begin{table}[t]
\centering
\footnotesize
\setlength{\tabcolsep}{4pt}
\renewcommand{\arraystretch}{1.15}
\begin{tabular}{@{}lccccc@{}}
\toprule
\textbf{Method}
 & \textbf{Cov.}
 & \textbf{Avg.\ AUROC $\uparrow$}
 & \textbf{Avg.\ $F_1$ $\uparrow$}
 & \textbf{Avg.\ Skel-$F_1$ $\uparrow$}
 & \textbf{Avg.\ SHD$/d$ $\downarrow$} \\
\midrule
\multicolumn{6}{l}{\emph{Classical baselines}} \\
PC ($\alpha{=}0.01$) \citep{spirtes2000causation}
          & $14/14$ & $0.713$ & $0.449$ & $0.665$ & $1.309$ \\
PC ($\alpha{=}0.05$) \citep{spirtes2000causation}
          & $14/14$ & $0.724$ & $0.451$ & $0.660$ & $1.394$ \\
FCI ($\alpha{=}0.05$) \citep{spirtes2000causation}
          & $14/14$ & $0.742$ & $0.463$ & $0.664$ & $1.463$ \\
GES \citep{chickering2002optimal}
          & $14/14$ & $0.760$ & $0.505$ & $0.686$ & $1.414$ \\
GRaSP \citep{lam2022greedy}
          & $14/14$ & $0.762$ & $0.516$ & $0.692$ & $1.380$ \\
DirectLiNGAM \citep{shimizu2011directlingam}
          & $14/14$ & $0.652$ & $0.295$ & $0.533$ & $2.453$ \\
ICA-LiNGAM \citep{shimizu2006linear}
          & $14/14$ & $0.666$ & $0.327$ & $0.554$ & $2.400$ \\
NOTEARS-Linear \citep{zheng2018dags}
          & $14/14$ & $0.608$ & $0.309$ & $0.532$ & $1.190$ \\
DAGMA (linear) \citep{bello2022dagma}
          & $14/14$ & $0.615$ & $0.309$ & $0.600$ & $1.268$ \\
SCORE \citep{rolland2022score}
          & $14/14$ & $0.672$ & $0.321$ & $0.565$ & $3.047$ \\
Naive correlation
          & $14/14$ & $0.682$ & $0.304$ & $0.461$ & $2.665$ \\
\addlinespace[3pt]
\multicolumn{6}{l}{\emph{Amortized / foundation-model methods}} \\
AVICI \citep{lorch2022avici}
          & $14/14$ & $0.804$ & $0.407$ & $0.532$ & $1.003$ \\
SEA \citep{wu2024recipe}
          & $14/14$ & $0.797$ & $0.311$ & $0.454$ & $1.755$ \\
TabCausal \citep{li2026tabcausal}
          & $14/14$ & $0.838$ & $0.495$ & $0.617$ & $0.956$ \\
TabCausal-v2 \citep{li2026tabcausal}
          & $14/14$ & $0.831$ & $0.509$ & $0.696$ & $0.984$ \\
Arrow \citep{thompson2026arrow}
          & $14/14$ & $0.744$ & $0.328$ & $0.549$ & $1.349$ \\
CauScale \citep{peng2026causcale}
          & $14/14$ & $0.800$ & $0.365$ & $0.524$ & $1.191$ \\
CDFM \citep{qiao2026cdfm}
          & $14/14$ & $0.856$ & $0.469$ & $0.631$ & $1.581$ \\
\addlinespace[3pt]
\textbf{\cdtname (ours)}
          & $14/14$ & $\bm{0.883}$ & $\bm{0.531}$ & $\bm{0.709}$ & $\bm{0.875}$ \\
\bottomrule
\end{tabular}
\vspace{-6pt}
\caption{Performance on 14 real-world and semi-synthetic benchmarks. Metrics are macro-averaged over the datasets; \emph{Cov.} gives the number each method completes. $F_1$ measures directed-edge recovery, Skel-$F_1$ undirected recovery, and $\mathrm{SHD}/d$ normalized graph-edit error. Arrows indicate the preferred direction. Per-dataset results are in Appendix~\ref{app:transformer_per_dataset}.}
\label{tab:method-comparison-avg}
\end{table}

\subsection{Real-World Benchmark Evaluation}
\label{sec:real_world_experiments}

We evaluate \cdtname on 14 real-world and semi-synthetic benchmarks covering Bayesian networks, nonlinear variants, biological systems and high-dimensional operational graphs.
These include \textsc{asia}, \textsc{child}, \textsc{insurance}, \textsc{sachs}, \textsc{causal\_chambers\_lt}, \textsc{ecoli\_like} and three \textsc{petshop} variants, with graph sizes up to $D=46$. We also evaluate on the T{\"u}bingen cause--effect pairs benchmark~\citep{mooij2016tuebingen}, a collection of real-world bivariate tasks where the goal is to infer whether $X \to Y$ or $Y \to X$. This setting is challenging because conditional-independence tests and multi-variable constraints are unavailable, requiring reliance on asymmetric distributional signals such as nonlinearity, non-Gaussianity, and noise--mechanism independence.
See Appendix~\ref{sec:datasets} for further dataset details.

\textbf{\notationchange{Baseline Methods.}} We compare representative classical methods based on conditional independence tests, score-based search, LiNGAM and continuous DAG optimization, along with SCORE and naive correlation (Table~\ref{tab:method-comparison-avg}). Amortized baselines are AVICI~\citep{lorch2022avici}, SEA~\citep{wu2024recipe}, TabCausal and its later checkpoint TabCausal-v2~\citep{li2026tabcausal}, Arrow~\citep{thompson2026arrow}, CauScale~\citep{peng2026causcale} and CDFM~\citep{qiao2026cdfm}, run with their public code and weights (Appendix~\ref{app:baselines}). For all baselines, missing values are mean-imputed per variable. For T{\"u}bingen, we omit PC, FCI, GES and GRaSP, which return Markov equivalence classes ($X \to Y$ and $Y \to X$ are Markov equivalent); NOTEARS-Linear, DAGMA and naive correlation, which cannot orient a standardized pair; SEA, whose released configuration needs at least five variables; and CauScale, which is pretrained on ten or more variables. For this benchmark, we add the bivariate methods SLOPPY~\citep{marx2019identifiability}, HECI~\citep{xu2022inferring} and RECI~\citep{bloebaum2018reci} (with degree-3 polynomial regression).

\textbf{\notationchange{Evaluation Protocol.}} Table~\ref{tab:method-comparison-avg} reports edge-ranking quality (AUROC), directed and skeleton $F_1$, normalized structural Hamming distance ($\mathrm{SHD}/d$), and coverage; Appendix~\ref{sec:eval} defines the metrics. \cdtname uses a single fixed checkpoint without retraining or per-dataset tuning. Its inference protocol (Appendix~\ref{app:IP}) uses \textcolor{black}{column sanitization, averaging over $K=10$ passes with column permutation and $90\%$ row subsampling, a fixed edge threshold of $0.5$ with no temperature scaling}, and self-consistency pruning. These inference-time steps are only applied to \cdtname.

\begin{table}[t]
\centering
\small
\renewcommand{\arraystretch}{1.15}
\begin{tabular}{@{}lccc@{}}
\toprule
\textbf{Method}
 & \textbf{Correct / Total}
 & \textbf{Unweighted $\uparrow$}
 & \textbf{Weighted $\uparrow$} \\
\midrule
\multicolumn{4}{l}{\emph{Classical baselines}} \\
DirectLiNGAM \citep{shimizu2011directlingam}
          & $50/102$ & $49.0\%$ & $49.8\%$ \\
ICA-LiNGAM \citep{shimizu2006linear}
          & $58.5/102$ & $57.4\%$ & $54.6\%$ \\
SCORE \citep{rolland2022score}
          & $64/102$ & $62.7\%$ & $62.1\%$ \\
\addlinespace[2pt]
\multicolumn{4}{l}{\emph{Amortized / foundation-model methods}} \\
AVICI \citep{lorch2022avici}$^{\ddagger}$
          & $56/102$ & $54.9\%$ & $54.8\%$ \\
TabCausal \citep{li2026tabcausal}
          & $67/102$ & $65.7\%$ & $61.0\%$ \\
TabCausal-v2 \citep{li2026tabcausal}
          & $62/102$ & $60.8\%$ & $59.8\%$ \\
Arrow \citep{thompson2026arrow}$^{\ddagger}$
          & $51/102$ & $50.0\%$ & $46.0\%$ \\
CDFM \citep{qiao2026cdfm}$^{\ddagger}$
          & $69/102$ & $67.6\%$ & $57.4\%$ \\
\addlinespace[2pt]
\multicolumn{4}{l}{\emph{Bivariate-specialized methods}} \\
RECI~\citep{bloebaum2018reci}
          & $60/102$ & $58.8\%$ & $63.7\%$ \\
HECI~\citep{xu2022inferring}$^{\dagger}$
          & $61.5/102$ & $60.3\%$ & $68.9\%$ \\
SLOPPY~\citep{marx2019identifiability}
          & $61.5/102$ & $60.3\%$ & $\bm{69.8\%}$ \\
\addlinespace[3pt]
\textbf{\cdtname (ours)}
          & $\bm{78/102}$ & $\bm{76.5\%}$ & $68.4\%$ \\
\bottomrule
\end{tabular}

\vspace{2pt}
{\footnotesize Unoriented or failed pairs count half. $^{\dagger}$Fails on three pairs with a binary variable. $^{\ddagger}$At most $2{,}000$ rows (CDFM: $5{,}000$).\par}
\vspace{-6pt}
\caption{Results on the T{\"u}bingen cause--effect pairs benchmark. 
We evaluate 102 pairs with known direction and positive weights. Weighted accuracy down-weights pairs from the same source data.
\textcolor{black}{\cdtname averages $P(X\to Y)$ and $P(Y\to X)$ over $K=10$ passes with random column order and, for large datasets, random row subsamples, then selects the direction with the larger average; column sanitization, thresholding, and pruning are omitted. The amortized baselines are likewise oriented by the larger of their two edge probabilities.}}
\label{tab:tuebingen}
\end{table}
Across the $14$ datasets, \cdtname has the best average on all four metrics. Its lead over the best baseline is clear in AUROC ($0.883$ versus $0.856$ for CDFM) and $\mathrm{SHD}/d$ ($0.875$ versus $0.956$ for TabCausal), but small in directed $F_1$ ($0.531$ versus $0.516$ for GRaSP) and skeleton $F_1$ ($0.709$ versus $0.696$ for TabCausal-v2). For the classical methods, which output only a graph, AUROC is computed from the binary prediction, i.e., at a single operating point (Appendix~\ref{sec:eval}). For \cdtname and the amortized baselines, it ranks the edge probabilities. No method is best on every dataset: \cdtname has the best $F_1$ on $3$ of the $14$ datasets, as does GRaSP (Appendix~\ref{app:transformer_per_dataset}).

On the T{\"u}bingen cause--effect pairs (Table~\ref{tab:tuebingen}), \cdtname has the highest unweighted accuracy overall ($76.5\%$; CDFM follows with $67.6\%$) and the highest weighted accuracy among amortized methods ($68.4\%$; TabCausal follows with $61.0\%$). It exceeds the strongest general-purpose classical baseline (SCORE) by $13.8$ percentage points in unweighted and $6.3$ points in weighted accuracy. The bivariate methods SLOPPY ($69.8\%$) and HECI ($68.9\%$) have a higher weighted accuracy; unweighted, both reach $60.3\%$ and RECI $58.8\%$. The weighted accuracy of \cdtname is below its unweighted accuracy, indicating that its errors occur on pairs with larger benchmark weights.

%(which is typically between $65\%$ to $70\%$) \textcolor{black}{check ?}.

%\textbf{Test-time ablations.}
%We perform test-time ablations by replacing intermediate activations with neutral values at inference, without retraining, isolating each component’s contribution. Pairwise statistics and the dedicated direction pathway are the primary drivers of performance, with large drops when removed, while triangular refinement provides additional gains via higher-order reasoning. Pooling and confounder feedback yield smaller improvements, indicating that performance is mainly driven by explicit causal inductive biases (Appendix~\ref{sec:test-time-ablation}).

\textbf{Missing Data.}
\cdtname takes incomplete tables without imputation (Appendix~\ref{sec:missing-data}).
Under purely random missingness, it remains first on all four averages at $10\%$, but at $30\%$ it is best only in AUROC and $\mathrm{SHD}/d$.
It degrades more under selection-driven missingness, where at $30\%$ CDFM with its native handling changes least among the amortized methods.

\begingroup\color{black}
\subsection{Training-Time Ablations}
\label{sec:train-ablation}

To quantify the contribution of individual architectural components, we remove each one \emph{before training} and retrain the model from scratch, so that the remaining network can adapt to the missing component. Each ablated model uses the same synthetic data generator, optimizer and $300$-epoch schedule as \cdtname, and differs only in the removed component.

\begin{itemize}[nosep,leftmargin=*,label=\notationchange{\textbullet}]
\item \textbf{Confounder Module.} The confounder tokens, their cross-attention, the loading network and the noisy-OR aggregation are removed ($10.6$M parameters), together with $\mathcal{L}_{\mathrm{conf}}$. The existence head receives a constant zero instead of $\hat C_{ij}$.

\item \textbf{Pairwise Statistics.} The $45$ pairwise statistics are not computed. This removes the stat-conditioned attention bias, the statistic projections of the existence and direction heads, the feature gate, and the statistics bias of the confounder module, so that dependence and orientation must be learned from the raw data matrix alone.

\item \textbf{Antisymmetric Features.} All direction-asymmetric statistics (the antisymmetric group and the asymmetric collider-child pair) are set to zero before any module reads them. The symmetric statistics, collider evidence, partial-correlation gap and reliability metadata are kept. Orientation can then come only from the embedding difference $\bm{z}_i-\bm{z}_j$, the role score $r_{ij}$ and triangular refinement.

\item \textbf{Triangular Refinement.} The pair tensor, the $R=3$ refinement rounds, the learned skeleton extraction $\mathrm{skel\_fn}$ (Appendix~\ref{app:architecture_details}) and the learned skeleton/direction blend are removed ($1.7$M parameters); $\mathcal{L}_{\mathrm{skel}}$ then uses $\max(\ell_{ij},\ell_{ji})$ instead of $\mathrm{skel\_fn}$. The edge logit is the per-pair decoder output plus the role score, so the decoder no longer combines evidence across pairs.
\end{itemize}

All models are evaluated at their last epoch on the 14 benchmarks using the inference protocol and metrics in Table~\ref{tab:method-comparison-avg}. Each ablated model is trained once. \notationchange{We fixed the configuration of \cdtname before running the ablations.}

\begin{table}[t]
\centering
\small
\color{black}
\renewcommand{\arraystretch}{1.15}
\begin{tabular}{@{}lccccc@{}}
\toprule
\textbf{Model}
 & \textbf{Cov.}
 & \textbf{Avg.\ AUROC $\uparrow$}
 & \textbf{Avg.\ $F_1$ $\uparrow$}
 & \textbf{Avg.\ Skel-$F_1$ $\uparrow$}
 & \textbf{Avg.\ SHD$/d$ $\downarrow$} \\
\midrule
\cdtname (full)           & $14/14$ & $\bm{0.883}$ & $0.531$ & $0.709$ & $0.875$ \\
\midrule
$-$Confounder module      & $14/14$ & $0.873$ & $\bm{0.571}$ & $\bm{0.715}$ & $\bm{0.828}$ \\
$-$Pairwise statistics    & $14/14$ & $0.690$ & $0.169$ & $0.299$ & $1.429$ \\
$-$Antisymmetric features & $14/14$ & $0.869$ & $0.506$ & $0.687$ & $0.976$ \\
$-$Triangular refinement  & $14/14$ & $0.849$ & $0.487$ & $0.647$ & $0.912$ \\
\bottomrule
\end{tabular}
\vspace{-6pt}
\caption{Training-time ablations of \cdtname on the 14 benchmark datasets. Each ablated model removes one component and is retrained from scratch. Metrics are macro-averaged across datasets as in Table~\ref{tab:method-comparison-avg}. Bold marks the best value in each column.}
\label{tab:train-ablation}
\end{table}

\notationchange{Without the pairwise statistics,} average $F_1$ drops from $0.531$ to $0.169$, skeleton $F_1$ from $0.709$ to $0.299$ and AUROC from $0.883$ to $0.690$, while $\mathrm{SHD}/d$ rises from $0.875$ to $1.429$ (Table~\ref{tab:train-ablation}). The raw-data encoder alone does poorly on these benchmarks. \notationchange{The other three ablations change $F_1$ by $0.025$ to $0.044$.}

Removing refinement reduces average $F_1$ by $0.044$, skeleton $F_1$ by $0.062$ and AUROC by $0.034$, and increases $\mathrm{SHD}/d$ from $0.875$ to $0.912$. The drop in skeleton $F_1$ is the largest after the pairwise statistics, consistent with the role of refinement: it aggregates evidence over chains, forks and colliders, which helps identify adjacent pairs beyond what per-pair features reveal (this ablation also removes the learned skeleton extraction).

\notationchange{Without the antisymmetric features,} average $F_1$ drops by $0.025$, skeleton $F_1$ by $0.022$ and AUROC by $0.014$, while $\mathrm{SHD}/d$ rises from $0.875$ to $0.976$, more than without refinement. On these multivariate benchmarks, the embedding difference, the role score and triangular refinement can partly compensate for the missing asymmetry statistics, for instance by orienting edges through V-structures.

\notationchange{Removing the confounder module improves average $F_1$ ($0.571$ vs.\ $0.531$), skeleton $F_1$ ($0.715$ vs.\ $0.709$) and $\mathrm{SHD}/d$ ($0.828$ vs.\ $0.875$), and lowers AUROC ($0.873$ vs.\ $0.883$). This ablation also removes the confounding output $\hat{\bm C}$, which we evaluate in Section~\ref{sec:confounder-eval}.}
\endgroup

\begin{table}[t]
\centering
\small
\setlength{\tabcolsep}{3.5pt}
\renewcommand{\arraystretch}{1.15}
\begin{tabular}{@{}>{\raggedright\arraybackslash}p{0.36\textwidth}cccccc@{}}
\toprule
 & \multicolumn{3}{c}{\textbf{All pairs}} & \multicolumn{3}{c}{\textbf{No ancestor relation}} \\
\cmidrule(lr){2-4}\cmidrule(lr){5-7}
\textbf{Setting} & FCI & CDFM & \cdtname & FCI & CDFM & \cdtname \\
\midrule
% columns: Setting & all pairs, F1 of: FCI & CDFM & FoundCause & pairs without ancestor relation, F1 of: FCI & CDFM & FoundCause
Simulation, in-distribution ($40$ tasks) & $0.197$ & $0.111$ & $\bm{0.299}$ & $0.209$ & $0.097$ & $\bm{0.334}$ \\
Semi-synthetic, $15\%$ hub variables hidden ($5$ networks, $42$ hidden sets) & $\bm{0.202}$ & $0.126$ & $0.152$ & $\bm{0.228}$ & $0.128$ & $0.158$ \\
Benchmark datasets, $15\%$ of variables hidden ($14$ datasets, $129$ hidden sets) & $0.064$ & $0.052$ & $\bm{0.087}$ & $0.075$ & $0.077$ & $\bm{0.158}$ \\
\bottomrule
\end{tabular}
\vspace{-6pt}
\caption{Latent confounder recovery: $F_1$ of the pairs flagged as confounded, with each method in its default configuration (FCI: bidirected edge; CDFM: edges in both directions at the threshold of its package; \cdtname: $\hat C_{ij}>0.5$), on all pairs of observed variables (\emph{All pairs}) and on the pairs in which neither variable is an ancestor of the other (\emph{No ancestor relation}). The first row pools the pairs of all tasks; the other two pool them per dataset and average over the datasets. Bold marks the highest value of the three methods.}
\label{tab:confounder-eval}
\end{table}

\begingroup\color{black}
\subsection{Latent Confounder Recovery}
\label{sec:confounder-eval}

We test whether $\hat{\bm C}$ identifies pairs that share a hidden common cause. A pair of observed variables is confounded if both are reached from the same hidden variable through hidden variables only, whether or not one also causes the other. Of the compared methods, only \cdtname has a separate, explicit confounding output: CDFM is trained to mark a confounded pair with edges in both directions of its single adjacency output, and FCI marks it with a bidirected edge. We compare the three methods by the $F_1$ of the pairs they flag, using their default configuration (Table~\ref{tab:confounder-eval}; details in Appendix~\ref{app:confounder}). \emph{All pairs} evaluates every pair of observed variables. \emph{No ancestor relation} excludes pairs where one variable is an ancestor of the other: in the maximal ancestral graph, such a confounded pair (e.g., $U\to X$, $U\to Y$, and $X\to Y$ with $U$ hidden) is joined by a directed edge, so a correct FCI output never marks it as bidirected, and CDFM is trained to prioritize a direct edge. The values below refer to all pairs. In distribution ($40$ tasks from the training generator), \cdtname has the highest $F_1$ ($0.299$; FCI $0.197$, CDFM $0.111$). With hub variables (at least two children; $15\%$ of the variables) hidden in five semi-synthetic benchmark networks, FCI has the highest score ($0.202$; \cdtname $0.152$, CDFM $0.126$). With a random $15\%$ of the variables hidden in each of the 14 benchmark datasets, all three have low scores (\cdtname $0.087$, FCI $0.064$, CDFM $0.052$), and each has the highest $F_1$ on four or five datasets. The best method in each setting is the same without the ancestor-related pairs.
\endgroup

\section{Conclusion}

We presented \cdtname, an amortized model that predicts directed causal relations and latent confounding from observational data. Trained entirely on synthetic SCMs, it achieves the best average AUROC and $\mathrm{SHD}/d$ among evaluated methods on 14 real-world and semi-synthetic benchmarks, and narrowly the best average $F_1$ and skeleton $F_1$. \notationchange{Its $F_1$ is highest on 3 of these datasets.} \notationchange{Its confounding output has the highest $F_1$ in distribution; the leading method varies across the benchmark settings with hidden variables.} \textcolor{black}{In training-time ablations, pairwise statistics are essential for edge recovery on the 14 benchmarks; triangular refinement and antisymmetric features have small effects on $F_1$. \notationchange{Removing the confounder module improves average $F_1$ and removes the confounding output.}}
% Extrapolation experiments demonstrate that edge ranking degrades gracefully beyond the training regime.
These results support combining amortized inference with statistical and, to a lesser extent, structural inductive biases. We plan to open-source \cdtname.

\section*{Statement on the Use of Generative AI Tools}
We disclose that we used generative AI tools (Claude Code and Codex) as assistants for three supporting tasks: (i)~\emph{writing}, to refine the wording and structure of text drafted by the authors and to draft additional text under their direction; (ii)~\emph{proofreading}, to check grammar, notation, cross-references and the consistency of reported numbers; and (iii)~\emph{experiments}, to help implement, run and check the evaluation scripts for baselines and analyses. Research questions, methodological decisions, the experimental design and the interpretation of results were made by the authors, who reviewed, verified and, where necessary, corrected every AI-assisted contribution. The authors take full responsibility for the content of this paper.

\bibliographystyle{abbrvnat}
\bibliography{ref}

\begin{thebibliography}{75}
\providecommand{\natexlab}[1]{#1}
\providecommand{\url}[1]{\texttt{#1}}
\expandafter\ifx\csname urlstyle\endcsname\relax
  \providecommand{\doi}[1]{doi: #1}\else
  \providecommand{\doi}{doi: \begingroup \urlstyle{rm}\Url}\fi

\bibitem[Annadani et~al.(2023)Annadani, Pawlowski, Jennings, Bauer, Zhang, and
  Gong]{annadani2023bayesdag}
Y.~Annadani, N.~Pawlowski, J.~Jennings, S.~Bauer, C.~Zhang, and W.~Gong.
\newblock {BayesDAG}: Gradient-based posterior inference for causal discovery.
\newblock In \emph{Advances in Neural Information Processing Systems},
  volume~36, 2023.

\bibitem[Ashman et~al.(2023)Ashman, Ma, Hilmkil, Jennings, and
  Zhang]{ashman2023neural}
M.~Ashman, C.~Ma, A.~Hilmkil, J.~Jennings, and C.~Zhang.
\newblock Causal reasoning in the presence of latent confounders via neural
  {ADMG} learning.
\newblock In \emph{International Conference on Learning Representations}, 2023.
\newblock URL \url{https://openreview.net/forum?id=dcN0CaXQhT}.

\bibitem[Balazadeh et~al.(2025)Balazadeh, Kamkari, Thomas, Ma, Li, Cresswell,
  and Krishnan]{balazadeh2026causalpfn}
V.~Balazadeh, H.~Kamkari, V.~Thomas, J.~Ma, B.~Li, J.~C. Cresswell, and
  R.~Krishnan.
\newblock Causal{PFN}: Amortized causal effect estimation via in-context
  learning.
\newblock In \emph{The Thirty-ninth Annual Conference on Neural Information
  Processing Systems}, 2025.
\newblock URL \url{https://openreview.net/forum?id=RblaNJGx8C}.

\bibitem[Bello et~al.(2022)Bello, Aragam, and Ravikumar]{bello2022dagma}
K.~Bello, B.~Aragam, and P.~Ravikumar.
\newblock {DAGMA}: Learning {DAGs} via {M}-matrices and a log-determinant
  acyclicity characterization.
\newblock \emph{Advances in Neural Information Processing Systems},
  35:\penalty0 8226--8239, 2022.

\bibitem[Bhattacharya et~al.(2021)Bhattacharya, Nagarajan, Malinsky, and
  Shpitser]{bhattacharya2021differentiable}
R.~Bhattacharya, T.~Nagarajan, D.~Malinsky, and I.~Shpitser.
\newblock Differentiable causal discovery under unmeasured confounding.
\newblock In \emph{Proceedings of the 24th International Conference on
  Artificial Intelligence and Statistics}, volume 130 of \emph{Proceedings of
  Machine Learning Research}, pages 2314--2322. PMLR, 2021.
\newblock URL \url{https://proceedings.mlr.press/v130/bhattacharya21a.html}.

\bibitem[Bl{\"o}baum et~al.(2018)Bl{\"o}baum, Janzing, Washio, Shimizu, and
  Sch{\"o}lkopf]{bloebaum2018reci}
P.~Bl{\"o}baum, D.~Janzing, T.~Washio, S.~Shimizu, and B.~Sch{\"o}lkopf.
\newblock Cause-effect inference by comparing regression errors.
\newblock In \emph{Proceedings of the Twenty-First International Conference on
  Artificial Intelligence and Statistics}, volume~84 of \emph{Proceedings of
  Machine Learning Research}, pages 900--909. PMLR, 2018.

\bibitem[Bl{\"o}baum et~al.(2024)Bl{\"o}baum, G{\"o}tz, Budhathoki, Mastakouri,
  and Janzing]{blobaum2024dowhy}
P.~Bl{\"o}baum, P.~G{\"o}tz, K.~Budhathoki, A.~A. Mastakouri, and D.~Janzing.
\newblock {DoWhy-GCM}: An extension of {DoWhy} for causal inference in
  graphical causal models.
\newblock \emph{Journal of Machine Learning Research}, 25\penalty0
  (147):\penalty0 1--7, 2024.

\bibitem[Brouillard et~al.(2020)Brouillard, Lachapelle, Lacoste,
  Lacoste-Julien, and Drouin]{brouillard2020dcdi}
P.~Brouillard, S.~Lachapelle, A.~Lacoste, S.~Lacoste-Julien, and A.~Drouin.
\newblock Differentiable causal discovery from interventional data.
\newblock In \emph{Advances in Neural Information Processing Systems},
  volume~33, 2020.
\newblock URL
  \url{https://papers.nips.cc/paper/2020/hash/f8b7aa3a0d349d9562b424160ad18612-Abstract.html}.

\bibitem[Bühlmann et~al.(2014)Bühlmann, Peters, and Ernest]{buhlmann2014cam}
P.~Bühlmann, J.~Peters, and J.~Ernest.
\newblock {CAM}: Causal additive models, high-dimensional order search and
  penalized regression.
\newblock \emph{The Annals of Statistics}, 42\penalty0 (6):\penalty0
  2526--2556, 2014.
\newblock \doi{10.1214/14-AOS1260}.

\bibitem[Chao et~al.(2024)Chao, Bl{\"o}baum, Patel, and
  Kasiviswanathan]{Chao2023ModelingCM}
P.~Chao, P.~Bl{\"o}baum, S.~Patel, and S.~P. Kasiviswanathan.
\newblock Modeling causal mechanisms with diffusion models for interventional
  and counterfactual queries.
\newblock \emph{Transactions on Machine Learning Research}, 2024.

\bibitem[Chen et~al.(2026)Chen, Guan, Qian, Zhang, Cui, Yang, Wu, and
  Kuang]{chen2026dagfm}
Y.~Chen, Z.~Guan, H.~Qian, X.~Zhang, P.~Cui, Y.~Yang, F.~Wu, and K.~Kuang.
\newblock {DAG-FM}: A foundation model for causal discovery under heterogeneous
  causal mechanisms.
\newblock \emph{arXiv preprint arXiv:2607.11510}, 2026.

\bibitem[Chickering(2002)]{chickering2002optimal}
D.~M. Chickering.
\newblock Optimal structure identification with greedy search.
\newblock \emph{Journal of Machine Learning Research}, 3:\penalty0 507--554,
  2002.

\bibitem[Colombo and Maathuis(2014)]{colombo2014order}
D.~Colombo and M.~H. Maathuis.
\newblock Order-independent constraint-based causal structure learning.
\newblock \emph{Journal of Machine Learning Research}, 15\penalty0
  (116):\penalty0 3921--3962, 2014.
\newblock URL \url{https://jmlr.org/papers/v15/colombo14a.html}.

\bibitem[Daniu{\v{s}}is et~al.(2010)Daniu{\v{s}}is, Janzing, Mooij,
  Zscheischler, Steudel, Zhang, and Sch{\"o}lkopf]{daniusis2010igci}
P.~Daniu{\v{s}}is, D.~Janzing, J.~Mooij, J.~Zscheischler, B.~Steudel, K.~Zhang,
  and B.~Sch{\"o}lkopf.
\newblock Inferring deterministic causal relations.
\newblock In \emph{Proceedings of the 26th Conference on Uncertainty in
  Artificial Intelligence (UAI)}, 2010.

\bibitem[Defazio et~al.(2024)Defazio, Yang, Mehta, Mishchenko, Khaled, and
  Cutkosky]{defazio2024road}
A.~Defazio, X.~Yang, H.~Mehta, K.~Mishchenko, A.~Khaled, and A.~Cutkosky.
\newblock The road less scheduled.
\newblock \emph{Advances in Neural Information Processing Systems},
  37:\penalty0 9974--10007, 2024.

\bibitem[Dhir et~al.(2025)Dhir, Ashman, Requeima, and van~der Wilk]{dhir2025a}
A.~Dhir, M.~Ashman, J.~Requeima, and M.~van~der Wilk.
\newblock A meta-learning approach to {B}ayesian causal discovery.
\newblock In \emph{The Thirteenth International Conference on Learning
  Representations}, 2025.
\newblock URL \url{https://openreview.net/forum?id=eeJz7eDWKO}.

\bibitem[Gamella et~al.(2025)Gamella, Peters, and
  B{\"u}hlmann]{gamella2025causal}
J.~L. Gamella, J.~Peters, and P.~B{\"u}hlmann.
\newblock Causal chambers as a real-world physical testbed for {AI}
  methodology.
\newblock \emph{Nature Machine Intelligence}, 7\penalty0 (1):\penalty0
  107--118, 2025.

\bibitem[Geffner et~al.(2024)Geffner, Antor{\'a}n, Foster, Gong, Ma,
  K{\i}c{\i}man, Sharma, Lamb, Kukla, Pawlowski, Hilmkil, Jennings, Scetbon,
  Allamanis, and Zhang]{geffner2022deci}
T.~Geffner, J.~Antor{\'a}n, A.~Foster, W.~Gong, C.~Ma, E.~K{\i}c{\i}man,
  A.~Sharma, A.~Lamb, M.~Kukla, N.~Pawlowski, A.~Hilmkil, J.~Jennings,
  M.~Scetbon, M.~Allamanis, and C.~Zhang.
\newblock Deep end-to-end causal inference.
\newblock \emph{Transactions on Machine Learning Research}, 2024.
\newblock ISSN 2835-8856.
\newblock URL \url{https://openreview.net/forum?id=e6sqttxEGX}.

\bibitem[Glymour et~al.(2019)Glymour, Zhang, and Spirtes]{glymour2019review}
C.~Glymour, K.~Zhang, and P.~Spirtes.
\newblock Review of causal discovery methods based on graphical models.
\newblock \emph{Frontiers in Genetics}, 10:\penalty0 524, 2019.

\bibitem[Guan et~al.(2026)Guan, Chen, He, Tong, Hu, Qian, Wu, and
  Kuang]{guan2026dcdpfn}
Z.~Guan, Y.~Chen, Y.~He, Y.~Tong, Z.~Hu, H.~Qian, F.~Wu, and K.~Kuang.
\newblock {DCD-PFN}: A decoupling-aware foundation model for causal discovery.
\newblock \emph{arXiv preprint arXiv:2606.21212}, 2026.

\bibitem[Hardt et~al.(2024)Hardt, Orchard, Bl{\"o}baum, Kirschbaum, and
  Kasiviswanathan]{hardt2024petshop}
M.~Hardt, W.~R. Orchard, P.~Bl{\"o}baum, E.~Kirschbaum, and S.~Kasiviswanathan.
\newblock The {PetShop} dataset—finding causes of performance issues across
  microservices.
\newblock In \emph{Causal Learning and Reasoning}, pages 957--978. PMLR, 2024.

\bibitem[Hendrycks and Gimpel(2016)]{hendrycks2016gaussian}
D.~Hendrycks and K.~Gimpel.
\newblock {G}aussian error linear units ({GELUs}).
\newblock \emph{arXiv preprint arXiv:1606.08415}, 2016.

\bibitem[Hoyer et~al.(2008)Hoyer, Janzing, Mooij, Peters, and
  Sch{\"o}lkopf]{anm}
P.~Hoyer, D.~Janzing, J.~Mooij, J.~Peters, and B.~Sch{\"o}lkopf.
\newblock Nonlinear causal discovery with additive noise models.
\newblock In D.~Koller, D.~Schuurmans, Y.~Bengio, and L.~Bottou, editors,
  \emph{Advances in Neural Information Processing Systems}, volume~21, pages
  689--696. Curran Associates, Inc., 2008.

\bibitem[Jumper et~al.(2021)Jumper, Evans, Pritzel,
  et~al.]{jumper2021alphafold}
J.~Jumper, R.~Evans, A.~Pritzel, et~al.
\newblock Highly accurate protein structure prediction with {AlphaFold}.
\newblock \emph{Nature}, 596\penalty0 (7873):\penalty0 583--589, 2021.

\bibitem[Ke et~al.(2023)Ke, Chiappa, Wang, Bornschein, Goyal, Rey, Weber,
  Botvinick, Mozer, and Rezende]{ke2023learning}
N.~R. Ke, S.~Chiappa, J.~Wang, J.~Bornschein, A.~Goyal, M.~Rey, T.~Weber,
  M.~Botvinick, M.~C. Mozer, and D.~J. Rezende.
\newblock Learning to induce causal structure.
\newblock In \emph{International Conference on Learning Representations
  (ICLR)}, 2023.
\newblock URL \url{https://openreview.net/forum?id=hp\_RwhKDJ5}.

\bibitem[Kim et~al.(2025)Kim, Gibbs, Yun, Song, and Cho]{kim2025largescale}
J.-H. Kim, C.~S. Gibbs, S.~Yun, H.~O. Song, and K.~Cho.
\newblock Large-scale targeted cause discovery via learning from simulated
  data.
\newblock \emph{Transactions on Machine Learning Research}, 2025.
\newblock ISSN 2835-8856.
\newblock URL \url{https://openreview.net/forum?id=NVgy29IQw8}.

\bibitem[Lachapelle et~al.(2020)Lachapelle, Brouillard, Deleu, and
  Lacoste-Julien]{lachapelle2020gradient}
S.~Lachapelle, P.~Brouillard, T.~Deleu, and S.~Lacoste-Julien.
\newblock Gradient-based neural {DAG} learning.
\newblock In \emph{International Conference on Learning Representations}, 2020.
\newblock URL \url{https://openreview.net/forum?id=rklbKA4YDS}.

\bibitem[Lam et~al.(2022)Lam, Andrews, and Ramsey]{lam2022greedy}
W.-Y. Lam, B.~Andrews, and J.~Ramsey.
\newblock Greedy relaxations of the sparsest permutation algorithm.
\newblock In \emph{Uncertainty in Artificial Intelligence}, pages 1052--1062.
  PMLR, 2022.

\bibitem[Lee et~al.(2019)Lee, Lee, Kim, Kosiorek, Choi, and Teh]{lee2019set}
J.~Lee, Y.~Lee, J.~Kim, A.~Kosiorek, S.~Choi, and Y.~W. Teh.
\newblock Set {Transformer}: A framework for attention-based
  permutation-invariant neural networks.
\newblock In K.~Chaudhuri and R.~Salakhutdinov, editors, \emph{Proceedings of
  the 36th International Conference on Machine Learning (ICML)}, volume~97 of
  \emph{Proceedings of Machine Learning Research}, pages 3744--3753. PMLR,
  2019.

\bibitem[Li et~al.(2026{\natexlab{a}})Li, Liu, Wang, and Ye]{li2026causalarena}
Z.-R. Li, S.-Y. Liu, T.-Z. Wang, and H.-J. Ye.
\newblock {CausalArena}: Benchmarking causal discovery in the foundation model
  era.
\newblock \emph{arXiv preprint arXiv:2609.11897}, 2026{\natexlab{a}}.

\bibitem[Li et~al.(2026{\natexlab{b}})Li, Liu, Wang, and Ye]{li2026tabcausal}
Z.-R. Li, S.-Y. Liu, T.-Z. Wang, and H.-J. Ye.
\newblock {TabCausal}: Pretraining across causal environments for tabular
  causal discovery.
\newblock \emph{arXiv preprint arXiv:2605.31156}, 2026{\natexlab{b}}.

\bibitem[Lippe et~al.(2022)Lippe, Cohen, and Gavves]{lippe2022enco}
P.~Lippe, T.~Cohen, and E.~Gavves.
\newblock Efficient neural causal discovery without acyclicity constraints.
\newblock In \emph{International Conference on Learning Representations}, 2022.
\newblock URL \url{https://openreview.net/forum?id=eYciPrLuUhG}.

\bibitem[Lopez-Paz et~al.(2015)Lopez-Paz, Muandet, Sch{\"o}lkopf, and
  Tolstikhin]{lopezpaz2015towards}
D.~Lopez-Paz, K.~Muandet, B.~Sch{\"o}lkopf, and I.~Tolstikhin.
\newblock Towards a learning theory of cause-effect inference.
\newblock In \emph{Proceedings of the 32nd International Conference on Machine
  Learning}, volume~37 of \emph{Proceedings of Machine Learning Research},
  pages 1452--1461. PMLR, 2015.

\bibitem[Lorch et~al.(2021)Lorch, Rothfuss, Sch{\"o}lkopf, and
  Krause]{lorch2021dibs}
L.~Lorch, J.~Rothfuss, B.~Sch{\"o}lkopf, and A.~Krause.
\newblock {DiBS}: Differentiable {B}ayesian structure learning.
\newblock In \emph{Advances in Neural Information Processing Systems},
  volume~34, 2021.
\newblock URL
  \url{https://proceedings.neurips.cc/paper/2021/hash/ca6ab34959489659f8c3776aaf1f8efd-Abstract.html}.

\bibitem[Lorch et~al.(2022)Lorch, Sussex, Rothfuss, Krause, and
  Sch{\"o}lkopf]{lorch2022avici}
L.~Lorch, S.~Sussex, J.~Rothfuss, A.~Krause, and B.~Sch{\"o}lkopf.
\newblock Amortized inference for causal structure learning.
\newblock In \emph{Advances in Neural Information Processing Systems
  (NeurIPS)}, 2022.

\bibitem[Ma et~al.(2024)Ma, Ding, Fu, Zhang, Wang, Han, and
  Zhang]{ma2024scalable}
P.~Ma, R.~Ding, Q.~Fu, J.~Zhang, S.~Wang, S.~Han, and D.~Zhang.
\newblock Scalable differentiable causal discovery in the presence of latent
  confounders with skeleton posterior.
\newblock In \emph{Proceedings of the 30th ACM SIGKDD Conference on Knowledge
  Discovery and Data Mining}, pages 2141--2152. Association for Computing
  Machinery, 2024.
\newblock \doi{10.1145/3637528.3672031}.
\newblock URL \url{https://dl.acm.org/doi/10.1145/3637528.3672031}.

\bibitem[Mahajan et~al.(2025)Mahajan, Gladrow, Hilmkil, Zhang, and
  Scetbon]{Mahajan2024amortized}
D.~Mahajan, J.~Gladrow, A.~Hilmkil, C.~Zhang, and M.~Scetbon.
\newblock Amortized inference of causal models via conditional fixed-point
  iterations.
\newblock \emph{Transactions on Machine Learning Research}, 2025.
\newblock ISSN 2835-8856.
\newblock URL \url{https://openreview.net/forum?id=D9pq25PGc5}.

\bibitem[Marx and Vreeken(2019)]{marx2019identifiability}
A.~Marx and J.~Vreeken.
\newblock Identifiability of cause and effect using regularized regression.
\newblock In \emph{Proceedings of the 25th ACM SIGKDD International Conference
  on Knowledge Discovery \& Data Mining}, pages 852--861, 2019.

\bibitem[Montagna et~al.(2025)Montagna, Cairney-Leeming, Sridhar, and
  Locatello]{montagna2024demystifying}
F.~Montagna, M.~Cairney-Leeming, D.~Sridhar, and F.~Locatello.
\newblock Demystifying amortized causal discovery with transformers.
\newblock \emph{Transactions on Machine Learning Research}, 2025.
\newblock ISSN 2835-8856.
\newblock URL \url{https://openreview.net/forum?id=9Lgy7IGSfp}.

\bibitem[Mooij et~al.(2016)Mooij, Peters, Janzing, Zscheischler, and
  Sch{\"o}lkopf]{mooij2016tuebingen}
J.~M. Mooij, J.~Peters, D.~Janzing, J.~Zscheischler, and B.~Sch{\"o}lkopf.
\newblock Distinguishing cause from effect using observational data: methods
  and benchmarks.
\newblock \emph{Journal of Machine Learning Research}, 17\penalty0
  (32):\penalty0 1--102, 2016.

\bibitem[Nasr-Esfahany et~al.(2023)Nasr-Esfahany, Alizadeh, and
  Shah]{nasr2023counterfactual}
A.~Nasr-Esfahany, M.~Alizadeh, and D.~Shah.
\newblock Counterfactual identifiability of bijective causal models.
\newblock In \emph{International Conference on Machine Learning}, 2023.

\bibitem[Nazaret and Blei(2024)]{nazaret2025xges}
A.~Nazaret and D.~Blei.
\newblock Extremely greedy equivalence search.
\newblock In \emph{The 40th Conference on Uncertainty in Artificial
  Intelligence}, 2024.
\newblock URL \url{https://openreview.net/forum?id=2gIMX9UxRN}.

\bibitem[Ng et~al.(2020)Ng, Ghassami, and Zhang]{ng2020role}
I.~Ng, A.~Ghassami, and K.~Zhang.
\newblock On the role of sparsity and {DAG} constraints for learning linear
  {DAGs}.
\newblock \emph{Advances in Neural Information Processing Systems (NeurIPS)},
  33:\penalty0 17943--17954, 2020.

\bibitem[Nilforoshan et~al.(2023)Nilforoshan, Moor, Roohani, Chen,
  {\v{S}}urina, Yasunaga, Oblak, and Leskovec]{nilforoshan2023zero}
H.~Nilforoshan, M.~Moor, Y.~Roohani, Y.~Chen, A.~{\v{S}}urina, M.~Yasunaga,
  S.~Oblak, and J.~Leskovec.
\newblock Zero-shot causal learning.
\newblock \emph{Advances in Neural Information Processing Systems},
  36:\penalty0 6862--6901, 2023.

\bibitem[Ogarrio et~al.(2016)Ogarrio, Spirtes, and Ramsey]{ogarrio2016gfci}
J.~M. Ogarrio, P.~Spirtes, and J.~Ramsey.
\newblock A hybrid causal search algorithm for latent variable models.
\newblock In \emph{Proceedings of the Eighth International Conference on
  Probabilistic Graphical Models}, volume~52 of \emph{Proceedings of Machine
  Learning Research}, pages 368--379. PMLR, 2016.
\newblock URL \url{https://proceedings.mlr.press/v52/ogarrio16.html}.

\bibitem[Peng et~al.(2026)Peng, Chen, Tian, Qiao, and Lu]{peng2026causcale}
B.~Peng, S.~Chen, J.~Tian, Y.~Qiao, and C.~Lu.
\newblock {CauScale}: Neural causal discovery at scale.
\newblock In \emph{Proceedings of the 43rd International Conference on Machine
  Learning}, volume 306 of \emph{Proceedings of Machine Learning Research},
  pages 97940--97957. PMLR, 2026.
\newblock URL \url{https://proceedings.mlr.press/v306/peng26c.html}.

\bibitem[Peters and Bühlmann(2014)]{peters2014identifiability}
J.~Peters and P.~Bühlmann.
\newblock Identifiability of {G}aussian structural equation models with equal
  error variances.
\newblock \emph{Biometrika}, 101\penalty0 (1):\penalty0 219--228, 03 2014.
\newblock ISSN 0006-3444.
\newblock \doi{10.1093/biomet/ast043}.
\newblock URL \url{https://doi.org/10.1093/biomet/ast043}.

\bibitem[Qiao et~al.(2026)Qiao, Cai, Li, Chen, Hua, Xu, Chen, Hao, and
  Cui]{qiao2026cdfm}
J.~Qiao, R.~Cai, Z.~Li, W.~Chen, P.~Hua, B.~Xu, Z.~Chen, Z.~Hao, and P.~Cui.
\newblock {CDFM}: Towards a general-purpose causal discovery foundation model.
\newblock \emph{arXiv preprint arXiv:2607.11508}, 2026.

\bibitem[Reisach et~al.(2021)Reisach, Seiler, and Weichwald]{reisach2021beware}
A.~Reisach, C.~Seiler, and S.~Weichwald.
\newblock Beware of the simulated {DAG}! {C}ausal discovery benchmarks may be
  easy to game.
\newblock \emph{Advances in Neural Information Processing Systems},
  34:\penalty0 27772--27784, 2021.

\bibitem[Reuter et~al.(2026)Reuter, Dhir, Diaconu, Robertson, Ossen, Hutter,
  Weller, van~der Wilk, and Sch{\"o}lkopf]{reuter2026use}
A.~Reuter, A.~Dhir, C.~Diaconu, J.~Robertson, O.~Ossen, F.~Hutter, A.~Weller,
  M.~van~der Wilk, and B.~Sch{\"o}lkopf.
\newblock Use what you know: Causal foundation models with partial graphs.
\newblock In \emph{Proceedings of the 43rd International Conference on Machine
  Learning}, volume 306 of \emph{Proceedings of Machine Learning Research},
  pages 104953--104987. PMLR, 2026.
\newblock URL \url{https://proceedings.mlr.press/v306/reuter26a.html}.

\bibitem[Robertson et~al.(2025)Robertson, Reuter, Guo, Hollmann, Hutter, and
  Sch{\"o}lkopf]{robertson2025pfn}
J.~Robertson, A.~Reuter, S.~Guo, N.~Hollmann, F.~Hutter, and B.~Sch{\"o}lkopf.
\newblock {Do-PFN}: In-context learning for causal effect estimation.
\newblock In \emph{Advances in Neural Information Processing Systems},
  volume~38, 2025.
\newblock \doi{10.52202/085713-5814}.

\bibitem[Rolland et~al.(2022)Rolland, Cevher, Kleindessner, Russell, Janzing,
  Sch{\"o}lkopf, and Locatello]{rolland2022score}
P.~Rolland, V.~Cevher, M.~Kleindessner, C.~Russell, D.~Janzing,
  B.~Sch{\"o}lkopf, and F.~Locatello.
\newblock Score matching enables causal discovery of nonlinear additive noise
  models.
\newblock In \emph{International Conference on Machine Learning}, pages
  18741--18753. PMLR, 2022.

\bibitem[Sachs et~al.(2005)Sachs, Perez, Pe'er, Lauffenburger, and
  Nolan]{sachs2005causal}
K.~Sachs, O.~Perez, D.~Pe'er, D.~A. Lauffenburger, and G.~P. Nolan.
\newblock Causal protein-signaling networks derived from multiparameter
  single-cell data.
\newblock \emph{Science}, 308\penalty0 (5721):\penalty0 523--529, 2005.

\bibitem[Scutari(2022)]{bnlearn_discrete_large}
M.~Scutari.
\newblock {B}ayesian network repository.
\newblock \url{https://www.bnlearn.com/bnrepository/}, 2022.
\newblock Accessed: 2026-05-03.

\bibitem[Sharma and K{\i}c{\i}man(2020)]{sharma2020dowhy}
A.~Sharma and E.~K{\i}c{\i}man.
\newblock {DoWhy}: An end-to-end library for causal inference.
\newblock \emph{arXiv preprint arXiv:2011.04216}, 2020.

\bibitem[Shazeer(2020)]{shazeer2020glu}
N.~Shazeer.
\newblock {GLU} variants improve transformer.
\newblock \emph{arXiv preprint arXiv:2002.05202}, 2020.

\bibitem[Shimizu et~al.(2006)Shimizu, Hoyer, Hyv{\"a}rinen, and
  Kerminen]{shimizu2006linear}
S.~Shimizu, P.~O. Hoyer, A.~Hyv{\"a}rinen, and A.~Kerminen.
\newblock A linear non-{G}aussian acyclic model for causal discovery.
\newblock \emph{Journal of Machine Learning Research}, 7:\penalty0 2003--2030,
  2006.

\bibitem[Shimizu et~al.(2011)Shimizu, Inazumi, Sogawa, Hyv{\"a}rinen, Kawahara,
  Washio, Hoyer, and Bollen]{shimizu2011directlingam}
S.~Shimizu, T.~Inazumi, Y.~Sogawa, A.~Hyv{\"a}rinen, Y.~Kawahara, T.~Washio,
  P.~O. Hoyer, and K.~Bollen.
\newblock {DirectLiNGAM}: A direct method for learning a linear non-{G}aussian
  structural equation model.
\newblock \emph{Journal of Machine Learning Research}, 12:\penalty0 1225--1248,
  2011.

\bibitem[Spirtes et~al.(2000)Spirtes, Glymour, and
  Scheines]{spirtes2000causation}
P.~Spirtes, C.~Glymour, and R.~Scheines.
\newblock \emph{Causation, Prediction, and Search}.
\newblock MIT Press, 2nd edition, 2000.

\bibitem[Stein et~al.(2024)Stein, Shadaydeh, and Denzler]{stein2024causal}
G.~Stein, M.~Shadaydeh, and J.~Denzler.
\newblock Embracing the black box: Heading towards foundation models for causal
  discovery from time series data.
\newblock \emph{arXiv preprint arXiv:2402.09305}, 2024.

\bibitem[Sypniewski et~al.(2025)Sypniewski, Olko, Gajewski, and
  Mi{\l}o{\'s}]{cdpfn2025}
M.~Sypniewski, M.~Olko, M.~Gajewski, and P.~Mi{\l}o{\'s}.
\newblock Amortized causal discovery with prior-fitted networks.
\newblock \emph{arXiv preprint arXiv:2512.11840}, 2025.

\bibitem[Thompson et~al.(2026)Thompson, Zhao, Steinberg, and
  Bonilla]{thompson2026arrow}
R.~Thompson, H.~Zhao, D.~M. Steinberg, and E.~V. Bonilla.
\newblock Arrow: A foundation model for causal discovery.
\newblock \emph{arXiv preprint arXiv:2605.07204}, 2026.

\bibitem[Uhler et~al.(2013)Uhler, Raskutti, B{\"u}hlmann, and
  Yu]{uhler2013geometry}
C.~Uhler, G.~Raskutti, P.~B{\"u}hlmann, and B.~Yu.
\newblock Geometry of the faithfulness assumption in causal inference.
\newblock \emph{The Annals of Statistics}, 41\penalty0 (2):\penalty0 436--463,
  2013.
\newblock \doi{10.1214/12-AOS1080}.

\bibitem[Vaswani et~al.(2017)Vaswani, Shazeer, Parmar, Uszkoreit, Jones, Gomez,
  Kaiser, and Polosukhin]{vaswani2017attention}
A.~Vaswani, N.~Shazeer, N.~Parmar, J.~Uszkoreit, L.~Jones, A.~N. Gomez,
  {\L}.~Kaiser, and I.~Polosukhin.
\newblock Attention is all you need.
\newblock \emph{Advances in Neural Information Processing Systems}, 30, 2017.

\bibitem[Wu et~al.(2025)Wu, Bao, Barzilay, and Jaakkola]{wu2024recipe}
M.~Wu, Y.~Bao, R.~Barzilay, and T.~Jaakkola.
\newblock Sample, estimate, aggregate: A recipe for causal discovery foundation
  models.
\newblock \emph{Transactions on Machine Learning Research}, 2025.
\newblock ISSN 2835-8856.
\newblock URL \url{https://openreview.net/forum?id=h434zx5SX0}.

\bibitem[Xu et~al.(2022)Xu, Mian, Marx, and Vreeken]{xu2022inferring}
S.~Xu, O.~A. Mian, A.~Marx, and J.~Vreeken.
\newblock Inferring cause and effect in the presence of heteroscedastic noise.
\newblock In \emph{International Conference on Machine Learning}, pages
  24615--24630. PMLR, 2022.

\bibitem[Xu et~al.(2026)Xu, He, Wang, Zhang, Kuang, Chen, and Cui]{kipt2026}
W.~Xu, Y.~He, Y.~Wang, X.~Zhang, K.~Kuang, Y.~Chen, and P.~Cui.
\newblock A knowledge-informed pretrained model for causal discovery.
\newblock \emph{arXiv preprint arXiv:2603.20842}, 2026.

\bibitem[Yin et~al.(2025)Yin, Gao, and Yu]{yin2025adag}
N.~Yin, T.~Gao, and Y.~Yu.
\newblock Learning causal graphs at scale: A foundation model approach.
\newblock \emph{arXiv preprint arXiv:2506.18285}, 2025.

\bibitem[Yu et~al.(2019)Yu, Chen, Gao, and Yu]{yu2019daggnn}
Y.~Yu, J.~Chen, T.~Gao, and M.~Yu.
\newblock {DAG-GNN}: {DAG} structure learning with graph neural networks.
\newblock In \emph{Proceedings of the 36th International Conference on Machine
  Learning}, volume~97 of \emph{Proceedings of Machine Learning Research},
  pages 7154--7163. PMLR, 2019.
\newblock URL \url{https://proceedings.mlr.press/v97/yu19a.html}.

\bibitem[Zhang and Sennrich(2019)]{zhang2019rmsnorm}
B.~Zhang and R.~Sennrich.
\newblock Root mean square layer normalization.
\newblock In \emph{Advances in Neural Information Processing Systems
  (NeurIPS)}, 2019.

\bibitem[Zhang(2008)]{zhang2008causality}
J.~Zhang.
\newblock On the completeness of orientation rules for causal discovery in the
  presence of latent confounders and selection bias.
\newblock \emph{Artificial Intelligence}, 172\penalty0 (16-17):\penalty0
  1873--1896, 2008.

\bibitem[Zhang and Hyv\"arinen(2009)]{Zhang_UAI}
K.~Zhang and A.~Hyv\"arinen.
\newblock On the identifiability of the post-nonlinear causal model.
\newblock In \emph{Proceedings of the 25th Conference on Uncertainty in
  Artificial Intelligence}, Montreal, Canada, 2009.

\bibitem[Zheng et~al.(2018)Zheng, Aragam, Ravikumar, and Xing]{zheng2018dags}
X.~Zheng, B.~Aragam, P.~Ravikumar, and E.~P. Xing.
\newblock {DAGs} with {NO TEARS}: Continuous optimization for structure
  learning.
\newblock In \emph{Advances in Neural Information Processing Systems
  (NeurIPS)}, 2018.

\bibitem[Zheng et~al.(2024)Zheng, Huang, Chen, Ramsey, Gong, Cai, Shimizu,
  Spirtes, and Zhang]{zheng2024causallearn}
Y.~Zheng, B.~Huang, W.~Chen, J.~Ramsey, M.~Gong, R.~Cai, S.~Shimizu,
  P.~Spirtes, and K.~Zhang.
\newblock Causal-learn: Causal discovery in {Python}.
\newblock \emph{Journal of Machine Learning Research}, 25\penalty0
  (60):\penalty0 1--8, 2024.

\bibitem[Zhu et~al.(2026)Zhu, Mansoldo, Wang, and Groha]{zhu2026tabpfncfm}
M.~Zhu, M.~Mansoldo, C.-H. Wang, and S.~Groha.
\newblock A causal foundation model for structure and outcome prediction.
\newblock \emph{arXiv preprint arXiv:2606.26467}, 2026.

\end{thebibliography}

\appendix

% ===== begin inlined test_time_ablation.tex =====
%\clearpage
%\appendix
%\noindent\rule{\textwidth}{1pt}
%\begin{center}
%\vspace{7pt}
%{\Large  Appendix}
%\end{center}
%\noindent\rule{\textwidth}{1pt}

\section{Evaluation Metrics}\label{sec:eval}

Let $d$ be the number of variables ($D$ in Section~\ref{sec:problem_setup}), let $\bm{D}^* \in \{0,1\}^{d \times d}$ denote the ground-truth directed adjacency matrix, and let
$\hat{\bm D} \in \{0,1\}^{d \times d}$ denote the predicted directed adjacency matrix (in this appendix $\hat{\bm D}$ is the final binary graph, not the edge probabilities $\hat D_{ij}=\sigma(\ell_{ij})$ of Section~\ref{sec:problem_setup}), with
diagonal entries excluded. We evaluate directed-edge recovery using precision, recall, and
$F_1$:
\[
\mathrm{Prec}_{\mathrm{dir}}
=
\frac{\sum_{i \neq j} \hat D_{ij}D^*_{ij}}
{\sum_{i \neq j} \hat D_{ij}},
\qquad
\mathrm{Rec}_{\mathrm{dir}}
=
\frac{\sum_{i \neq j} \hat D_{ij}D^*_{ij}}
{\sum_{i \neq j} D^*_{ij}},
\]
and
\[
F_1
=
\frac{2\,\mathrm{Prec}_{\mathrm{dir}}\mathrm{Rec}_{\mathrm{dir}}}
{\mathrm{Prec}_{\mathrm{dir}}+\mathrm{Rec}_{\mathrm{dir}}}.
\]
Thus, a predicted edge $i \to j$ is counted as correct only when the true DAG also contains
$i \to j$; predicting $j \to i$ is treated as an orientation error.
Some baselines return edges without a single direction: the undirected edges of the CPDAGs of PC, GES, and GRaSP, every edge of FCI's PAG other than $\rightarrow$ and $\circ\!\!\rightarrow$ (which are read as directed toward the arrowhead), and every edge of the naive correlation baseline. Such an edge enters $\hat{\bm D}$ in both directions: on a true edge it gives one correct and one incorrect directed entry, and in the skeleton it counts as one adjacency. The same applies when an amortized model assigns a probability above its threshold to both $i \to j$ and $j \to i$.

For direction-agnostic adjacency recovery, we report skeleton $F_1$. Define the true and
predicted skeletons over unordered pairs by
\[
S^*_{ij} = \mathbbm{1}\{D^*_{ij}=1 \ \text{or}\ D^*_{ji}=1\},
\qquad
\hat S_{ij} = \mathbbm{1}\{\hat D_{ij}=1 \ \text{or}\ \hat D_{ji}=1\},
\qquad i<j.
\]
Skeleton precision and recall are then
\[
\mathrm{Prec}_{\mathrm{skel}}
=
\frac{\sum_{i<j} \hat S_{ij}S^*_{ij}}
{\sum_{i<j} \hat S_{ij}},
\qquad
\mathrm{Rec}_{\mathrm{skel}}
=
\frac{\sum_{i<j} \hat S_{ij}S^*_{ij}}
{\sum_{i<j} S^*_{ij}},
\]
with
\[
\mathrm{Skel}\text{-}F_1
=
\frac{2\,\mathrm{Prec}_{\mathrm{skel}}\mathrm{Rec}_{\mathrm{skel}}}
{\mathrm{Prec}_{\mathrm{skel}}+\mathrm{Rec}_{\mathrm{skel}}}.
\]
This metric measures whether the correct variable pairs are connected, regardless of edge
orientation.

We also report structural Hamming distance normalized by graph size, denoted $\mathrm{SHD}/d$.
SHD counts the number of graph edits required to transform the predicted graph into the true
DAG, where each missing, extra, or reversed edge counts as one edit (a reversal is not counted as a deletion plus an insertion; an edge predicted in both directions also costs one edit, on a true edge as well as between two non-adjacent variables):
\[
\mathrm{SHD}/d = \frac{\mathrm{SHD}(\hat{\bm D},\bm{D}^*)}{d}.
\]
Normalizing by $d$ makes structural errors more comparable across datasets with different
numbers of variables. Lower $\mathrm{SHD}/d$ indicates better graph recovery.

Finally, we report AUROC for directed-edge prediction over the ordered pairs $(i,j)$ with
$i \neq j$. When probability scores are available, AUROC measures the ability of the model to
rank true directed edges above non-edges. When a method returns only a graph (the eleven classical baselines), AUROC is computed from the binary matrix $\hat{\bm D}$ and
corresponds to the ROC performance at the single operating point induced by that
graph:
\[
\mathrm{AUROC}
=
\Pr\!\left(q_{ij} > q_{kl} \mid D^*_{ij}=1,\ D^*_{kl}=0\right)
+
\frac{1}{2}
\Pr\!\left(q_{ij} = q_{kl} \mid D^*_{ij}=1,\ D^*_{kl}=0\right),
\]
where $q_{ij}$ is the predicted edge probability (\cdtname and the amortized baselines) or the binary prediction $\hat D_{ij}$ (classical baselines).
Higher AUROC indicates better discrimination between true directed edges and absent directed
edges.

% ===== begin inlined dataset_descriptions.tex =====
% ============================================================
% Dataset Descriptions — section for NeurIPS paper
% ------------------------------------------------------------
% Covers only the 15 in-distribution evaluation datasets plus
% the Tuebingen cause-effect pairs. Out-of-dimension datasets
% (DREAM4, K562, HEPAR II, etc.) are omitted.
%
% Required preamble:
%   \usepackage{booktabs}
%   \usepackage{array}
% ============================================================

\section{Evaluation Datasets}
\label{sec:datasets}

We evaluate \cdtname on 14 real-world and semi-synthetic causal discovery benchmarks spanning $D \in [8,46]$ variables, which lie within the training range $D \in [2,50]$, as well as on the T{\"u}bingen cause--effect pairs benchmark for bivariate direction.
% In addition, we include four datasets with $D>50$ to assess generalization beyond the training regime.
\notationchange{We train the model on synthetic SCMs, excluding the benchmark samples and ground-truth graphs (Appendix~\ref{sec:training-data}). Table~\ref{tab:datasets} summarizes the benchmark suite.}
\begin{table}[h]
\centering
\small
\setlength{\tabcolsep}{4pt}
\renewcommand{\arraystretch}{1.1}

\begin{tabular}{@{}llrrrrl@{}}
\toprule
\textbf{Dataset} & \textbf{Domain} & $D$ & $N$ & $|E|$ & $\rho$ & \textbf{Type} \\
\midrule
\multicolumn{7}{l}{\emph{Semi-synthetic Bayesian networks (bnlearn repository;~\citealp{bnlearn_discrete_large})}} \\
\textsc{asia}                 & Medical (lung cancer)        &  8 &  1{,}000 &   8 & $0.143$ & Linear Gaussian \\
\textsc{asia\_nonlinear}      & Medical (lung cancer)        &  8 &  1{,}000 &   8 & $0.143$ & Nonlinear variant \\
\textsc{child}                & Medical (congenital heart)   & 20 &  1{,}000 &  25 & $0.066$ & Discrete BN \\
\textsc{child\_nonlinear}     & Medical (congenital heart)   & 20 &  1{,}000 &  25 & $0.066$ & Nonlinear variant \\
\textsc{insurance}            & Insurance risk assessment    & 27 &  1{,}500 &  52 & $0.074$ & Discrete BN \\
\textsc{alarm\_like}          & Medical (ICU monitoring)     & 37 &  2{,}000 &  46 & $0.035$ & Linear Gaussian \\
\textsc{ecoli\_like}          & Gene regulation (E.\ coli)   & 46 &  1{,}000 &  \textcolor{black}{70} & \textcolor{black}{$0.034$} & Linear Gaussian \\
\addlinespace[2pt]
\multicolumn{7}{l}{\emph{Sachs protein signaling~\citep{sachs2005causal}}} \\
\textsc{sachs\_obs}           & Protein signaling (obs.)    & 11 &    853 &  17 & $0.155$ & Flow cytometry \\
\textsc{sachs\_full}          & Protein signaling (mixed)   & 11 &  7{,}466 &  17 & $0.155$ & Flow cytometry \\
\textsc{sachs\_nonlinear}     & Protein signaling           & 11 &  1{,}000 &  17 & $0.155$ & Nonlinear variant \\
\addlinespace[2pt]
\multicolumn{7}{l}{\emph{Real physical / operational systems}} \\
\textsc{causal\_chambers\_lt} & Light tunnel (ETH Z\"urich)  & 20 & 10{,}000 &  39 & $0.103$ & Physical measurements \\
\textsc{petshop\_high\_traffic}  & Microservice telemetry    & 39 &    589 &  42 & $0.028$ & Operational metrics \\
\textsc{petshop\_low\_traffic}   & Microservice telemetry    & 41 &    589 &  43 & $0.026$ & Operational metrics \\
\textsc{petshop\_temporal}       & Microservice (temporal)   & 41 &  1{,}652 &  43 & $0.026$ & Time-series metrics \\
% \midrule
% \multicolumn{7}{l}{\emph{Large Bayesian networks (bnlearn repository;~\citealp{bnlearn_discrete_large}, beyond training range)}} \\
% \textsc{hailfinder}           & Weather forecasting          & 56 &  5{,}000 &  66 & $0.021$ & Discrete BN \\
% \textsc{hepar2}               & Medical (liver disorders)    & 70 &  5{,}000 & 123 & $0.025$ & Discrete BN \\
% \textsc{win95pts}             & Printer troubleshooting      & 76 &  5{,}000 & 112 & $0.020$ & Discrete BN \\
\midrule
\multicolumn{7}{l}{\emph{Bivariate direction benchmark}} \\
\textsc{tuebingen}            & Mixed (102 pairs)             &  2 & variable & 1 & $0.5$   & Real-world pairs \\
\bottomrule
\end{tabular}
\caption{Evaluation benchmarks.
$D$ denotes the number of observed variables, $N$ the number of samples in the dataset, $|E|$ the number of edges in the ground-truth DAG, and $\rho = |E|/(D(D-1))$ the edge density. 
\emph{Type} indicates how the data were generated or measured (BN = Bayesian network). For the \textsc{petshop} tables, missing measurements are kept as missing for \cdtname and replaced by the per-variable mean for the baselines (Appendix~\ref{app:baselines}), and edges point from a called service to its caller.}
\label{tab:datasets}
\end{table}

\paragraph{\notationchange{Semi-Synthetic Bayesian Networks.}}
\textsc{asia}, \textsc{child}, \textsc{insurance}, and \textsc{alarm\_like} are derived from the bnlearn repository~\citep{bnlearn_discrete_large}. 
\textsc{child} and \textsc{insurance} are generated from fixed expert-designed directed acyclic graphs (DAGs) with discrete conditional probability tables (CPTs) via ancestral sampling, while \textsc{asia} and \textsc{alarm\_like} use a linear-Gaussian parameterization of their graphs (Table~\ref{tab:datasets}); in \textsc{asia}, one of the two root variables is drawn from a Student-$t$ distribution with five degrees of freedom instead of a Gaussian. 
The ground-truth graph is therefore known exactly. 
\textsc{asia} is a classic eight-variable lung cancer network; \textsc{child} models congenital heart disease diagnosis; \textsc{insurance} captures insurance risk factors; and \textsc{alarm\_like} is based on an ICU monitoring network.

We additionally evaluate on \textsc{asia\_nonlinear} and \textsc{child\_nonlinear}, which retain the same DAG structures but apply nonlinear mechanisms ($\tanh$, quadratic, or exponential transforms of the parents) with additive Gaussian noise, yielding continuous observations under the same causal graph. 
\textsc{ecoli\_like} is a linear-Gaussian SCM constructed from a published \emph{E.\ coli} gene regulatory network (46 variables, \textcolor{black}{70} edges), providing a biologically motivated benchmark at moderate scale.

\paragraph{Sachs Protein Signaling.}
The Sachs dataset~\citep{sachs2005causal} contains measurements of 11 phosphorylated proteins and phospholipids collected via flow cytometry under multiple experimental conditions. 
\textsc{sachs\_obs} uses the observational subset ($N=853$; general anti-CD3/CD28 stimulation without an intervention on any measured protein), \textsc{sachs\_full} uses the full dataset including interventional samples ($N=7{,}466$, treated as observational during evaluation), and \textsc{sachs\_nonlinear} is a nonlinear simulation on the 17-edge consensus graph reported in the original study. 
All three variants are scored against this graph.

\paragraph{Causal Chambers (Light Tunnel).}
The Causal Chamber light-tunnel system~\citep{gamella2025causal} is a physical experimental setup at ETH Z\"urich consisting of a controllable light source, rotating polarizers, a camera (whose images are not used here), and wavelength-specific intensity sensors. 
Variables correspond to sensor measurements and actuator settings, and the ground-truth causal graph is determined by the known physical interactions among these components. 
The \textsc{causal\_chambers\_lt} dataset contains $N=10{,}000$ randomly selected observations of an experiment with actuator settings drawn uniformly at random (no interventions), restricted to $D=20$ variables; the omitted variables are parentless sensor-parameter actuators, so the restricted graph is the latent projection of the full one. 

\paragraph{PetShop Operational Telemetry.}
The PetShop dataset~\citep{hardt2024petshop} consists of runtime telemetry from a microservice-based e-commerce application. 
Each variable is the average latency of one service; missing measurements are kept as missing for \cdtname (via the observation mask) and replaced by the per-variable mean for the baselines. The ground-truth DAG is the reversed service dependency graph, with an edge from a called service to its caller. 
We evaluate three variants: \textsc{petshop\_high\_traffic} and \textsc{petshop\_low\_traffic}, which reflect steady-state regimes ($D=39$ and $41$, $N=589$), and \textsc{petshop\_temporal}, which includes temporal variability ($D=41$, $N=1{,}652$). The original collection lists two temporal variants whose tables hold the same measurements, with the columns in a different order; we keep one of them. 
These datasets are challenging due to strong correlations induced by service dependencies and non-additive interactions arising from system dynamics such as queuing, retries, and cascading effects.

% \paragraph{Large Bayesian Networks (Dimension Extrapolation).}
% To evaluate generalization beyond the training range $D \in [2,50]$, we consider three expert-designed Bayesian networks from the bnlearn repository~\citep{bnlearn_discrete_large} with $D \in \{56,70,76\}$. 
% \textsc{hailfinder}~\citep{abramson1996hailfinder} is a 56-node, 66-edge network for severe weather forecasting, constructed from meteorological expert knowledge, with relatively sparse connectivity (average degree $2.36$). 
% \textsc{hepar2}~\citep{onisko2003probabilistic} is a 70-node, 123-edge medical diagnostic network for liver disorders, with higher edge density (average degree $3.51$). 
% \textsc{win95pts}~\citep{bnlearn_discrete_large} is a 76-node, 112-edge troubleshooting network for Windows 95 printing issues, exhibiting dense local structure with large Markov blankets and in-degree up to $7$. 
% For each network, we generate $N=5{,}000$ samples via ancestral sampling from the corresponding conditional probability tables, using the published DAG as ground truth. 
% These datasets represent strict zero-shot evaluation for \cdtname, as no graphs with $D>50$ are seen during training.
\paragraph{T{\"u}bingen Cause--Effect Pairs.}
We further evaluate bivariate direction identification using the T{\"u}bingen cause--effect pairs benchmark~\citep{mooij2016tuebingen}, which consists of real-world variable pairs with known causal direction across diverse domains. 
We use the 102 pairs that have a single cause and a single effect variable and a positive evaluation weight, and report both unweighted and weighted accuracy; the benchmark's weights down-weight groups of pairs drawn from the same source data, so that near-duplicate pairs do not dominate the average~\citep{mooij2016tuebingen}. 
This benchmark isolates the directionality task independently of edge existence, providing a complementary evaluation of causal orientation performance.

\section{Inference Protocol} \label{app:IP}
All multivariate datasets are evaluated using a fixed inference pipeline (the T{\"u}bingen protocol is given in the caption of Table~\ref{tab:tuebingen}). 
Given an input dataset, we first apply column sanitization: columns with fewer than $\max(10,\,0.10N)$ observed values and constant columns are set aside and returned as isolated nodes, and an exact duplicate of another column inherits the predictions of the column it copies, with no edge between the two. 
We then perform $K=10$ inference passes, each combining a random permutation of variable indices with a $90\%$ row subsample drawn without replacement (all rows for very small datasets; at most $4{,}500$ rows per pass, which applies to \textsc{sachs\_full} and \textsc{causal\_chambers\_lt}), and average the resulting edge probabilities after mapping predictions back to the original ordering. 
No temperature scaling is applied: a pair $(i,j)$ is predicted as adjacent when $\max(\hat D_{ij},\hat D_{ji})$ exceeds the fixed threshold $0.5$, and it is oriented toward the larger of the two probabilities. 
Finally, self-consistency pruning retains only adjacencies found, in either direction, in at least $50\%$ of the passes. 
\notationchange{We use a single fixed checkpoint across all datasets, with fixed hyperparameters and without fine-tuning.} The \cdtname results in Tables~\ref{tab:method-comparison-avg}--\ref{tab:confounder-eval}, \ref{tab:f1-per-dataset}, and~\ref{tab:per-dataset-transformer}--\ref{tab:confounder-per-dataset} were computed in single precision (float32) on CPU. The protocol of the baselines is given in Appendix~\ref{app:baselines}.

\section{Baseline Protocol}
\label{app:baselines}
\notationchange{We ran every baseline on the 14 benchmark tables, using public implementations and, for the amortized models, the released checkpoints.} All baselines are run on CPU, as is \cdtname (Appendix~\ref{app:IP}), and no method is tuned per dataset. Each run of a classical baseline is limited to two hours; no run on the complete tables reached the limit.

Only the three \textsc{petshop} tables contain missing values ($18{,}410$ cells in total). \cdtname receives them as missing; for every baseline they are replaced by the mean of the observed values of the variable. The classical methods then receive $z$-scored columns. The amortized models receive the imputed table and apply the preprocessing of their own code, which standardizes each variable in all of them except SEA.

\paragraph{\notationchange{Classical Methods.}}
PC ($\alpha\in\{0.01,0.05\}$) and FCI ($\alpha=0.05$) use the Fisher-$z$ test, and GES and GRaSP use the BIC score, all in causal-learn~\citep{zheng2024causallearn}, version~0.1.4.4. DirectLiNGAM and ICA-LiNGAM use the \texttt{lingam} package (1.13.0) with its default settings. NOTEARS-Linear uses the reference implementation of its authors with $\lambda_1=0.1$, and DAGMA uses the \texttt{dagma} package (1.1.1) with $\lambda_1=0.02$; both drop weights below $0.3$ in absolute value. SCORE uses the \texttt{dodiscover} implementation with its default settings. The naive correlation baseline joins two variables by an undirected edge when $|r|>0.3$. A constant column is returned as an isolated node. All eleven methods complete all 14 datasets. Appendix~\ref{sec:eval} states how edges without a direction are scored.

\paragraph{\notationchange{Amortized Methods.}}
\notationchange{Each model uses its released configuration with the row limits specified below.} Unless stated otherwise, the graph contains the edges whose probability exceeds $0.5$, the threshold of the respective authors, and AUROC is computed from the edge probabilities.

\emph{AVICI}~\citep{lorch2022avici}: the \texttt{scm-v0} checkpoint (package version 1.0.7), which its authors recommend for real-valued data. Tables with more than $2{,}000$ rows are subsampled to $2{,}000$ rows to bound memory.

\emph{SEA}~\citep{wu2024recipe}: the released pipeline with its FCI-based checkpoint for observational data and the configuration of the repository ($500$ subsets of $5$ variables with $500$ rows each, FCI with the Fisher-$z$ test at $\alpha=0.05$ on every subset). As in the public code, the input is not rescaled.

\emph{TabCausal}~\citep{li2026tabcausal}: the public inference code with two released checkpoints, reported as two rows: the checkpoint of the code release (TabCausal) and a later checkpoint of the same authors (TabCausal-v2). Both use all rows.

\emph{Arrow}~\citep{thompson2026arrow}: the released checkpoint. Tables with more than $2{,}000$ rows, the largest number of rows in its training data, are subsampled to $2{,}000$ rows. The graph follows the rule of its authors: it contains the pairs with skeleton probability above $0.5$, each oriented by the model's variable-order score, which always gives a DAG. $F_1$, Skel-$F_1$, and SHD are computed from this graph, and AUROC from the model's directed-edge probabilities.

\emph{CauScale}~\citep{peng2026causcale}: the public code with the checkpoint trained on synthetic SCMs and the settings of the authors' inference script, which feeds every row of a table with fewer than $2{,}000$ rows and about $1{,}000$ rows otherwise. CauScale is pretrained on 10 to 500 variables; we run it also on the two \textsc{asia} tables ($D=8$), which are below this range, but not on the cause--effect pairs.

\emph{CDFM}~\citep{qiao2026cdfm}: the public package (version 0.1.0) and checkpoint with default settings. Tables with more than $5{,}000$ rows are subsampled to $5{,}000$ rows. The graph uses the threshold that the package computes for each table from the edge probabilities (between $0.474$ and $0.902$ on the 14 tables). CDFM can also take missing values directly. In Table~\ref{tab:method-comparison-avg} it receives mean-imputed input like the other baselines; with its own handling of missing values, its averages are AUROC $0.860$, $F_1$ $0.456$, Skel-$F_1$ $0.622$, and $\mathrm{SHD}/d$ $1.658$, and the two runs differ only on the three \textsc{petshop} tables.

\paragraph{\notationchange{Cause--Effect Pairs.}}
On the T{\"u}bingen pairs (Table~\ref{tab:tuebingen}), AVICI, TabCausal, Arrow, and CDFM use the same checkpoints and row limits as above ($24$ pairs have more than $2{,}000$ rows and $12$ have more than $5{,}000$). Each pair is given to the model in both column orders and is oriented toward the direction with the larger average edge probability, without a threshold. DirectLiNGAM, ICA-LiNGAM, and SCORE are run as above. RECI is our implementation of Algorithm~1 of \citet{bloebaum2018reci}: both variables are rescaled to $[0,1]$, and the direction with the smaller mean squared error of a degree-3 polynomial regression is taken as causal. HECI uses the code of its authors with default settings; SLOPPY uses the code of its authors with smoothing splines and the AIC score, which their paper states for this benchmark. The classical and bivariate methods are also run in both column orders; a pair on which the two orders give opposite directions counts half (ICA-LiNGAM: $9$ pairs; no other method).
% ===== end inlined dataset_descriptions.tex =====
% ===== end inlined test_time_ablation.tex =====

% ===== begin inlined cdt_v9_architecture_figure.tex =====
% ============================================================
% CDT v9 architecture figures
% ------------------------------------------------------------
% Four separate figures, each full-width:
%   Fig 1 — top-level architecture
%   Fig 2 — (a) encoder block pair
%   Fig 3 — (b) factored edge predictor
%   Fig 4 — (c) confounder module
%
% Required preamble:
%   \usepackage{amsmath, amssymb, bm}
%   \usepackage{tikz}
%   \usetikzlibrary{positioning, arrows.meta, fit, backgrounds, calc, shapes.geometric}
%   \usepackage{xcolor}
%
% No \usepackage{subcaption} or \usepackage{makecell} needed.
% ============================================================

% ── Colour palette ──
\definecolor{cInput}{HTML}{E8F1F8}
\definecolor{cEncoder}{HTML}{DDEEFF}
\definecolor{cStats}{HTML}{FDECD3}
\definecolor{cEdge}{HTML}{D9EAD3}
\definecolor{cTri}{HTML}{FFE5E5}
\definecolor{cConf}{HTML}{E6D9F2}
\definecolor{cOutput}{HTML}{F5F5F5}
\definecolor{cBorder}{HTML}{333333}
\definecolor{cArrow}{HTML}{555555}

% ── Shared styles ──
\tikzset{
  block/.style={draw=cBorder, rounded corners=2pt, minimum width=2.3cm,
                minimum height=0.75cm, align=center, font=\small,
                line width=0.5pt, thick},
  subblock/.style={draw=cBorder, rounded corners=1.5pt, minimum width=1.9cm,
                   minimum height=0.6cm, align=center, font=\footnotesize,
                   line width=0.4pt},
  tensor/.style={draw=cBorder, rectangle, rounded corners=1pt,
                 minimum height=0.55cm, align=center,
                 font=\scriptsize\ttfamily, line width=0.4pt, fill=cInput},
  mod/.style={thick, -{Stealth[length=2mm]}, color=cArrow},
}
% ===== end inlined cdt_v9_architecture_figure.tex =====

% ════════════════════════════════════════════════════════════════

\section{Architecture Details}
\label{app:architecture_details}

\begin{table}[h]
\centering
\small

\begin{tabular}{ll}
\toprule
\textbf{Parameter} & \textbf{Value} \\
\midrule
Hidden dimension $d_h$ & $768$ \\
Number of attention heads $H$ & $8$ \\
Number of encoder layers $L$ & $8$ ($16$ alternating blocks) \\
FFN inner dimension $d_f$ & $\frac{2}{3}\times 4d_h=2048$ \\
Triangular pair dimension $d_{\mathrm{pair}}$ & $192$ \\
Triangular refinement rounds $R$ & $3$ \\
Number of confounder tokens $K_c$ & $8$ \\
Number of global context tokens $K_g$ & $12$ \\
PMA query tokens $K_q$ & $8$ \\
PMA attention heads & $4$ \\
Number of pairwise statistics & $45$ \\
Logit bias initialization & $-2.0$ \\
Number of parameters & $\approx 139$M \\
\bottomrule
\end{tabular}
\caption{Architectural hyperparameters.}
\label{tab:hyperparams}
\end{table}

\noindent\textbf{Normalization and Feedforward Blocks.}
Each encoder block operates on input representations $\bm{H}\in\mathbb{R}^{N\times D\times d_h}$ and outputs tensors of the same shape. 
All blocks use RMSNorm~\citep{zhang2019rmsnorm} applied along the feature dimension:
\[
\RMSNorm(\bm{x})
=
\frac{\bm{x}}
{\sqrt{d_h^{-1}\sum_{r=1}^{d_h}x_r^2+\epsilon}}
\odot \bm{\gamma},
\qquad \epsilon=10^{-6},
\]
where $\bm{x}\in\mathbb{R}^{d_h}$ is a feature vector and $\bm{\gamma}\in\mathbb{R}^{d_h}$ is a learned scale parameter.

The feedforward sublayer uses a SwiGLU activation~\citep{shazeer2020glu}, mapping $\bm{x}\in\mathbb{R}^{d_h}$ to $\mathbb{R}^{d_h}$ via
\[
\mathrm{SwiGLU}(\bm{x})
=
\bm{W}_2
\left[
\SiLU(\bm{W}_g\bm{x})\odot \bm{W}_1\bm{x}
\right],
\]
where $\bm{W}_g,\bm{W}_1\in\mathbb{R}^{d_f\times d_h}$ and $\bm{W}_2\in\mathbb{R}^{d_h\times d_f}$, with inner dimension $d_f$ (Table~\ref{tab:hyperparams}).

Both attention and feedforward outputs are added through residual connections, and output projections are initialized with standard deviation $0.02/\sqrt{2L}$ to stabilize deep stacking of $2L$ blocks. 
No dropout layers are used in the encoder; instead, regularization is provided by continuous data regeneration, feature-level dropout applied to selected pairwise statistics (each of them is dropped with probability $0.2$, for all variable pairs of a batch at once), and two training-only input ablations: in a random $20\%$ of the tasks with more than two variables, all observations of one variable are masked out while the targets are left unchanged (variable dropout), and the embedding difference $\bm{z}_i-\bm{z}_j$ in the input of the feature gate (see below) is zeroed in $40\%$ of the bivariate and $5\%$ of the other tasks (z-ablation) so that the gate must rely on the pairwise statistics.

\noindent\textbf{\notationchange{Multi-Layer Feature Fusion.}}
Let $\bm{H}^{(\ell)}\in\mathbb{R}^{N\times D\times d_h}$ denote the encoder output at layer $\ell$. 
We retain the representations after every fourth block, $\{\bm{H}^{(\ell_k)}\}_{k=1}^4$: the first three are taken after the intermediate RMSNorm and the last is the final encoder output (after the final RMSNorm). We combine them prior to pooling via a learned convex combination:
\[
\bm{H}_{\mathrm{fused}}
=
\sum_{k=1}^{4}
\frac{\exp(\alpha_k)}
{\sum_{k'}\exp(\alpha_{k'})}
\bm{H}^{(\ell_k)},
\]
where $\alpha_k\in\mathbb{R}$ are learnable scalar weights.

The fused representation $\bm{H}_{\mathrm{fused}}\in\mathbb{R}^{N\times D\times d_h}$ is the encoder output $\bm{H}^{\mathrm{enc}}$ of Section~\ref{sec:architecture} and is passed to the pooling module. 
This fusion allows downstream prediction to leverage causal signals that may emerge at different depths of the encoder.

\noindent\textbf{Pairwise Statistics.}
Given input data $\bm{X}\in\mathbb{R}^{N\times D}$, the model computes a tensor of pairwise statistics $\bm{S}\in\mathbb{R}^{D\times D\times d_s}$ with $d_s=45$, where $\bm{s}_{ij}\in\mathbb{R}^{d_s}$ summarizes statistical relationships between variables $i$ and $j$. 
These statistics are computed directly from the raw data as fixed, parameter-free input features; no gradients flow through them. Each variable is first standardized to zero mean and unit variance over its observed entries (after clipping extreme values), and all statistics but four are computed on the standardized values. Three of the four (the variance of local $R^2$ values across bins, the conditional-variance features, and the ratio of standard deviation to range in the reliability metadata) are ratios of the unstandardized values and do not depend on the scale of the variables until fixed numerical floors are reached: for the local-$R^2$ variance of strongly dependent variables when the product of the two standard deviations falls to a few thousandths, for the conditional-variance features only at standard deviations of a few thousandths or below, and for the ratio of standard deviation to range only when the range of a variable falls below $10^{-8}$. The fourth is the conditional-entropy asymmetry (a binned estimate of $H(X_j\mid X_i)-H(X_i\mid X_j)$), which is computed on the unstandardized values and therefore contains the log-ratio of the two marginal standard deviations. Through these statistics, the model inputs are not invariant to the measurement units of the variables.

The feature vector $\bm{s}_{ij}$ is partitioned into four groups: symmetric features (invariant under swapping $i$ and $j$), antisymmetric features (changing sign or direction under swapping), V-structure features capturing collider-like patterns, and reliability metadata describing statistical quality. 
The symmetric features include Pearson and Spearman correlation, partial correlation, nonlinear dependence measures, and higher-order moment interactions. 
The antisymmetric features include residual-variance asymmetry, kernel-based dependence asymmetry (HSIC-RFF), cross-moment asymmetries, regression-based coefficients, conditional-variance features, and directional statistics such as Chatterjee’s $\xi$ and IGCI-style scores. 
The V-structure features capture patterns indicative of collider structures, while the reliability metadata comprise a conditioning proxy (the log-ratio of the largest to the smallest diagonal entry of the inverse of the shrinkage-regularized correlation matrix), the ratio $D/N$ (clipped to $[0,1]$), the ratio of standard deviation to range (the smaller of the two variables' values), and the normalized observation count $\log(1+N)/\log(1+10^{4})$; here $N$ is the mean number of observed values per variable.

These metadata features are always retained during training, allowing the model to learn when other statistics are unreliable, while only the antisymmetric and V-structure features are stochastically dropped for regularization (see below). The rank-based statistics (Spearman correlation, the IGCI-style scores and Chatterjee’s $\xi$) break ties with a small random jitter, so for data with tied values the model output depends on this draw and is permutation-equivariant in distribution rather than exactly.

\noindent\textbf{\notationchange{Statistic-Conditioned Attention Bias.}}
The pairwise statistics $\bm{S}$ are used to construct attention biases for variable-wise self-attention. 
For each layer $\ell$ and attention head $h\in\{1,\ldots,H\}$, a nonlinear projection maps $\bm{s}_{ij}$ to a scalar bias:
\[
b^{(h,\ell)}_{ij}
=
\left[
\bm{W}^{(\ell)}_2
\GELU\!\left(\bm{W}^{(\ell)}_1\bm{s}_{ij}\right)
\right]_h,
\]
where $\bm{W}^{(\ell)}_1\in\mathbb{R}^{d_b\times d_s}$, $\bm{W}^{(\ell)}_2\in\mathbb{R}^{H\times d_b}$, and $d_b=48$ is the hidden dimension of the bias network. 
The resulting biases form a tensor $\bm{B}^{(\ell)}\in\mathbb{R}^{H\times D\times D}$, which is added to the attention logits.

To ensure numerical stability, biases are clamped to the range $[-10,10]$. 
During training, antisymmetric and V-structure features are subjected to inverted dropout, while symmetric features and reliability metadata are always retained. 
This design allows the model to leverage classical dependence cues while learning robustness to noisy or unreliable statistical signals.

\noindent\textbf{\notationchange{Encoder-Conditioned Feature Gate.}}
For each ordered pair $(i,j)$, the direction head constructs a feature gate $\bm{g}_{ij}\in[0,1]^{d_a}$, where $d_a$ is the number of direction-asymmetric statistics (the antisymmetric group and the asymmetric collider-child pair, with paired entries differenced; here $d_a=20$). 
The gate is computed from the variable embeddings $\bm{z}_i,\bm{z}_j\in\mathbb{R}^{d_h}$ (after a LayerNorm that the edge predictor also applies to $\bm{Z}$ before both heads; the confounder module and the role score use the un-normalized $\bm{Z}$) and pairwise statistics $\bm{s}_{ij}\in\mathbb{R}^{d_s}$ as
\[
\bm{g}_{ij}
=
\sigma\!\left(
\bm{W}_2
\GELU\!\left(
\bm{W}_1[\bm{z}_i-\bm{z}_j;\bm{s}_{ij}]
\right)
\right),
\]
where $\bm{W}_1\in\mathbb{R}^{d_g\times(d_h+d_s)}$, $\bm{W}_2\in\mathbb{R}^{d_a\times d_g}$, and $d_g=256$ is the hidden width of the gate (Figure~\ref{fig:cdt-arch}). 
The gate modulates antisymmetric features used for direction prediction, allowing the model to adaptively select informative causal signals for each variable pair.

To prevent saturation, we apply a soft bounds regularization that penalizes gate values outside $[0.1,0.9]$ (Table~\ref{tab:losses}), encouraging flexible feature usage across pairs.

\noindent\textbf{Triangular Refinement.}
Let $\bm{P}\in\mathbb{R}^{D\times D\times d_{\mathrm{pair}}}$ denote the pairwise representation tensor, initialized from the concatenated existence and direction features. 
We refine $\bm{P}$ for $R$ rounds using triangle-based updates that aggregate information over intermediate variables $k\in\{1,\ldots,D\}$. 
Each round consists of outgoing, incoming (fork), and collider updates, followed by row-wise self-attention over pairs.

The outgoing update is given by
\[
\Delta \bm{p}^{\mathrm{out}}_{ij}
=
\frac{1}{\sqrt{D}}
\sum_{k=1}^D
\left(
\frac{1}{2}\tanh(\bm{W}_t \bm{p}_{ik})+\frac{1}{2}
\right)
\odot
\bm{W}_v \bm{p}_{kj},
\]
where $\bm{p}_{ij}\in\mathbb{R}^{d_{\mathrm{pair}}}$, $\bm{W}_t,\bm{W}_v\in\mathbb{R}^{d_{\mathrm{pair}}\times d_{\mathrm{pair}}}$ (biases omitted), and the gate acts channel-wise through the element-wise product $\odot$. The sum runs over the variables that are present; during training, where every task is padded to $D_{\max}=65$ variables (Appendix~\ref{sec:training-data}), the prefactor is $1/\sqrt{D_{\max}}$. 
Incoming and collider updates use analogous gated aggregation with index permutations corresponding to fork and collider motifs. 

Each round first applies a LayerNorm to $\bm{P}$; each update is then added residually, $\bm{p}_{ij}\leftarrow\bm{p}_{ij}+\mathrm{LayerNorm}(\bm{W}_o\,\Delta\bm{p}_{ij})$ with $\bm{W}_o\in\mathbb{R}^{d_{\mathrm{pair}}\times d_{\mathrm{pair}}}$ initialized to zero, and the refined tensor remains in $\mathbb{R}^{D\times D\times d_{\mathrm{pair}}}$. 
The final refined logits are obtained from $\bm{P}$ and combined with the base decoder outputs using learned mixing weights for skeleton and direction components. 
This refinement enables reasoning over higher-order causal structures such as chains, forks, and colliders.

\noindent\textbf{Learned Skeleton Extraction.}
To obtain an undirected skeleton from directed logits, the model learns a smooth function $\mathrm{skel\_fn}:\mathbb{R}^2\to\mathbb{R}$ applied to each pair $(i,j)$:
\[
\mathrm{skel\_fn}(\ell_{ij},\ell_{ji})
=
\bm{w}^{\top}
\bigl[
\ell_{ij}+\ell_{ji},\,
\ell_{ij}\ell_{ji}
\bigr]
+b,
\]
where $\bm{w}\in\mathbb{R}^2$ and $b\in\mathbb{R}$ are learned parameters. 
Inside the decoder, $\mathrm{skel\_fn}$ is applied separately to the base and the refined logits: for $x\in\{\mathrm{base},\mathrm{tri}\}$, the skeleton component is $u^{x}_{ij}=\mathrm{skel\_fn}(\ell^{x}_{ij},\ell^{x}_{ji})$ and the direction component is the antisymmetric part $a^{x}_{ij}=\ell^{x}_{ij}-\ell^{x}_{ji}$. The blended logit is
\[
\tilde\ell_{ij}
=
\psi\,\Bigl[
\sigma(\alpha_e)\,u^{\mathrm{tri}}_{ij}+\bigl(1-\sigma(\alpha_e)\bigr)\,u^{\mathrm{base}}_{ij}
+\sigma(\alpha_d)\,a^{\mathrm{tri}}_{ij}+\bigl(1-\sigma(\alpha_d)\bigr)\,a^{\mathrm{base}}_{ij}
\Bigr],
\]
where $\psi=\mathrm{clamp}\bigl(\sigma(\alpha_s),0.3,0.7\bigr)$ is a learned overall scale and $\alpha_e,\alpha_d,\alpha_s\in\mathbb{R}$ are learned scalars, initialized so that $\sigma(\alpha_e)=0.5$, $\sigma(\alpha_d)\approx 0.7$ and $\psi=0.5$. The role score is then added with the fixed weight $\omega=0.1$ to give the final logits $\ell_{ij}=\tilde\ell_{ij}+\omega r_{ij}$; $\ell^{\mathrm{base}}_{ij}$, $\ell^{\mathrm{tri}}_{ij}$, $\tilde\ell_{ij}$ and $\ell_{ij}$ are each clamped to $\pm 15$. 
$\mathrm{skel\_fn}$ applied to the final logits gives the prediction in $\mathcal{L}_{\mathrm{skel}}$, a differentiable training-time readout that preserves information from both directions; at inference, an edge between $i$ and $j$ is declared (before self-consistency pruning) when the larger of the two averaged directed probabilities exceeds the threshold, and it is oriented toward the larger one.

\noindent\textbf{\notationchange{Confounder Module.}}
Let $\bm{Z}\in\mathbb{R}^{D\times d_h}$ denote the variable embeddings and $\bm{T}\in\mathbb{R}^{K_c\times d_h}$ the learnable confounder tokens. 
The tokens attend to $\bm{Z}$ via two cross-attention layers, producing updated token representations $\bm{T}'\in\mathbb{R}^{K_c\times d_h}$. 
Each layer applies LayerNorm to the tokens, multi-head attention with the normalized tokens as queries and $\bm{Z}$ as keys and values, and a GELU feedforward network, both with residual connections, preserving the token dimensionality.

For each variable $i$ and confounder $k$, a loading network computes $S_{ik}\in(0,1)$, and pairwise confounding probabilities are obtained via the noisy-OR aggregation
\[
C^{\mathrm{OR}}_{ij}=1-\prod_{k=1}^{K_c}(1-S_{ik}S_{jk}).
\]
This computation is performed in float32 for numerical stability, and the resulting probabilities are clamped before conversion to logits.

The loading network bias is initialized to $-2.0$, so that initial loadings satisfy $S_{ik}\approx\sigma(-2)\approx 0.12$, encouraging sparse confounder assignments at the start of training. 
In addition, a pairwise feature projection maps symmetric and gated antisymmetric statistics $\bm{s}_{ij}$ to a scalar bias, which is added to the symmetrized noisy-OR logit before the diagonal and padding entries are masked and a sigmoid is applied. 
Because gated antisymmetric statistics enter this bias, the forward-pass $\hat{\bm{C}}\in[0,1]^{D\times D}$ is not exactly symmetric; the reported confounding matrix is symmetrized at inference, $(\hat{\bm{C}}+\hat{\bm{C}}^\top)/2$.

\noindent\textbf{Acyclicity.}
Let $\bm{\ell}\in\mathbb{R}^{D\times D}$ denote the directed edge logits and $\sigma(\bm{\ell})$ the corresponding edge probabilities. 
An optional acyclicity penalty is defined using the spectral radius $\rho(\sigma(\bm{\ell}))$, estimated via power iteration. 
The penalty takes the form
\[
\mathcal{L}_{\mathrm{acyc}}=\mathrm{ReLU}\bigl(\rho(\sigma(\bm{\ell}))-1\bigr),
\]
which penalizes only spectral radii above $1$. The adjacency matrix of a DAG has $\rho=0$, but a single directed cycle whose edge probabilities are all below $1$ also has $\rho<1$, so the term discourages dense, strongly cyclic graphs rather than excluding cycles.

In the default configuration, this penalty is disabled during training, and acyclicity is not enforced at inference, so the predicted graph may contain cycles. 
Greedy DAG post-processing of the predicted graph (its edges are kept in descending order of probability, and an edge that would close a cycle is dropped) is available as an optional step that returns a valid directed acyclic graph.

\begin{figure}[t]
\centering
\begin{tikzpicture}[
  sl/.style={subblock, fill=white, minimum height=0.62cm, inner sep=2pt},
  addn/.style={circle, draw=cBorder, fill=white, minimum size=0.42cm, inner sep=0pt, line width=0.5pt,
    path picture={\draw[cBorder, line width=0.6pt]
      ([xshift=-1.2mm]path picture bounding box.center) -- ([xshift=1.2mm]path picture bounding box.center)
      ([yshift=-1.2mm]path picture bounding box.center) -- ([yshift=1.2mm]path picture bounding box.center);}},
  bp/.style={circle, fill=cArrow, inner sep=0pt, minimum size=3.6pt},
  swp/.style={subblock, fill=cArrow!20, minimum width=1.75cm, minimum height=0.62cm, inner sep=2pt},
  ten/.style={tensor, font=\footnotesize},
  hdr/.style={anchor=north west, inner sep=0pt, font=\footnotesize, align=left},
  aux/.style={mod, dashed},
  frame/.style={draw=cBorder, fill=cEncoder, rounded corners=3pt, thick},
]
% ---------------- one encoder block as a horizontal chain ----------------
% #1 = name prefix, #2 = stream y
\newcommand{\encchain}[2]{%
  \coordinate (#1in) at (0.45,#2);
  \node[bp] (#1b1) at (0.9,#2) {};
  \node[sl, right=0.38cm of #1b1, minimum width=1.6cm] (#1n1) {RMSNorm};
  \node[sl, right=0.38cm of #1n1, minimum width=1.75cm] (#1qkv) {Q/K/V proj.};
  \node[sl, right=0.38cm of #1qkv, minimum width=1.2cm] (#1att) {SDPA};
  \node[addn, right=0.38cm of #1att] (#1a1) {};
  \node[bp, right=0.38cm of #1a1] (#1b2) {};
  \node[sl, right=0.38cm of #1b2, minimum width=1.6cm] (#1n2) {RMSNorm};
  \node[sl, right=0.38cm of #1n2, minimum width=1.4cm] (#1ffn) {SwiGLU};
  \node[addn, right=0.38cm of #1ffn] (#1a2) {};
  % main path through the sub-layers
  \draw[mod] (#1b1) -- (#1n1);
  \draw[mod] (#1n1) -- (#1qkv);
  \draw[mod] (#1qkv) -- (#1att);
  \draw[mod] (#1att) -- (#1a1);
  \draw[thick, color=cArrow] (#1a1) -- (#1b2);
  \draw[mod] (#1b2) -- (#1n2);
  \draw[mod] (#1n2) -- (#1ffn);
  \draw[mod] (#1ffn) -- (#1a2);
  % residual skips: block input -> around the sub-layer -> add
  \draw[mod, rounded corners=2pt] (#1b1) -- ++(0,0.62) -| (#1a1.north);
  \draw[mod, rounded corners=2pt] (#1b2) -- ++(0,0.62) -| (#1a2.north);
}

% ---------------- input ----------------
\node[ten, anchor=west, minimum width=3.3cm] (hin) at (0.1,1.95)
  {$\bm{H}^{(\ell)} \in \mathbb{R}^{N \times D \times d_h}$};

% ---------------- variable-attention block (even) ----------------
\encchain{v}{0}
\node[swp, align=center] (sw1) at (14.1,0) {\texttt{swapaxes}\\[-1pt]$N \leftrightarrow D$};
\node[subblock, fill=cStats, below=0.55cm of vatt, minimum height=0.55cm, inner sep=2pt] (sb)
  {stat bias from pairwise statistics (clamp $\pm 10$)};
\draw[aux] (sb) -- (vatt);
\node[hdr] at (0.7,1.2)
  {{\small\bfseries Variable-attention block} (even): attends across the $D$ variables $+$ $K_g{=}12$ global tokens};

% ---------------- sample-attention block (odd) ----------------
\encchain{s}{-3.9}
\node[swp, align=center] (sw2) at (14.1,-3.9) {\texttt{swapaxes}\\[-1pt]$N \leftrightarrow D$};
\node[hdr] at (0.7,-2.3)
  {{\small\bfseries Sample-attention block} (odd): attends across the $N$ samples; no stat bias or global tokens;\\
   missing samples are masked as keys (key-padding mask)};

% ---------------- output ----------------
\node[ten, minimum width=1.75cm] (hout) at (14.1,-5.1) {$\bm{H}^{(\ell+1)}$};

% ---------------- inter-block wiring ----------------
\draw[mod] ([xshift=0.35cm]hin.south west) |- (vn1.west);
\draw[mod] (va2) -- (sw1);
\draw[mod, rounded corners=2pt] (sw1.south) |- (6,-1.9) -| (sin) -- (sn1.west);
\draw[mod] (sa2) -- (sw2);
\draw[mod] (sw2) -- (hout);

% ---------------- frames (behind) ----------------
\begin{scope}[on background layer]
  \draw[frame] (0.1,1.3) rectangle (12.95,-1.6);
  \draw[frame] (0.1,-2.2) rectangle (12.95,-4.5);
\end{scope}

% ---------------- legend ----------------
\node[anchor=west, font=\footnotesize, inner sep=0pt] (leg) at (0.1,-5.1) {%
  \tikz[baseline=-0.6ex]\node[addn]{};~residual add (skip connection around each sub-layer)\quad
  \tikz[baseline=-0.6ex]\draw[mod](0,0)--(0.6,0);~data path\quad
  \tikz[baseline=-0.6ex]\draw[aux](0,0)--(0.6,0);~auxiliary input};
\end{tikzpicture}
\caption{\textbf{Encoder Block Pair (one of eight).} The encoder consists of $2L=16$ attention blocks that alternate between variable-axis attention (even-indexed blocks) and sample-axis attention (odd-indexed blocks).
Variable-attention blocks attend across variables and incorporate stat-conditioned attention bias and $K_g=12$ global context tokens, while sample-attention blocks attend across samples and use neither. Each block follows a pre-normalization design with RMSNorm (computed in float32 with $\varepsilon=10^{-6}$), multi-head scaled dot-product attention (SDPA), and a SwiGLU feedforward network.
SDPA is PyTorch's scaled dot-product attention; where the padding mask is merged with the stat bias, masked keys receive $-10^{9}$ instead of $-\infty$ for numerical stability under bfloat16 precision.
The SwiGLU gated product is clamped to $\pm 10^4$ before the output projection. Residual output projections use scaled initialization with standard deviation $0.02/\sqrt{2L}$.
An additional RMSNorm is applied to the residual stream after every fourth block. Embeddings at missing entries are reset to zero after every block.}
\label{fig:cdt-encoder}
\end{figure}

\begin{figure}[t]
\centering
\begin{tikzpicture}[
  inp/.style={tensor, font=\footnotesize, minimum height=0.75cm, inner xsep=5pt},
  head/.style={block, fill=cEdge, minimum height=1.2cm, inner sep=3pt},
  grouptitle/.style={font=\footnotesize\itshape, color=cBorder, anchor=south},
  hlab/.style={font=\footnotesize, text=black, right, xshift=2pt},
  aux/.style={mod, dashed},
]

% ── Symmetric inputs (existence head, left column) ──
\node[inp, minimum width=1.6cm] (zsum) at (0,0) {$\bm{z}_i{+}\bm{z}_j$};
\node[inp, minimum width=1.6cm, right=0.25cm of zsum] (zprod) {$\bm{z}_i{\odot}\bm{z}_j$};
\node[inp, fill=cStats, minimum width=2.6cm, right=0.25cm of zprod] (sym)
  {$\mathrm{proj}_{\mathrm{sym}}(\bm{s}^{\mathrm{sym}})$};
\node[inp, fill=cConf, minimum width=2.6cm, right=0.25cm of sym] (cscore)
  {$\hat{C}_{ij}.\mathtt{detach}()$};

% ── Antisymmetric inputs (direction head, right column) ──
\node[inp, minimum width=1.6cm, right=0.6cm of cscore] (zdiff) {$\bm{z}_i{-}\bm{z}_j$};
\node[inp, fill=cStats, minimum width=3.3cm, right=0.25cm of zdiff] (asym)
  {$\mathrm{proj}_{\mathrm{asym}}(\bm{g}{\odot}\bm{s}^{\mathrm{asym}})$};

% Group titles
\node[grouptitle] at ($(zsum.north west)!0.5!(cscore.north east) + (0,0.08)$) {Existence-head inputs};
\node[grouptitle] at ($(zdiff.north west)!0.5!(asym.north east) + (0,0.08)$) {Direction-head inputs};

% ── Heads: each spans exactly its own inputs ──
\path (zsum.west |- 0,-1.7) coordinate (eL) (cscore.east |- 0,-1.7) coordinate (eR);
\node[head, fit=(eL)(eR), inner sep=0pt, minimum height=1.2cm] (exist) {};
\node[align=center, font=\small] at (exist.center)
  {\textbf{Existence head}\\[1pt]
   {\footnotesize Linear $\to$ GELU $\to \bm{h}^{\mathrm{exist}} \in \mathbb{R}^{d_h}$}};

\path (zdiff.west |- 0,-1.7) coordinate (dL) (asym.east |- 0,-1.7) coordinate (dR);
\node[head, fit=(dL)(dR), inner sep=0pt, minimum height=1.2cm] (dir) {};
\node[align=center, font=\small] at (dir.center)
  {\textbf{Direction head}\\[1pt]
   {\footnotesize Linear $\to$ GELU $\to \bm{h}^{\mathrm{dir}} \in \mathbb{R}^{d_h}$}};

% Inputs -> heads (straight down). Pairwise-statistic inputs dashed.
\draw[mod] (zsum.south)   -- (zsum.south   |- exist.north);
\draw[mod] (zprod.south)  -- (zprod.south  |- exist.north);
\draw[aux] (sym.south)    -- (sym.south    |- exist.north);
\draw[mod] (cscore.south) -- (cscore.south |- exist.north);
\draw[mod] (zdiff.south)  -- (zdiff.south  |- dir.north);
\draw[aux] (asym.south)   -- (asym.south   |- dir.north);

% ── Fusion MLP: full width, below both heads ──
\path (zsum.west |- 0,-3.75) coordinate (fL) (asym.east |- 0,-3.75) coordinate (fR);
\node[head, fit=(fL)(fR), inner sep=0pt, minimum height=1.3cm] (fus) {};
\node[align=center, font=\small] at (fus.center)
  {\textbf{Fusion MLP} (fp32): concat$[\bm{h}^{\mathrm{exist}};\, \bm{h}^{\mathrm{dir}}] \in \mathbb{R}^{2 d_h}$\\[1pt]
   {\footnotesize Linear$(2 d_h \to d_h) \to$ GELU $\to$ Linear$(d_h \to 1)$}};

% Heads -> fusion (straight down, no crossings)
\draw[mod] (exist.south) -- node[hlab] {$\bm{h}^{\mathrm{exist}}$} (exist.south |- fus.north);
\draw[mod] (dir.south)   -- node[hlab] {$\bm{h}^{\mathrm{dir}}$}   (dir.south   |- fus.north);

% ── Output ──
\node[inp, fill=cOutput, minimum width=4.5cm, below=0.55cm of fus] (ell)
  {$\ell^{\mathrm{base}}_{ij} \in \mathbb{R}$ \quad (clamp $\pm 15$)};
\draw[mod] (fus.south) -- (ell.north);

\end{tikzpicture}
\caption{\textbf{Factored Edge Predictor.}
The model predicts each directed edge $(i,j)$ by separating \emph{edge existence} and \emph{edge direction}, using symmetry-aware features. The \emph{existence head} (left) operates on the symmetric inputs $\bm{z}_i+\bm{z}_j$, $\bm{z}_i\odot\bm{z}_j$ and a projection of the symmetric pairwise statistics $\bm{s}^{\mathrm{sym}}$, together with the confounding score $\hat{C}_{ij}$ (detached from gradients). These features capture whether two variables are related, regardless of direction. The \emph{direction head} (right) operates on direction-sensitive inputs: the antisymmetric difference $\bm{z}_i-\bm{z}_j$ and a projection of the encoder-gated antisymmetric pairwise statistics $\bm{g}\odot\bm{s}^{\mathrm{asym}}$. These features capture directional asymmetry between variables. Each head produces a $d_h$-dimensional hidden representation (computed in float32 for numerical stability). These are concatenated into a $2d_h$-dimensional vector and passed to a fusion MLP (bottom), which combines existence and directional evidence to produce a single logit $\ell^{\mathrm{base}}_{ij}$ for the ordered pair $(i,j)$. The logit is clamped to $\pm 15$ and later combined with the triangular-refinement logits; the refinement itself is initialized from the concatenated vector $[\bm{h}^{\mathrm{exist}};\bm{h}^{\mathrm{dir}}]$, not from $\ell^{\mathrm{base}}_{ij}$.
Dashed arrows indicate inputs from pairwise statistics: $\bm{s}^{\mathrm{sym}}$ and $\bm{s}^{\mathrm{asym}}$ are the symmetric and the $d_a{=}20$ antisymmetric statistics, V-structure scores included (Appendix~\ref{app:architecture_details}).}
\label{fig:cdt-edge}
\end{figure}

\begin{figure}[t]
\centering
\begin{tikzpicture}[
  inp/.style={tensor, font=\footnotesize, minimum height=0.95cm, inner sep=3pt},
  cblk/.style={block, fill=cConf, minimum width=5.0cm, minimum height=1.05cm},
  sblk/.style={block, fill=cStats, minimum width=5.0cm, minimum height=1.05cm},
  sumnode/.style={circle, draw=cBorder, fill=white, thick, inner sep=0pt,
                  minimum size=0.62cm, font=\small},
  aux/.style={mod, dashed},
  elab/.style={font=\scriptsize, inner sep=2pt, text=black},
  junc/.style={circle, fill=cArrow, inner sep=0pt, minimum size=3.2pt},
]

% ── Middle column: token path ─────────────────────────────────────
\node[inp, fill=cConf, minimum width=5.0cm] (tok0)
  {Learnable tokens\\$\bm{T} \in \mathbb{R}^{K_c \times d_h}$,\; $K_c{=}8$};

\node[cblk, below=0.55cm of tok0] (ca1)
  {\textbf{Cross-attn layer 1}\\
   {\footnotesize multi-head attn., $Q = \mathrm{LN}(\text{tokens})$}\\
   {\footnotesize $+$ GELU FFN ($2\times$ width)}};

\node[cblk, below=0.45cm of ca1] (ca2)
  {\textbf{Cross-attn layer 2}\\
   {\footnotesize multi-head attn., $Q = \mathrm{LN}(\text{tokens})$}\\
   {\footnotesize $+$ GELU FFN ($2\times$ width)}};

\node[cblk, below=0.45cm of ca2] (load)
  {\textbf{Loading network}\\
   {\footnotesize $S_{ik} = \sigma(\mathrm{LoadNet}([\bm{z}_i;\, \bm{t}'_k]))$}};

\node[cblk, below=0.45cm of load] (nor)
  {\textbf{Noisy-OR}\\
   {\footnotesize $C^{\mathrm{OR}}_{ij} = 1 - \prod_{k=1}^{K_c} (1 - S_{ik} S_{jk})$}\\
   {\footnotesize clamped to $[10^{-3},\, 1-10^{-3}]$}};

% ── Left column: variable embeddings ──────────────────────────────
\node[inp, fill=cInput, minimum width=3.3cm, left=0.8cm of tok0] (Z2)
  {Variable embeddings\\$\bm{Z} \in \mathbb{R}^{D \times d_h}$};

% ── Right column: pairwise-statistics branch and output ───────────
\node[inp, fill=cStats, minimum width=5.0cm, right=0.8cm of tok0] (sfeat)
  {Pairwise statistics\\
   $[\,\bm{s}^{\mathrm{sym}}_{(9)}\;;\; (\bm{g}\!\odot\!\bm{s}^{\mathrm{asym}})_{(20)}\,]$};

\node[sblk] (confstat) at (sfeat.center |- load.center)
  {\textbf{Pairwise feature projection}\\
   {\footnotesize $\mathrm{MLP}(29 \to 48 \to 1)$, bias init $-1$}};

\node[sumnode] (plus) at (sfeat.center |- nor.center) {$+$};

\node[inp, fill=cOutput, minimum width=5.0cm, below=0.55cm of plus] (C)
  {Confounding matrix\\$\hat{\bm{C}} \in [0,1]^{D \times D}$};

% ── Token path (solid) ────────────────────────────────────────────
\draw[mod] (tok0.south) -- (ca1.north);
\draw[mod] (ca1.south) -- (ca2.north);
\draw[mod] (ca2.south) -- (load.north);
\draw[mod] (load.south) -- (nor.north);

% ── Z bus: K,V of both cross-attention layers, z_i of the loading net
\draw[mod] (Z2.south) |- node[elab, pos=0.78, above] {$\bm{z}_i$} (load.west);
\draw[mod] (Z2.south |- ca1.west) -- node[elab, above] {$K, V$} (ca1.west);
\draw[mod] (Z2.south |- ca2.west) -- node[elab, above] {$K, V$} (ca2.west);
\node[junc] at (Z2.south |- ca1.west) {};
\node[junc] at (Z2.south |- ca2.west) {};

% ── Statistics branch (dashed) and logit-space sum ────────────────
\draw[aux] (sfeat.south) -- (confstat.north);
\draw[aux] (confstat.south) -- node[elab, right, xshift=1pt] {logit bias} (plus.north);
\draw[mod] (nor.east) -- node[elab, above] {$\mathrm{logit}\, C^{\mathrm{OR}}_{ij}$} (plus.west);
\draw[mod] (plus.south) -- node[elab, right, xshift=1pt] {$\sigma(\cdot)$} (C.north);

% ── Legend (bottom left) ──────────────────────────────────────────
\coordinate (lg) at ([yshift=0.12cm]Z2.west |- C.center);
\draw[mod] (lg) -- ++(0.7cm,0)
  node[right, font=\footnotesize, text=black] {main data path};
\draw[aux] ([yshift=-0.45cm]lg) -- ++(0.7cm,0)
  node[right, font=\footnotesize, text=black] {pairwise-statistics branch};

\end{tikzpicture}
\caption{\textbf{Confounder Module.}
The model represents latent confounders using $K_c=8$ learnable tokens, which cross-attend to the variable embeddings $\bm{Z}\in\mathbb{R}^{D\times d_h}$ over two layers. Each layer applies LayerNorm (LN) to the tokens, multi-head attention with the normalized tokens as queries, and a GELU feedforward network (with $2\times$ hidden width), both with residual connections; computations are performed in float32 for numerical stability. Token activations are clamped to $\pm 5000$ after each layer. A loading network maps each variable--token pair to a probability $S_{ik}\in(0,1)$, indicating whether confounder $k$ influences variable $i$. Pairwise confounding probabilities $C^{\mathrm{OR}}_{ij}$ are then computed via noisy-OR aggregation, with computation in float32 and outputs clamped away from $0$ and $1$. In parallel, a pairwise feature projection maps symmetric and gated antisymmetric statistics to a scalar logit bias, which is added to the noisy-OR logits. The resulting confounding matrix $\hat{\bm{C}}\in[0,1]^{D\times D}$ is used both as an output and as input to the edge predictor, where $\hat{C}_{ij}$ is detached to prevent edge-prediction gradients from suppressing confounder learning.}
\label{fig:cdt-conf}
\end{figure}
\clearpage
\section{Loss Function Details}
\label{app:loss_table}

As introduced in~\eqref{eq:mainloss}, the training objective combines a primary directed-edge binary cross-entropy loss with a set of auxiliary terms that separately regularize directionality, skeleton recovery, latent confounding, calibration, sparsity, and representation diversity. This decomposition reflects the fact that causal discovery contains several coupled but distinct subproblems: detecting whether two variables are adjacent, orienting the edge correctly, avoiding overconfident reverse predictions, accounting for hidden common causes, and controlling graph density. In particular, $\mathcal{L}_{\mathrm{asym}}$ and $\mathcal{L}_{\mathrm{bdir}}$ provide explicit supervision for edge orientation, $\mathcal{L}_{\mathrm{skel}}$ gives a direction-agnostic signal for adjacency recovery, and $\mathcal{L}_{\mathrm{conf}}$ trains the confounder module to identify latent common causes. The calibration, false-positive, density, PMA-diversity, and gate-regularization terms are intended to stabilize training and to discourage pathological solutions such as saturated logits, overly dense graphs, collapsed pooling queries, or feature gates that over-specialize to synthetic mechanisms.

\begin{table}[th]

\centering
\small

\begin{tabular}{@{}p{0.06\linewidth}p{0.56\linewidth}>{\raggedright\arraybackslash}p{0.32\linewidth}@{}}
\toprule
Term & Definition & Weight / setting \\
\midrule
$\mathcal{L}_{\mathrm{asym}}$ &
$\displaystyle 
\frac{1}{n_+}\sum_{(i,j):D^*_{ij}=1}
\mathrm{ReLU}\!\left(\sigma(\ell_{ji})-\delta\right)V_{ij}$ &
$\lambda_{\mathrm{asym}}=0.08$, $\delta=0.15$ \\[1.0em]

$\mathcal{L}_{\mathrm{cal}}$ &
$\displaystyle
\frac{1}{n_+}\sum_{(i,j):D^*_{ij}=1}
\left[
\mathrm{ReLU}\!\left(|\ell_{ij}-\ell_{ji}|-\Delta_{\max}\right)
\right]^2$ &
$\lambda_{\mathrm{cal}}=0.1$, $\Delta_{\max}=6.0$ \\[1.0em]

$\mathcal{L}_{\mathrm{bdir}}$ &
$\displaystyle
\frac{1}{|\mathcal{A}|}\sum_{(i,j)\in\mathcal{A}}
\mathrm{BCE}(\ell_{ij}-\ell_{ji},1)$, \newline
where 
$\mathcal{A}=\{(i,j):D^*_{ij}=1,\ \ell_{ij}>-3\ \text{or}\ \ell_{ji}>-3\}$ &
$\lambda_{\mathrm{bdir}}=0.25$ \\[1.0em]

$\mathcal{L}_{\mathrm{skel}}$ &
$\displaystyle
\frac{1}{|V_\triangle|}
\sum_{i<j}
w^{\mathrm{skel}}_{ij}V_{ij}
\mathrm{BCE}\!\left(
\mathrm{skel\_fn}(\ell_{ij},\ell_{ji}),
\max(D^*_{ij},D^*_{ji})
\right)$ &
$\lambda_{\mathrm{skel}}=0.5$; positive weight capped at $2.0$ \\[1.0em]

$\mathcal{L}_{\mathrm{conf}}$ &
$\displaystyle
\frac{1}{B}\sum_{b=1}^B
\frac{1}{|V^{(b)}|}
\sum_{ij}
w^{(b)}_{ij}V^{(b)}_{ij}
\mathrm{BCE}\!\left(\mathrm{logit}(\hat C^{(b)}_{ij}),C^{*(b)}_{ij}\right)$ &
$\lambda_{\mathrm{conf}}=0.2$; per-graph positive weight $\sqrt{n_-^{(b)}/n_+^{(b)}}$ (clamped to $[1,15]$), with $n_\pm^{(b)}$ counted on $\bm{C}^*$ \\[1.0em]

$\mathcal{L}_{\mathrm{fp}}$ &
$\displaystyle
\frac{\lambda_{\mathrm{fp}}^{\mathrm{eff}}}{|V|}
\sum_{ij}
\mathrm{ReLU}\!\left(
\sigma(\ell_{ij})(1-D^*_{ij})V_{ij}-0.15
\right)$,\newline
where
$\lambda_{\mathrm{fp}}^{\mathrm{eff}}
=
\lambda_{\mathrm{fp}}\left[1+3\,\mathrm{clamp}(p^*-\hat p_t,-0.3,0.3)\right]$ &
$\lambda_{\mathrm{fp}}=0.15$, $p^*=0.50$ \\[1.0em]

$\mathcal{L}_{\mathrm{den}}$ &
$\displaystyle
\frac{1}{B}\sum_{b=1}^B
\frac{
\mathrm{softplus}\!\left(
\sum_{ij}\sigma(\ell^{(b)}_{ij})V^{(b)}_{ij}
-d_{\mathrm{eff}}^{(b)}\bar e
\right)}
{2d_{\mathrm{eff}}^{(b)}}$ &
$\lambda_{\mathrm{den}}=0.05$; $\bar e=3$ is a fixed target number of edges per variable (not adapted during training) \\[1.0em]

$\mathcal{L}_{\mathrm{pma}}$ &
$\displaystyle
\frac{1}{B D_{\max} K_q^2}
\sum_{b,i}\sum_{a\neq c}
\left[\mathrm{ReLU}\!\left(\cos(\bm o^{(b)}_{ia},\bm o^{(b)}_{ic})-0.5\right)\right]^2$ &
$\lambda_{\mathrm{pma}}=0.01$ \\[1.0em]

$\mathcal{L}_{\mathrm{gate}}$ &
$\displaystyle
\frac{1}{d_a|V|}\sum_{ij}V_{ij}\sum_{c=1}^{d_a}
\Bigl[\mathrm{ReLU}\bigl(0.1-g_{ij,c}\bigr)+\mathrm{ReLU}\bigl(g_{ij,c}-0.9\bigr)\Bigr]$ &
$\lambda_{\mathrm{gate}}=0.05$ (fixed) \\
\bottomrule
\end{tabular}
\caption{Auxiliary loss terms. All sums are masked by valid pairs unless stated otherwise.}
\label{tab:losses}
\end{table}

$\mathcal{L}_{\mathrm{asym}}$ uses a dead zone, so it stops penalizing the reverse direction of a true edge once $\sigma(\ell_{ji})$ falls below $\delta$; $\mathcal{L}_{\mathrm{bdir}}$ is gated on detection, so it acts only on true edges with $\ell_{ij}>-3$ or $\ell_{ji}>-3$, and its gradient shrinks (but does not vanish) as the logit gap $\ell_{ij}-\ell_{ji}$ grows. The skeleton and confounding losses use capped positive-class weights to address class imbalance while preventing rare positive labels from overwhelming shared encoder representations. The false-positive and density penalties control graph sparsity at local and global levels, respectively, while the calibration penalty limits excessive logit gaps that can harm transfer under distribution shift. In $\mathcal{L}_{\mathrm{fp}}$, $p^*$ is a fixed precision target and $\hat p_t=0.95\,\hat p_{t-1}+0.05\,p_t$ is the moving average of the training precision $p_t$ of epoch $t$, computed like $r_t$ (directed entries thresholded at $0.5$, true-positive and false-positive counts summed over all GPUs, replayed hard examples excluded) and updated once per epoch; $\hat r_0=\hat p_0=0.5$. Finally, $\mathcal{L}_{\mathrm{pma}}$, computed on the PMA outputs $\bm o^{(b)}_{ia}$ of query $a$ for variable $i$ of task $b$, encourages the attention-pooling queries to specialize to distinct sample-distribution patterns, and $\mathcal{L}_{\mathrm{gate}}$ keeps the statistic gate from saturating, preserving adaptive use of pairwise causal features across datasets.

In Table~\ref{tab:losses}, $B=3$ is the number of graphs that one GPU processes per step (batch size 24 on 8 GPUs). $\mathcal{L}_{\mathrm{asym}}$, $\mathcal{L}_{\mathrm{cal}}$, $\mathcal{L}_{\mathrm{bdir}}$, $\mathcal{L}_{\mathrm{fp}}$ and $\mathcal{L}_{\mathrm{gate}}$ pool the pairs of these $B$ graphs, so $n_+$, $|V|$ and $|\mathcal{A}|$ in their rows are counts over all $B$ graphs, as are $n_+$ and $n_-$ in $w_+$. In $\mathcal{L}_{\mathrm{skel}}$, $|V_\triangle|=\sum_{i<j}V_{ij}$, and $w^{\mathrm{skel}}_{ij}=\mathrm{clamp}\bigl(\sqrt{n^{\mathrm{skel}}_-/n^{\mathrm{skel}}_+},1,2\bigr)$ for pairs with an edge and $1$ otherwise, where $n^{\mathrm{skel}}_+$ and $n^{\mathrm{skel}}_-$ count the unordered pairs with and without an edge among the $B$ graphs. In $\mathcal{L}_{\mathrm{conf}}$, $w^{(b)}_{ij}$ is the positive weight for pairs with $C^{*(b)}_{ij}=1$ and $1$ otherwise, and $\mathrm{logit}(\hat C^{(b)}_{ij})$ is the confounding logit before the sigmoid. In $\mathcal{L}_{\mathrm{den}}$, $d^{(b)}_{\mathrm{eff}}$ is the number of variables of graph $b$. $\mathcal{L}_{\mathrm{pma}}$ is averaged over all $D_{\max}$ columns of the padded input (Appendix~\ref{sec:training-data}); a padded column adds a constant. In $\mathcal{L}_{\mathrm{gate}}$, $g_{ij,c}$ is component $c$ of the feature gate $\bm{g}_{ij}\in[0,1]^{d_a}$ (Appendix~\ref{app:architecture_details}).

Over the graphs of a batch, $\mathcal{L}_{\mathrm{dir}}$ and $\mathcal{L}_{\mathrm{skel}}$ are weighted means of the per-graph losses, with weight $2$ for replayed hard examples and weight $1$ otherwise; the other terms do not use this weight. Replayed hard examples are selected per GPU by a confidence-weighted error of the directed-edge probabilities, the mean over valid pairs of $|\hat D_{ij}-D^*_{ij}|\,|2\hat D_{ij}-1|$: the $10\%$ of an epoch's tasks with the highest score are kept for up to five epochs, and further tasks of the last epoch in which at least $1\%$ of the pairs are wrong with a probability of at least $0.7$ fill the replay pool up to its capacity.

% ===== begin inlined training_data_section.tex =====
% ============================================================
% Training Data Construction — section for NeurIPS paper
% ------------------------------------------------------------
% Designed to fit on ~1 page (with the subsection on DoWhy).
% Required preamble:
%   \usepackage{amsmath, amssymb, bm}
%   \usepackage{booktabs}
% ============================================================

\section{Synthetic Training Dataset Creation}
\label{sec:training-data}

\notationchange{We train \cdtname using only synthetic structural causal models (SCMs).
The 14 benchmarks reported in Section~\ref{sec:exp} and the T{\"u}bingen
cause--effect pairs are excluded from training.}  The training distribution is
constructed to span the breadth of generative assumptions encountered in
practice, namely, graph topology, functional mechanism, noise family,
dimensionality, sample size, missingness, and confounding, so that the
learned amortized inference transfers zero-shot to real data.

\notationchange{At the start of every epoch, we regenerate the training tasks using a pool
of CPU workers.}  An epoch holds about $15{,}200$ tasks: about $13{,}000$
multivariate tasks, $2{,}000$ bivariate tasks, and $200$ copies of
bivariate tasks whose two columns are shuffled independently and that
are labeled with no edge and no confounding.  About $2{,}200$ (at most $3{,}000$) of
the multivariate tasks are replaced by tasks replayed from recent epochs,
on which the model was confidently wrong about directed edges
(Section~\ref{sec:optimisation}); all other tasks are previously unseen
SCMs.  This continuous regeneration is the main
regularizer; training additionally uses weight decay
(Section~\ref{sec:optimisation}) and the feature-level dropout and input
ablations of Appendix~\ref{app:architecture_details}, while the network
itself has no dropout layers.

\paragraph{Task Definition.}
A training task is a tuple $(\bm{X},\,\bm{M},\,\bm{D}^*,\,\bm{C}^*,\,
\bm{v})$ where
\begin{itemize}[nosep,itemsep=1pt]
  \item $\bm{X} \in \mathbb{R}^{N \times D_{\max}}$ is the observational
        data matrix (padded to $D_{\max}{=}65$);
  \item $\bm{M} \in \{0,1\}^{N \times D_{\max}}$ is the observation
        mask (observed versus missing);
  \item $\bm{D}^* \in \{0,1\}^{D_{\max} \times D_{\max}}$ is the
        ground-truth directed-edge matrix among \emph{observed} variables
        (entries outside the observed sub-block are zero); it is acyclic
        except in subpopulation-mixing tasks (see below);
  \item $\bm{C}^* \in \{0,1\}^{D_{\max} \times D_{\max}}$ is the
        ground-truth confounding matrix (symmetric) indicating pairs
        sharing a hidden common cause: a hidden variable that reaches
        both through hidden-only paths (Appendix~\ref{sec:dowhy-path}), or
        a latent factor introduced by the special task types and
        post-processing steps described below;
  \item $\bm{v} \in \{0,1\}^{D_{\max}}$ is the variable-validity mask.
\end{itemize}
Per task, the number of observed variables $D$ is sampled uniformly from
$[2, 50]$ and the number of samples from $[100, 599]$ (tables are padded
to $600$ rows).  The number of hidden variables is $\mathrm{round}(u D)$
with $u$ uniform on $[0,\, 0.30]$ ($[0,\, 0.50]$ for the dedicated
bivariate tasks, so about half of them have one hidden variable); a task
with $D \ge 4$ that draws no hidden variable receives one with
probability $0.5$.

To improve robustness in low-edge-density regimes, $8\%$ of the multivariate training tasks are generated as ultra-sparse SCMs with $15$--$50$ observed variables, with the edge-density parameter of Appendix~\ref{sec:dowhy-path} drawn from $[0.02,0.04]$ before its cap. In addition, $5\%$ of synthetic tasks are generated as discrete Bayesian networks: conditional probability tables are sampled from Dirichlet distributions, and observations are produced by ancestral sampling from the resulting directed graphical model.

\paragraph{Task Difficulty.}
Each task is generated at a difficulty $\zeta \in [0,1]$; larger values
give more nonlinear mechanisms, more non-Gaussian noise, and more
artifacts.  A multivariate task draws $\zeta$ uniformly from $[0, 0.2]$
with probability $0.25$ and from $[0.2, 1]$ otherwise (mean $0.475$); for
the dedicated bivariate tasks these probabilities are $0.10$ and $0.90$
(mean $0.55$); correlation-fog and ultra-sparse tasks use $\zeta = 1$.
Below, ``full difficulty'' means $\zeta = 1$.  For multivariate tasks drawn with the first of these distributions, the DoWhy path is used with probability
$0.5\,\zeta$ ($24\%$ on average) and missing values are injected with
probability $0.15\,\zeta$ ($7\%$ on average).

\paragraph{Two Generation Paths.}
A task of difficulty $\zeta$ is produced by the DoWhy path described in
Appendix~\ref{sec:dowhy-path} with probability $0.5\,\zeta$ (correlation-fog tasks never are); the other tasks use a custom diverse-graph
generator supporting scale-free (Barab\'asi--Albert), small-world
(Watts--Strogatz), stochastic block model, geometric, star/hub, deep single-root, and biological (hub regulators with feed-forward loops)
topologies.  On this path a non-root variable is linear with probability $0.9 - 0.6\,\zeta$, with random signs and weight magnitudes uniform on $[0.25, w_{\max}]$, $w_{\max}$ uniform on $[0.5, 3]$ (with probability $0.1\,\zeta$ the magnitudes are instead uniform on $[0.15, 0.5]$); otherwise it draws one of 18+ nonlinear mechanism families (among them GP via
random Fourier features, tanh and sigmoid MLPs, Michaelis--Menten, Hill,
piecewise-linear, polynomial, step, interaction, competitive inhibition,
sinusoidal, exponential, logarithmic, log-space ANM, threshold-binary,
categorical), some with heteroscedastic noise inside the mechanism; in $10\%$ of the tasks all non-root variables share one mechanism family instead.  Noise distributions include Gaussian, Laplace,
uniform, Gaussian mixtures, and Cauchy (Table~\ref{tab:noise-models}).  The generator itself injects MCAR / MNAR / MAR missingness (probability $0.15\,\zeta$ per task; Appendix~\ref{sec:missing-data}) and, on the diverse path, multiplicative measurement noise and zero-inflation of single variables (probability $0.15\,\zeta$ and $0.08\,\zeta$ per variable). A separate
post-processing pipeline adds further artifacts: measurement noise, zero-inflation,
discretization, rounding, and log-uniform scale
randomization.  These anti-shortcut perturbations are applied to causes and effects alike,
so that an artifact by itself does not indicate the causal direction. A deliberate exception is the T{\"u}bingen-like augmentation of bivariate tasks described below, which alters the cause or the effect specifically.

At batch assembly time, variables are randomly permuted on 100\% of
training batches (with consistent relabeling of $\bm{D}^*$ and
$\bm{C}^*$), so that no variable-index order produced by the generator can be exploited; marginal-variance ordering (var-sortability;~\citealp{reisach2021beware}) is addressed separately: the encoder inputs are standardized per variable and all pairwise statistics but one are scale-free up to numerical floors (Appendix~\ref{app:architecture_details}).

\paragraph{Noise Models.}
To encourage robustness across heterogeneous data-generating processes, synthetic SCMs are generated with a mixture of additive noise distributions (Table~\ref{tab:noise-models}). Each non-root variable of the diverse path first draws a noise family: Gaussian with weight $0.9 - 0.6\,\zeta$ ($30\%$ at full difficulty), with the remaining probability mass split equally across Laplace, uniform, mixture, and Cauchy noise. For a share $0.35 + 0.3\,\zeta$ of these variables the draw is then replaced by Gaussian noise at one of three scales: near-deterministic ($15\%$, standard deviation in $[0.001, 0.03]$), high signal-to-noise ($10\%$, $[0.03, 0.15]$), or moderate ($10\% + 30\%\,\zeta$, $[0.3, 1.5]$). At full difficulty the additive noise is therefore Gaussian for about $75.5\%$ of these variables and of each other family for about $6.1\%$; for a part of the variables it is then scaled with the signal (heteroscedastic) or replaced by multiplicative log-normal noise. This design exposes the model to light-tailed, heavy-tailed, bounded, multimodal, and pathological noise regimes. In particular, the inclusion of Laplace and Cauchy noise tests robustness to heavy-tailed perturbations, uniform noise tests robustness to bounded-support noise, and mixture noise introduces multimodality that cannot be summarized by low-order moments alone. Cauchy noise is clipped at $50\times$ the nominal standard-deviation scale used for sampling to avoid numerical instabilities while retaining its extreme-tail behavior.

\begin{table}[h]
\centering
\small
\renewcommand{\arraystretch}{1.12}

\begin{tabular}{lcl}
\toprule
\textbf{Noise model} & \textbf{Weight} & \textbf{Properties} \\
\midrule
Gaussian & $30\%$ & Standard additive noise \\
Laplace  & $17.5\%$ & Heavy-tailed noise \\
Uniform  & $17.5\%$ & Bounded-support noise \\
Mixture  & $17.5\%$ & Multimodal noise with $2$--$4$ components \\
Cauchy   & $17.5\%$ & Undefined variance; clipped at $50\times$ nominal scale \\
\bottomrule
\end{tabular}
\caption{\textbf{Noise models used for synthetic SCM generation.}
Weights of the first draw for non-root variables of the diverse path at full difficulty ($\zeta = 1$); the text describes the Gaussian replacement that follows.}
\label{tab:noise-models}
\end{table}

\paragraph{\notationchange{Post-Processing and Special Task Types.}}
After mechanism generation, each synthetic dataset is passed through a stochastic post-processing pipeline designed to mimic common artifacts in real observational data. The pipeline injects AR(1) autocorrelation along the rows of every variable with probability $5\%$, and batch effects (mean shifts on a random subset of variables, each at its own random split point) with probability $3\%$; adds independent noise with probability $5\%$, measurement noise with probability $7.5\%$, and outliers with probability $5\%$ at magnitudes between $3$ and $8$ standard deviations, skipping these continuous perturbations for discrete variables; applies discretization with probability $5\%$ using $5$--$100$ bins; introduces bounded or positive transforms with probability $4\%$; applies monotone post-transforms such as log, square-root, power, sigmoid, and softplus with probability $3\%$; adds zero-inflation or censoring with probability $4\%$; rounds one or two randomly chosen variables to $0$--$3$ decimals with probability $15\%$; randomizes variable scales with probability $10\%$; and introduces extreme scale heterogeneity with probability $2.5\%$ using multiplicative factors in $10^{[-3,3]}$. To model latent dependence and near-collinearity, the generator also introduces a latent noise factor shared by several continuous variables with probability $12\%$, labeling their pairs with absolute correlation $\ge 0.15$ as confounded, and grouped collinearity with probability $5\%$, where one observed source variable is copied, with small noise, onto $3$--$6$ leaf variables (fewer when fewer leaves exist); each copy is labeled as a direct child of the source, its previous edges and confounding entries are removed, and no confounding is added among the copies. Additional augmentations include T{\"u}bingen-like bivariate transformations for $35\%$ of $D=2$ tasks and log-transform augmentation with probability $6\%$ to create positive, skewed, Sachs-like data. Discrete variables are automatically detected and protected from incompatible continuous-noise perturbations. Finally, the synthetic mixture includes several special task types: the $2{,}000$ dedicated bivariate tasks per epoch are drawn with the difficulty distribution given above; log-space tasks comprise $10\%$ and exponentiate all continuous variables (after clipping to $[-8,8]$), with the values of each variable below a low percentile ($5$--$15\%$, one draw per task) set to that percentile, as at a detection limit; correlation-fog tasks comprise $12\%$ and use all-linear mechanisms with very low noise in $[0.001,0.05]$, forcing the model to distinguish direct edges from strong indirect correlation; subpopulation-mixing tasks comprise $5\%$ and combine the rows of two SCMs with the same number of variables, using the union of their edge sets and confounding labels, teaching robustness against heterogeneous populations; a pair that the two SCMs orient in opposite directions receives no directed edge and $\bm{C}^*_{ij}{=}1$, and the union can still contain directed cycles over three or more variables; and intervention-mixture tasks comprise $8\%$ and pool observations across $3$--$6$ experimental conditions, each but the first with a mean shift on one to three variables; the condition index acts as a hidden common cause, so a pair of shifted variables whose absolute correlation in the pooled data is at least $0.15$ is labeled $\bm{C}^*_{ij}{=}1$. The shares of the special task types refer to the multivariate tasks and are nominal: the types are applied one after another to independently drawn tasks, so a later type can replace or further modify a task of an earlier one. In addition, a fraction of base tasks are replaced by dense, transitively collinear linear SCMs with strong edges, and a fraction of diverse-path tasks receive mechanism drift: a latent regime process over the rows modulates the mechanism parameters of a subset of variables, and pairs of modulated variables with detectable pooled dependence are labeled $\bm{C}^*_{ij}{=}1$.

% ------------------------------------------------------------
\subsection{The DoWhy Generator Path}
\label{sec:dowhy-path}
% ------------------------------------------------------------

A training task of difficulty $\zeta$ is generated with probability $0.5\,\zeta$ by a modified version of the random structural causal model (SCM) generator of DoWhy~\citep{sharma2020dowhy,blobaum2024dowhy}, described below; samples are drawn from these SCMs with DoWhy's sampling routine.

\paragraph{Graph Topology.}
The generator constructs a DAG by sequential random attachment: every
non-root node draws $1 + \mathrm{round}(q\,(D_{\mathrm{tot}}-2))$ parents
(at most the number of earlier nodes) uniformly among the earlier nodes,
where $D_{\mathrm{tot}}$ counts observed and hidden variables; edges are
then added so that no root is childless and the graph is connected.  The
root fraction is sampled from $[0.05, 0.40]$.  The density parameter is
$q = \min(q_0,\, k/(D_{\mathrm{tot}}-1))$, with $q_0$ uniform on
$[0.10 - 0.08\,\zeta,\; 0.40]$ and a per-task parent budget
$k = \max(1, \mathrm{round}(a))$, $a$ exponentially distributed with mean
$3$ ($k = 1$ in $39\%$ of the tasks, $k > 4$ in $22\%$).  The cap keeps
most large graphs sparse; the same $q$ is passed to the diverse-graph
generator.

\paragraph{Mechanism Types.}
For each non-root node $Y$, the data-generating process assigns one of three structural mechanisms: a linear model with probability $0.30$, a non-additive-noise MLP with probability $0.35\,\zeta$, and an additive MLP otherwise (the percentages in the list are the values at full difficulty). Let $\bm{X}_{\mathrm{Pa}(Y)}$ denote the parent variables of $Y$.

\begin{itemize}[nosep,itemsep=1pt]
  \item \textit{Linear} (30\%): 
  \[
  Y = \bm{w}^{\top} \bm{X}_{\mathrm{Pa}(Y)} + \varepsilon,
  \]
  where each entry of $\bm{w}$ is drawn independently from a uniform distribution symmetric about zero (not bounded away from zero) and $\varepsilon$ is independent noise.

  \item \textit{Nonlinear additive MLP} (35\%): 
  \[
  Y = f_{\mathrm{NN}}(\bm{X}_{\mathrm{Pa}(Y)}) + \varepsilon,
  \]
  where $f_{\mathrm{NN}}$ is a multilayer perceptron (2--4 layers, 4--64 hidden units per layer) with spectral normalization applied to the weights and a scaling factor to induce strong nonlinearity.

  \item \textit{Non-additive noise MLP} (35\%): 
  \[
  Y = f_{\mathrm{NN}}([\bm{X}_{\mathrm{Pa}(Y)}; \varepsilon]),
  \]
  where the noise $\varepsilon$ is concatenated with the parent variables before being processed by the network, resulting in non-additive noise effects.
\end{itemize}

Of the two additive mechanisms, $20\%$ are heteroscedastic, $Y = f(\bm{X}_{\mathrm{Pa}(Y)}) + (|g(\bm{X}_{\mathrm{Pa}(Y)})| + 0.1)\,\varepsilon$ with $g$ a second random MLP, and $10\%$ of the others act in log space, $Y = s\,\exp(f(\tilde{\bm{X}}_{\mathrm{Pa}(Y)}) + \varepsilon)$, where $\tilde{x} = \mathrm{sign}(x)\log(1+|x|)$ is applied to every parent value, the exponent is clipped to $[-10, 10]$, and $s \in \{-1, +1\}$ is one random sign per variable; $f$ is the linear function or the MLP of the mechanism.

To stabilize the range of generated values, we perform an initial calibration step by drawing $1{,}000$ samples from the SCM and recording per-variable output ranges. 
Subsequently, outputs of neural mechanisms are rescaled to lie within $[-2, 2]$, preventing value explosion as signals propagate through the graph. Each node's output, roots included, may then be clipped at zero or discretized into equal-width bins, with probabilities that grow with task difficulty, and in a small fraction of tasks the non-root mechanisms are either all linear or all neural.

\paragraph{Noise Distributions.}
With probability $1 - 0.5\,\zeta$ the noise of a non-root node is
unimodal: Gaussian, Laplace, uniform, Student-$t$, or Cauchy (clipped at
$50\times$ the noise scale) with weights
$0.25 / 0.25 / 0.15 / 0.15 / 0.20$; otherwise it is a zero-centered
$2$--$6$ component Gaussian mixture.  The noise scale is log-uniform on
$[0.01,\, 0.25]$.  Root variables
additionally sample from log-normal, exponential, $\beta$, and
$\chi^2$ distributions with $65\%$ of root marginals replaced by a
$2$--$6$ component Gaussian mixture (component $\mathrm{std}{=}0.4$,
means in $[-3,3]$) to ensure multimodal root behavior.

\paragraph{Hidden Confounders.}
To induce latent-confounding structure, both generation paths first build an extended
SCM with $D_{\mathrm{tot}} = D + K_H$ variables where $K_H$ equals the requested latent
count; a weighted selection favoring nodes with $\ge 2$ children is
then applied to hide variables.  With probability $0.15\,\zeta$, in tasks
with $D \ge 8$ and $K_H \ge 1$, every hidden variable $h$ receives additional observed children up to a target of
$5$--$10$ children (edges that would close a cycle are skipped), and each
new edge is realized after sampling by adding $w\,h$ to the child, with
$|w|$ uniform on $[0.3, 1.5]$.  The ground-truth DAG $\bm{D}^*$ over
observed variables is computed as the transitive closure through
hidden-only paths (so an observed $i$ causes observed $j$ whenever
there is a directed path $i \to \cdots \to j$ passing only through
hidden nodes), and $\bm{C}^*_{ij}{=}1$ whenever observed $i$ and $j$
share a hidden common ancestor reachable only through hidden nodes; the $\bm{C}^*$ label of a pair is removed when the absolute Pearson correlation of the two variables over their jointly observed rows is below $0.1$ (a pair with fewer than $30$ such rows or with a constant variable keeps its label), while $\bm{D}^*$ is not filtered.
$\bm{D}^*$ and $\bm{C}^*$ are therefore the latent projection of the generating graph onto the observed variables, with this correlation floor applied to $\bm{C}^*$.  For these added edges the graph differs from the sampling process in one respect: the added term $w\,h$ is not propagated to the descendants of the child, whose values were generated before the addition, while their edges are labeled from the graph.

\paragraph{Sanitization.}
Deep nonlinear chains at large $D$ occasionally produce NaN or Inf
values; such entries are replaced with zero (they remain observed in
$\bm{M}$) and each column is clipped around its mean; non-finite values
introduced later by post-processing are zeroed and marked as missing in
$\bm{M}$.  A final guard ensures
$\max(10,\, 0.10N)$ observed samples per variable to prevent degenerate
columns that would cause all-masked softmaxes downstream: a variable
below this number is restored to fully observed in the missingness step
and replaced by a fully observed all-zero column after post-processing.

The main DoWhy settings are summarised in Table~\ref{tab:dowhy-cfg}.

\begin{table}[h]
\centering
\small
\begin{tabular}{ll}
\toprule
\textbf{Parameter} & \textbf{Value / range} \\
\midrule
$|\mathcal{V}_{\mathrm{obs}}|$ (observed variables) & uniform $[2, 50]$ \\
$|\mathcal{V}_{\mathrm{hidden}}|/|\mathcal{V}_{\mathrm{obs}}|$ (latent frac.) & $[0.00, 0.30]$ \\
Root fraction & $[0.05, 0.40]$ \\
Edge density parameter $q$ & $[0.10-0.08\,\zeta,\; 0.40]$, capped at $k/(D_{\mathrm{tot}}-1)$ \\
Number of samples $N$ & uniform $[100, 599]$ \\
Mechanism: linear / tanh-NN / non-additive & $30\%$ / $70\% - 35\%\,\zeta$ / $35\%\,\zeta$ \\
NN hidden layers & $[2, 4]$ \\
NN hidden units/layer & $[4, 64]$ \\
NN weight scale (spectral-normed) & $5.0$ \\
NN output normalization & $[-2, 2]$ \\
Noise std (log-uniform) & $[0.01, 0.25]$ \\
Unimodal noise (probability $1-0.5\,\zeta$) & Gaussian / Laplace / uniform / $t$ / Cauchy \\
\quad weights & $25\% / 25\% / 15\% / 15\% / 20\%$ \\
Noise otherwise & $2$--$6$ component Gaussian mixture \\
Root mixture fraction & $65\%$ ($2$--$6$ Gaussian components) \\
Prob.\ missing data injection & $0.15\,\zeta$ per task (Appendix~\ref{sec:missing-data}) \\
\bottomrule
\end{tabular}
\caption{DoWhy generator configuration; this path is used with probability $0.5\,\zeta$
for a task of difficulty $\zeta$.}
\label{tab:dowhy-cfg}
\end{table}
% ===== end inlined training_data_section.tex =====

\section{Per-Dataset Results}
\label{app:transformer_per_dataset}
Table~\ref{tab:f1-per-dataset} lists the directed-edge $F_1$ of all methods on each of the 14 datasets. \cdtname has the highest $F_1$ on 3 of the 14 datasets (\textsc{asia}, \textsc{child}, \textsc{petshop\_temporal}) and is among the three best methods on 6. It is weakest on \textsc{sachs\_obs} ($0.077$, rank 18 of the 19 rows of Table~\ref{tab:f1-per-dataset}; it finds 9 of the 17 edges but orients only 1 of them correctly), \textsc{child\_nonlinear} ($0.227$, rank 13) and \textsc{petshop\_low\_traffic} ($0.188$, rank 12). Tables~\ref{tab:grasp-per-dataset}, \ref{tab:ges-per-dataset}, and~\ref{tab:pc001-per-dataset} give per-dataset results for GRaSP, GES, and PC ($\alpha=0.01$); Table~\ref{tab:per-dataset-transformer} gives them for \cdtname.
\begin{table}[h]
\centering
\small
\setlength{\tabcolsep}{3pt}
\renewcommand{\arraystretch}{1.12}
\resizebox{\textwidth}{!}{%
\begin{tabular}{l*{14}{c}}
\toprule
\textbf{Method}
& \rotatebox{90}{\textsc{alarm\_like}}
& \rotatebox{90}{\textsc{asia}}
& \rotatebox{90}{\textsc{asia\_nonlinear}}
& \rotatebox{90}{\textsc{causal\_chambers\_lt}}
& \rotatebox{90}{\textsc{child}}
& \rotatebox{90}{\textsc{child\_nonlinear}}
& \rotatebox{90}{\textsc{ecoli\_like}}
& \rotatebox{90}{\textsc{insurance}}
& \rotatebox{90}{\textsc{petshop\_high\_traffic}}
& \rotatebox{90}{\textsc{petshop\_low\_traffic}}
& \rotatebox{90}{\textsc{petshop\_temporal}}
& \rotatebox{90}{\textsc{sachs\_full}}
& \rotatebox{90}{\textsc{sachs\_nonlinear}}
& \rotatebox{90}{\textsc{sachs\_obs}} \\
\midrule
\multicolumn{15}{l}{\emph{Classical baselines}} \\
PC ($\alpha{=}0.01$) & $0.745$ & $0.842$ & $0.571$ & $0.097$ & $0.379$ & $0.125$ & $0.592$ & $0.472$ & $0.427$ & $0.366$ & $0.272$ & $0.439$ & $0.571$ & $0.387$ \\
PC ($\alpha{=}0.05$) & $0.745$ & $0.842$ & $0.571$ & $0.131$ & $0.388$ & $0.094$ & $0.544$ & $0.495$ & $\bm{0.451}$ & $0.358$ & $0.264$ & $0.465$ & $0.579$ & $0.387$ \\
FCI ($\alpha{=}0.05$) & $0.777$ & $0.842$ & $0.571$ & $0.303$ & $0.479$ & $0.254$ & $0.563$ & $0.444$ & $0.436$ & $0.349$ & $0.254$ & $0.400$ & $0.429$ & $0.387$ \\
GES & $0.750$ & $0.842$ & $0.571$ & $0.571$ & $0.540$ & $0.216$ & $\bm{0.815}$ & $0.440$ & $0.420$ & $\bm{0.435}$ & $0.338$ & $0.192$ & $0.550$ & $0.387$ \\
GRaSP & $\bm{0.958}$ & $0.842$ & $0.571$ & $0.600$ & $0.517$ & $0.217$ & $\bm{0.815}$ & $\bm{0.648}$ & $0.333$ & $0.371$ & $0.229$ & $0.355$ & $0.378$ & $0.387$ \\
DirectLiNGAM & $0.287$ & $0.370$ & $0.100$ & $0.494$ & $0.278$ & $0.148$ & $0.258$ & $0.184$ & $0.319$ & $0.256$ & $0.228$ & $0.415$ & $0.429$ & $0.357$ \\
ICA-LiNGAM & $0.224$ & $0.545$ & $0.824$ & $0.382$ & $0.354$ & $0.186$ & $0.449$ & $0.200$ & $0.147$ & $0.137$ & $0.156$ & $0.267$ & $0.350$ & $0.357$ \\
NOTEARS-Linear & $0.533$ & $0.400$ & $0.000$ & $0.038$ & $0.438$ & $0.408$ & $0.541$ & $0.257$ & $0.265$ & $0.246$ & $0.164$ & $0.174$ & $0.571$ & $0.286$ \\
DAGMA (linear) & $0.336$ & $0.250$ & $0.000$ & $0.068$ & $0.571$ & $0.491$ & $0.571$ & $0.312$ & $0.220$ & $0.192$ & $0.173$ & $0.222$ & $0.645$ & $0.273$ \\
SCORE & $0.275$ & $0.211$ & $0.632$ & $0.220$ & $0.576$ & $0.494$ & $0.152$ & $0.200$ & $0.087$ & $0.074$ & $0.104$ & $0.323$ & $\bm{0.944}$ & $0.200$ \\
Naive correlation & $0.221$ & $0.333$ & $0.417$ & $0.296$ & $0.393$ & $0.229$ & $0.173$ & $0.315$ & $0.306$ & $0.286$ & $0.243$ & $0.189$ & $0.488$ & $0.370$ \\
\addlinespace[3pt]
\multicolumn{15}{l}{\emph{Amortized / foundation-model methods}} \\
AVICI & $0.595$ & $0.571$ & $\bm{0.875}$ & $0.000$ & $0.143$ & $0.889$ & $0.589$ & $0.102$ & $0.179$ & $0.133$ & $0.034$ & $0.519$ & $0.813$ & $0.250$ \\
SEA & $0.787$ & $0.714$ & $0.286$ & $0.098$ & $0.314$ & $0.509$ & $0.688$ & $0.247$ & $0.043$ & $0.043$ & $0.025$ & $0.000$ & $0.606$ & $0.000$ \\
TabCausal & $0.841$ & $0.588$ & $\bm{0.875}$ & $0.548$ & $0.077$ & $\bm{0.902}$ & $0.740$ & $0.000$ & $0.253$ & $0.174$ & $0.414$ & $0.364$ & $0.839$ & $0.320$ \\
TabCausal-v2 & $0.771$ & $0.533$ & $0.842$ & $0.587$ & $0.059$ & $0.694$ & $0.661$ & $0.338$ & $0.368$ & $0.293$ & $0.360$ & $\bm{0.529}$ & $0.667$ & $\bm{0.429}$ \\
Arrow & $0.500$ & $0.632$ & $0.533$ & $0.351$ & $0.356$ & $0.542$ & $0.588$ & $0.164$ & $0.115$ & $0.114$ & $0.089$ & $0.148$ & $0.375$ & $0.091$ \\
CauScale & $0.571$ & $0.571$ & $0.400$ & $0.255$ & $0.585$ & $0.549$ & $0.693$ & $0.420$ & $0.000$ & $0.162$ & $0.167$ & $0.200$ & $0.429$ & $0.105$ \\
CDFM & $0.533$ & $0.400$ & $0.632$ & $\bm{0.689}$ & $0.483$ & $0.603$ & $0.584$ & $0.444$ & $0.394$ & $0.354$ & $0.272$ & $0.345$ & $0.500$ & $0.333$ \\
\addlinespace[3pt]
\textbf{\cdtname (ours)} & $0.894$ & $\bm{0.875}$ & $0.667$ & $0.587$ & $\bm{0.694}$ & $0.227$ & $0.694$ & $0.537$ & $0.324$ & $0.188$ & $\bm{0.423}$ & $0.462$ & $0.789$ & $0.077$ \\
\bottomrule
\end{tabular}%
}
\caption{\textbf{Directed-edge $F_1$ of all methods on each of the 14 benchmark datasets.}
The best value per dataset is in bold; tied methods are all in bold. The averages over the datasets are in
Table~\ref{tab:method-comparison-avg}.}
\label{tab:f1-per-dataset}
\end{table}

\begin{table}[h]
\centering
\small
\renewcommand{\arraystretch}{1.12}
\begin{tabular}{lcccccc}
\toprule
\textbf{Dataset} & $F_1 \uparrow$ & SHD $\downarrow$ & Found & Oriented & Extra & Undirected \\
\midrule
\textsc{alarm\_like} & $0.958$ & $4$ & $46/46$ & $46/46$ & $0$ & $4$ \\
\textsc{asia} & $0.842$ & $3$ & $8/8$ & $8/8$ & $0$ & $3$ \\
\textsc{asia\_nonlinear} & $0.571$ & $8$ & $6/8$ & $6/6$ & $3$ & $5$ \\
\textsc{causal\_chambers\_lt} & $0.600$ & $28$ & $21/39$ & $21/21$ & $8$ & $2$ \\
\textsc{child} & $0.517$ & $21$ & $22/25$ & $15/22$ & $10$ & $1$ \\
\textsc{child\_nonlinear} & $0.217$ & $57$ & $20/25$ & $10/20$ & $47$ & $5$ \\
\textsc{ecoli\_like} & $0.815$ & $30$ & $66/70$ & $66/66$ & $1$ & $25$ \\
\textsc{insurance} & $0.648$ & $34$ & $39/52$ & $35/39$ & $10$ & $7$ \\
\textsc{petshop\_high\_traffic} & $0.333$ & $61$ & $27/42$ & $19/27$ & $40$ & $12$ \\
\textsc{petshop\_low\_traffic} & $0.371$ & $48$ & $20/43$ & $18/20$ & $32$ & $13$ \\
\textsc{petshop\_temporal} & $0.229$ & $85$ & $23/43$ & $15/23$ & $64$ & $9$ \\
\textsc{sachs\_full} & $0.355$ & $30$ & $14/17$ & $11/14$ & $25$ & $13$ \\
\textsc{sachs\_nonlinear} & $0.378$ & $17$ & $13/17$ & $7/13$ & $4$ & $3$ \\
\textsc{sachs\_obs} & $0.387$ & $17$ & $8/17$ & $6/8$ & $0$ & $6$ \\
\midrule
Average & $0.516$ & -- & -- & -- & -- & -- \\
\bottomrule
\end{tabular}
\caption{\textbf{Per-dataset results for GRaSP.}
GRaSP has the highest average directed-edge $F_1$ of the classical baselines,
$F_1=0.516$ across the 14 benchmark datasets.
GRaSP, GES, and PC return partially directed graphs, and we score an undirected edge in both directions.
On a true edge it counts as found and as oriented, and its reverse direction as one false positive in $F_1$
and one edit in SHD; between two variables that are not adjacent in the true graph it counts twice in
\emph{Extra} and once in SHD. \emph{Undirected} is the number of undirected edges in the output.}
\label{tab:grasp-per-dataset}
\end{table}

\begin{table}[h]
\centering
\small
\renewcommand{\arraystretch}{1.12}
\begin{tabular}{lcccccc}
\toprule
\textbf{Dataset} & $F_1 \uparrow$ & SHD $\downarrow$ & Found & Oriented & Extra & Undirected \\
\midrule
\textsc{alarm\_like} & $0.750$ & $17$ & $46/46$ & $39/46$ & $8$ & $6$ \\
\textsc{asia} & $0.842$ & $3$ & $8/8$ & $8/8$ & $0$ & $3$ \\
\textsc{asia\_nonlinear} & $0.571$ & $8$ & $6/8$ & $6/6$ & $3$ & $5$ \\
\textsc{causal\_chambers\_lt} & $0.571$ & $29$ & $20/39$ & $20/20$ & $9$ & $3$ \\
\textsc{child} & $0.540$ & $24$ & $22/25$ & $17/22$ & $9$ & $7$ \\
\textsc{child\_nonlinear} & $0.216$ & $59$ & $20/25$ & $11/20$ & $56$ & $13$ \\
\textsc{ecoli\_like} & $0.815$ & $30$ & $66/70$ & $66/66$ & $1$ & $25$ \\
\textsc{insurance} & $0.440$ & $47$ & $31/52$ & $22/31$ & $16$ & $1$ \\
\textsc{petshop\_high\_traffic} & $0.420$ & $55$ & $29/42$ & $25/29$ & $39$ & $19$ \\
\textsc{petshop\_low\_traffic} & $0.435$ & $43$ & $23/43$ & $20/23$ & $23$ & $9$ \\
\textsc{petshop\_temporal} & $0.338$ & $73$ & $25/43$ & $22/25$ & $62$ & $10$ \\
\textsc{sachs\_full} & $0.192$ & $31$ & $13/17$ & $5/13$ & $21$ & $4$ \\
\textsc{sachs\_nonlinear} & $0.550$ & $15$ & $13/17$ & $11/13$ & $5$ & $6$ \\
\textsc{sachs\_obs} & $0.387$ & $17$ & $8/17$ & $6/8$ & $0$ & $6$ \\
\midrule
Average & $0.505$ & -- & -- & -- & -- & -- \\
\bottomrule
\end{tabular}
\caption{\textbf{Per-dataset results for GES.}
GES obtains average directed-edge $F_1=0.505$ across the 14 benchmark datasets.
Columns as in Table~\ref{tab:grasp-per-dataset}.}
\label{tab:ges-per-dataset}
\end{table}

\begin{table}[t]
\centering
\small
\renewcommand{\arraystretch}{1.12}
\begin{tabular}{lcccccc}
\toprule
\textbf{Dataset} & $F_1 \uparrow$ & SHD $\downarrow$ & Found & Oriented & Extra & Undirected \\
\midrule
\textsc{alarm\_like} & $0.745$ & $17$ & $42/46$ & $35/42$ & $1$ & $5$ \\
\textsc{asia} & $0.842$ & $3$ & $8/8$ & $8/8$ & $0$ & $3$ \\
\textsc{asia\_nonlinear} & $0.571$ & $8$ & $6/8$ & $6/6$ & $3$ & $5$ \\
\textsc{causal\_chambers\_lt} & $0.097$ & $44$ & $11/39$ & $3/11$ & $10$ & $6$ \\
\textsc{child} & $0.379$ & $25$ & $21/25$ & $11/21$ & $4$ & $9$ \\
\textsc{child\_nonlinear} & $0.125$ & $38$ & $15/25$ & $4/15$ & $23$ & $8$ \\
\textsc{ecoli\_like} & $0.592$ & $56$ & $51/70$ & $45/51$ & $2$ & $29$ \\
\textsc{insurance} & $0.472$ & $40$ & $28/52$ & $21/28$ & $6$ & $3$ \\
\textsc{petshop\_high\_traffic} & $0.427$ & $44$ & $22/42$ & $19/22$ & $18$ & $11$ \\
\textsc{petshop\_low\_traffic} & $0.366$ & $46$ & $17/43$ & $15/17$ & $16$ & $10$ \\
\textsc{petshop\_temporal} & $0.272$ & $64$ & $18/43$ & $14/18$ & $36$ & $13$ \\
\textsc{sachs\_full} & $0.439$ & $21$ & $11/17$ & $9/11$ & $13$ & $0$ \\
\textsc{sachs\_nonlinear} & $0.571$ & $13$ & $11/17$ & $10/11$ & $3$ & $5$ \\
\textsc{sachs\_obs} & $0.387$ & $17$ & $8/17$ & $6/8$ & $0$ & $6$ \\
\midrule
Average & $0.449$ & -- & -- & -- & -- & -- \\
\bottomrule
\end{tabular}
\caption{\textbf{Per-dataset results for PC with $\alpha=0.01$.}
PC obtains average directed-edge $F_1=0.449$ across the 14 benchmark datasets.
Columns as in Table~\ref{tab:grasp-per-dataset}.}
\label{tab:pc001-per-dataset}
\end{table}

\begin{table}[h]
\centering
\scriptsize
\renewcommand{\arraystretch}{1.12}
\resizebox{\textwidth}{!}{%
\begin{tabular}{lrrrrrrrrrr}
\toprule
\textbf{Dataset}
& $D$
& $N$
& $F_1 \uparrow$
& Skel-$F_1 \uparrow$
& SHD $\downarrow$
& SHD$/d \downarrow$
& Found
& Oriented
& Extra
& AUROC $\uparrow$ \\
\midrule
\textsc{alarm\_like}
& $37$ & $2000$ & $0.894$ & $0.936$ & $8$ & $0.216$ & $44/46$ & $42/44$ & $4$ & $0.998$ \\
\textsc{asia}
& $8$ & $1000$ & $0.875$ & $1.000$ & $1$ & $0.125$ & $8/8$ & $7/8$ & $0$ & $0.984$ \\
\textsc{asia\_nonlinear}
& $8$ & $1000$ & $0.667$ & $0.889$ & $4$ & $0.500$ & $8/8$ & $6/8$ & $2$ & $0.958$ \\
\textsc{causal\_chambers\_lt}
& $20$ & $10000$ & $0.587$ & $0.640$ & $29$ & $1.450$ & $24/39$ & $22/24$ & $12$ & $0.874$ \\
\textsc{child}
& $20$ & $1000$ & $0.694$ & $0.898$ & $10$ & $0.500$ & $22/25$ & $17/22$ & $2$ & $0.969$ \\
\textsc{child\_nonlinear}
& $20$ & $1000$ & $0.227$ & $0.500$ & $28$ & $1.400$ & $11/25$ & $5/11$ & $8$ & $0.895$ \\
\textsc{ecoli\_like}
& $46$ & $1000$ & $0.694$ & $0.790$ & $32$ & $0.696$ & $49/70$ & $43/49$ & $5$ & $0.974$ \\
\textsc{insurance}
& $27$ & $1500$ & $0.537$ & $0.683$ & $32$ & $1.185$ & $28/52$ & $22/28$ & $2$ & $0.874$ \\
\textsc{petshop\_high\_traffic}
& $39$ & $589$ & $0.324$ & $0.559$ & $38$ & $0.974$ & $19/42$ & $11/19$ & $7$ & $0.849$ \\
\textsc{petshop\_low\_traffic}
& $41$ & $589$ & $0.188$ & $0.375$ & $46$ & $1.122$ & $12/43$ & $6/12$ & $9$ & $0.757$ \\
\textsc{petshop\_temporal}
& $41$ & $1652$ & $0.423$ & $0.535$ & $37$ & $0.902$ & $19/43$ & $15/19$ & $9$ & $0.819$ \\
\textsc{sachs\_full}
& $11$ & $7466$ & $0.462$ & $0.538$ & $13$ & $1.182$ & $7/17$ & $6/7$ & $2$ & $0.821$ \\
\textsc{sachs\_nonlinear}
& $11$ & $1000$ & $0.789$ & $0.895$ & $6$ & $0.545$ & $17/17$ & $15/17$ & $4$ & $0.978$ \\
\textsc{sachs\_obs}
& $11$ & $853$ & $0.077$ & $0.692$ & $16$ & $1.455$ & $9/17$ & $1/9$ & $0$ & $0.607$ \\
\midrule
Average
& -- & -- & $0.531$ & $0.709$ & $21.4$ & $0.875$ & -- & -- & -- & $0.883$ \\
\bottomrule
\end{tabular}%
}
\caption{\textbf{Per-dataset results for \cdtname on the 14 benchmark datasets.}
We report directed-edge $F_1$, skeleton $F_1$, structural Hamming distance
(SHD), normalized SHD, number of true edges found, number of found edges
oriented correctly, number of extra predicted edges, and directed-edge AUROC.
The final row reports macro-averages across datasets where applicable.}
\label{tab:per-dataset-transformer}
\end{table}

\clearpage
\section{Missing Data}
\label{sec:missing-data}

A practical causal discovery method must handle incomplete observations, which are ubiquitous in real-world data (e.g., missing medical records, failed sensors, detection limits). \cdtname receives the observation mask as part of its input (Section~\ref{sec:architecture}), so it takes a table with missing values as it is, without imputation.
This appendix evaluates \cdtname and the baselines of Table~\ref{tab:method-comparison-avg} on incomplete copies of the 14 benchmark datasets.
The copies are made with two mechanisms, purely random and selection-driven missingness, at two levels each.

\paragraph{\notationchange{Missingness Generation.}}
For each dataset with variables $\mathcal{V}$ and each level, we draw a \emph{driver} set $\mathcal{S}$ of $\lfloor D/2 \rfloor$ variables uniformly at random among the variables without missing values; all other variables form the \emph{target} set $\mathcal{T} = \mathcal{V} \setminus \mathcal{S}$.
Driver variables stay fully observed, and only cells of target variables are removed.
The level, $10\%$ or $30\%$, is the expected share of cells removed from each target variable.
At a given level, both mechanisms use the same driver and target sets; the sets are drawn separately for the two levels.

\emph{Purely random missingness} (missing completely at random, MCAR): each cell of a target variable is removed independently with probability $10\%$ or $30\%$.

\emph{Selection-driven missingness} (missing at random, MAR): whether a cell of a target variable is removed depends on the driver values of the same sample.
For each target variable $j \in \mathcal{T}$ and sample $n$, the observation mask is drawn as
\[
m_{nj} \sim \mathrm{Bernoulli}\!\left(1 - \sigma\!\left(\alpha_j + \bm{\beta}_j^{\top} \bm{X}^{\mathrm{std}}_{n,\mathcal{S}}\right)\right),
\]
where $\bm{X}^{\mathrm{std}}_{n,\mathcal{S}}$ denotes the standardized driver values (zero mean, unit variance), $\bm{\beta}_j \sim \mathcal{N}(\bm{0}, \bm{I})$ is drawn independently for each target variable, and $\alpha_j$ is set so that the expected missing rate of variable $j$ equals the level.
Missingness thus depends only on fully observed variables.

Cells that are already missing in the three \textsc{petshop} tables stay missing in every copy.
We draw one realization of the missing cells per dataset, mechanism, and level.
Since about half of the variables are targets, the $10\%$ and $30\%$ levels remove about $5\%$ and $15\%$ of all cells of a complete table: on the 11 datasets without earlier gaps, $5.2\%$ and $15.5\%$ of the cells are missing on average, under both mechanisms.
With the three \textsc{petshop} tables and their earlier gaps included, the average over the 14 datasets is $8.1\%$ at the $10\%$ level and $17.7\%$ (purely random) or $17.8\%$ (selection-driven) at the $30\%$ level.
For comparison, a training task of difficulty $\zeta$ receives missing values with probability $0.15\,\zeta$ (Appendix~\ref{sec:training-data}). Each such task draws a rate $r$ uniformly from $[2\%, 12\%]$. In half of these tasks every cell is removed independently with probability $r$; in $30\%$ a detection limit (missing not at random, MNAR) removes the lowest $5$--$15\%$ of the values of about half of the variables; in the remaining $20\%$ about $30\%$ of the variables lose each cell with probability $3r\,\sigma(x - \mathrm{med}(x))$, where $x$ is the value of another variable in the same row and $\mathrm{med}(x)$ the median of that variable.

\cdtname receives each incomplete table with the missing cells marked as missing.
CDFM, the only baseline that handles missing values itself (Table~\ref{tab:fm-comparison}), is run twice: with its own handling (\emph{native}) and on the mean-imputed table.
Every other baseline receives the table with each missing value replaced by the mean of the observed values of its variable, as on the clean tables.
All other settings are those of the clean evaluation (Appendices~\ref{app:IP} and~\ref{app:baselines}).
\notationchange{We did not evaluate the baselines with multiple imputation or missingness-aware independence tests.}

\paragraph{\notationchange{MCAR Results.}}
Tables~\ref{tab:missing-mcar-10} and~\ref{tab:missing-mcar-30} report the averages over the 14 datasets and, in parentheses, the change against the same method on the clean tables.
At the $10\%$ level, \cdtname has the best average on all four metrics; its lead in $F_1$ is small ($0.502$ versus $0.495$ for TabCausal-v2), and its $F_1$ is lower by $0.029$ than on the clean tables.
At the $30\%$ level, it keeps the best AUROC ($0.853$) and $\mathrm{SHD}/d$ ($1.065$).
Its $F_1$ falls to $0.434$, behind GRaSP ($0.458$), CDFM with native handling ($0.447$), and GES ($0.445$), and its Skel-$F_1$ falls to $0.594$, behind seven rows, of which GES is the best ($0.615$).

\begin{table}[h]
\centering
\small
\setlength{\tabcolsep}{4pt}
\renewcommand{\arraystretch}{1.15}
\begin{tabular}{@{}lccccc@{}}
\toprule
\textbf{Method}
 & \textbf{Cov.}
 & \textbf{AUROC $\uparrow$ ($\Delta$)}
 & \textbf{$F_1$ $\uparrow$ ($\Delta$)}
 & \textbf{Skel-$F_1$ $\uparrow$ ($\Delta$)}
 & \textbf{SHD$/d$ $\downarrow$ ($\Delta$)} \\
\midrule
\multicolumn{6}{l}{\emph{Classical baselines (mean imputation)}} \\
PC ($\alpha{=}0.01$)
 & $14/14$ & $0.683\,(-0.030)$ & $0.389\,(-0.060)$ & $0.626\,(-0.039)$ & $1.450\,(+0.141)$ \\
PC ($\alpha{=}0.05$)
 & $14/14$ & $0.694\,(-0.030)$ & $0.393\,(-0.058)$ & $0.621\,(-0.039)$ & $1.554\,(+0.160)$ \\
FCI ($\alpha{=}0.05$)
 & $14/14$ & $0.721\,(-0.021)$ & $0.418\,(-0.045)$ & $0.625\,(-0.039)$ & $1.582\,(+0.119)$ \\
GES
 & $14/14$ & $0.726\,(-0.034)$ & $0.433\,(-0.072)$ & $0.621\,(-0.065)$ & $1.651\,(+0.237)$ \\
GRaSP
 & $14/14$ & $0.722\,(-0.040)$ & $0.419\,(-0.097)$ & $0.606\,(-0.086)$ & $1.640\,(+0.260)$ \\
DirectLiNGAM
 & $14/14$ & $0.630\,(-0.022)$ & $0.274\,(-0.021)$ & $0.524\,(-0.009)$ & $2.538\,(+0.085)$ \\
ICA-LiNGAM
 & $14/14$ & $0.674\,(+0.008)$ & $0.316\,(-0.011)$ & $0.510\,(-0.044)$ & $2.548\,(+0.148)$ \\
NOTEARS-Linear
 & $14/14$ & $0.590\,(-0.018)$ & $0.277\,(-0.032)$ & $0.517\,(-0.015)$ & $1.212\,(+0.022)$ \\
DAGMA (linear)
 & $14/14$ & $0.589\,(-0.026)$ & $0.254\,(-0.055)$ & $0.542\,(-0.058)$ & $1.343\,(+0.075)$ \\
SCORE
 & $14/14$ & $0.643\,(-0.029)$ & $0.275\,(-0.046)$ & $0.519\,(-0.046)$ & $3.309\,(+0.262)$ \\
Naive correlation
 & $14/14$ & $0.686\,(+0.004)$ & $0.311\,(+0.007)$ & $0.468\,(+0.007)$ & $2.588\,(-0.077)$ \\
\addlinespace[3pt]
\multicolumn{6}{l}{\emph{Amortized / foundation-model methods (mean imputation, except CDFM (native))}} \\
AVICI
 & $14/14$ & $0.781\,(-0.023)$ & $0.365\,(-0.042)$ & $0.503\,(-0.029)$ & $1.102\,(+0.099)$ \\
SEA
 & $14/14$ & $0.784\,(-0.013)$ & $0.295\,(-0.016)$ & $0.431\,(-0.023)$ & $1.746\,(-0.009)$ \\
TabCausal
 & $14/14$ & $0.819\,(-0.019)$ & $0.468\,(-0.027)$ & $0.605\,(-0.012)$ & $1.096\,(+0.140)$ \\
TabCausal-v2
 & $14/14$ & $0.818\,(-0.013)$ & $0.495\,(-0.014)$ & $0.646\,(-0.050)$ & $1.117\,(+0.133)$ \\
Arrow
 & $14/14$ & $0.738\,(-0.006)$ & $0.318\,(-0.010)$ & $0.544\,(-0.005)$ & $1.429\,(+0.080)$ \\
CauScale
 & $14/14$ & $0.771\,(-0.029)$ & $0.280\,(-0.085)$ & $0.446\,(-0.078)$ & $1.309\,(+0.118)$ \\
CDFM (mean-imputed)
 & $14/14$ & $0.841\,(-0.015)$ & $0.454\,(-0.015)$ & $0.599\,(-0.032)$ & $1.645\,(+0.064)$ \\
CDFM (native)
 & $14/14$ & $0.853\,(-0.007)$ & $0.462\,(+0.006)$ & $0.617\,(-0.005)$ & $1.646\,(-0.012)$ \\
\addlinespace[3pt]
\multicolumn{6}{l}{\emph{No imputation (missing values kept as missing)}} \\
\textbf{\cdtname (ours)}
 & $14/14$ & $\bm{0.870}\,(-0.013)$ & $\bm{0.502}\,(-0.029)$ & $\bm{0.685}\,(-0.024)$ & $\bm{0.921}\,(+0.046)$ \\
\bottomrule
\end{tabular}
\caption{Purely random (MCAR) missingness at the $10\%$ level: averages over the 14 benchmark datasets.
\emph{Cov.} is the number of datasets a method completed.
The value in parentheses is the change against the same method on the clean tables (Table~\ref{tab:method-comparison-avg}; for CDFM (native), against its run with native handling on the clean tables, Appendix~\ref{app:baselines}).
Bold marks the best average in each column.}
\label{tab:missing-mcar-10}
\end{table}

\begin{table}[h]
\centering
\small
\setlength{\tabcolsep}{4pt}
\renewcommand{\arraystretch}{1.15}
\begin{tabular}{@{}lccccc@{}}
\toprule
\textbf{Method}
 & \textbf{Cov.}
 & \textbf{AUROC $\uparrow$ ($\Delta$)}
 & \textbf{$F_1$ $\uparrow$ ($\Delta$)}
 & \textbf{Skel-$F_1$ $\uparrow$ ($\Delta$)}
 & \textbf{SHD$/d$ $\downarrow$ ($\Delta$)} \\
\midrule
\multicolumn{6}{l}{\emph{Classical baselines (mean imputation)}} \\
PC ($\alpha{=}0.01$)
 & $14/14$ & $0.678\,(-0.035)$ & $0.375\,(-0.074)$ & $0.598\,(-0.067)$ & $1.552\,(+0.243)$ \\
PC ($\alpha{=}0.05$)
 & $14/14$ & $0.685\,(-0.039)$ & $0.384\,(-0.067)$ & $0.603\,(-0.057)$ & $1.613\,(+0.219)$ \\
FCI ($\alpha{=}0.05$)
 & $14/14$ & $0.715\,(-0.027)$ & $0.402\,(-0.061)$ & $0.606\,(-0.058)$ & $1.692\,(+0.229)$ \\
GES
 & $14/14$ & $0.730\,(-0.030)$ & $0.445\,(-0.060)$ & $\bm{0.615}\,(-0.071)$ & $1.573\,(+0.159)$ \\
GRaSP
 & $14/14$ & $0.745\,(-0.017)$ & $\bm{0.458}\,(-0.058)$ & $0.602\,(-0.090)$ & $1.621\,(+0.241)$ \\
DirectLiNGAM
 & $14/14$ & $0.609\,(-0.043)$ & $0.245\,(-0.050)$ & $0.503\,(-0.030)$ & $2.521\,(+0.068)$ \\
ICA-LiNGAM
 & $14/14$ & $0.672\,(+0.006)$ & $0.327\,(0.000)$ & $0.523\,(-0.031)$ & $2.314\,(-0.086)$ \\
NOTEARS-Linear
 & $14/14$ & $0.584\,(-0.024)$ & $0.256\,(-0.053)$ & $0.468\,(-0.064)$ & $1.245\,(+0.055)$ \\
DAGMA (linear)
 & $14/14$ & $0.564\,(-0.051)$ & $0.210\,(-0.099)$ & $0.528\,(-0.072)$ & $1.372\,(+0.104)$ \\
SCORE
 & $14/14$ & $0.619\,(-0.053)$ & $0.240\,(-0.081)$ & $0.493\,(-0.072)$ & $3.477\,(+0.430)$ \\
Naive correlation
 & $14/14$ & $0.681\,(-0.001)$ & $0.314\,(+0.010)$ & $0.466\,(+0.005)$ & $2.408\,(-0.257)$ \\
\addlinespace[3pt]
\multicolumn{6}{l}{\emph{Amortized / foundation-model methods (mean imputation, except CDFM (native))}} \\
AVICI
 & $14/14$ & $0.678\,(-0.126)$ & $0.283\,(-0.124)$ & $0.473\,(-0.059)$ & $1.282\,(+0.279)$ \\
SEA
 & $14/14$ & $0.773\,(-0.024)$ & $0.258\,(-0.053)$ & $0.400\,(-0.054)$ & $1.756\,(+0.001)$ \\
TabCausal
 & $14/14$ & $0.743\,(-0.095)$ & $0.336\,(-0.159)$ & $0.495\,(-0.122)$ & $1.369\,(+0.413)$ \\
TabCausal-v2
 & $14/14$ & $0.720\,(-0.111)$ & $0.340\,(-0.169)$ & $0.556\,(-0.140)$ & $1.480\,(+0.496)$ \\
Arrow
 & $14/14$ & $0.704\,(-0.040)$ & $0.301\,(-0.027)$ & $0.489\,(-0.060)$ & $1.475\,(+0.126)$ \\
CauScale
 & $14/14$ & $0.762\,(-0.038)$ & $0.273\,(-0.092)$ & $0.460\,(-0.064)$ & $1.599\,(+0.408)$ \\
CDFM (mean-imputed)
 & $14/14$ & $0.777\,(-0.079)$ & $0.402\,(-0.067)$ & $0.595\,(-0.036)$ & $1.722\,(+0.141)$ \\
CDFM (native)
 & $14/14$ & $0.841\,(-0.019)$ & $0.447\,(-0.009)$ & $0.606\,(-0.016)$ & $1.633\,(-0.025)$ \\
\addlinespace[3pt]
\multicolumn{6}{l}{\emph{No imputation (missing values kept as missing)}} \\
\textbf{\cdtname (ours)}
 & $14/14$ & $\bm{0.853}\,(-0.030)$ & $0.434\,(-0.097)$ & $0.594\,(-0.115)$ & $\bm{1.065}\,(+0.190)$ \\
\bottomrule
\end{tabular}
\caption{Purely random (MCAR) missingness at the $30\%$ level: averages over the 14 benchmark datasets, with the change against the same method on the clean tables in parentheses, as in Table~\ref{tab:missing-mcar-10}.}
\label{tab:missing-mcar-30}
\end{table}
% ------------------------------------------------------------

\begin{table}[h]
\centering
\small
\setlength{\tabcolsep}{4pt}
\renewcommand{\arraystretch}{1.15}
\begin{tabular}{@{}lccccc@{}}
\toprule
\textbf{Method}
 & \textbf{Cov.}
 & \textbf{AUROC $\uparrow$ ($\Delta$)}
 & \textbf{$F_1$ $\uparrow$ ($\Delta$)}
 & \textbf{Skel-$F_1$ $\uparrow$ ($\Delta$)}
 & \textbf{SHD$/d$ $\downarrow$ ($\Delta$)} \\
\midrule
\multicolumn{6}{l}{\emph{Classical baselines (mean imputation)}} \\
PC ($\alpha{=}0.01$)
 & $14/14$ & $0.652\,(-0.061)$ & $0.342\,(-0.107)$ & $0.596\,(-0.069)$ & $1.554\,(+0.245)$ \\
PC ($\alpha{=}0.05$)
 & $14/14$ & $0.653\,(-0.071)$ & $0.332\,(-0.119)$ & $0.580\,(-0.080)$ & $1.713\,(+0.319)$ \\
FCI ($\alpha{=}0.05$)
 & $14/14$ & $0.698\,(-0.044)$ & $0.374\,(-0.089)$ & $0.586\,(-0.078)$ & $1.751\,(+0.288)$ \\
GES
 & $14/14$ & $0.692\,(-0.068)$ & $0.380\,(-0.125)$ & $0.585\,(-0.101)$ & $1.794\,(+0.380)$ \\
GRaSP
 & $14/14$ & $0.690\,(-0.072)$ & $0.367\,(-0.149)$ & $0.562\,(-0.130)$ & $1.883\,(+0.503)$ \\
DirectLiNGAM
 & $14/14$ & $0.632\,(-0.020)$ & $0.265\,(-0.030)$ & $0.493\,(-0.040)$ & $2.741\,(+0.288)$ \\
ICA-LiNGAM
 & $14/14$ & $0.647\,(-0.019)$ & $0.285\,(-0.042)$ & $0.508\,(-0.046)$ & $2.664\,(+0.264)$ \\
NOTEARS-Linear
 & $14/14$ & $0.589\,(-0.019)$ & $0.262\,(-0.047)$ & $0.504\,(-0.028)$ & $1.215\,(+0.025)$ \\
DAGMA (linear)
 & $14/14$ & $0.598\,(-0.017)$ & $0.277\,(-0.032)$ & $0.550\,(-0.050)$ & $1.300\,(+0.032)$ \\
SCORE
 & $14/14$ & $0.607\,(-0.065)$ & $0.229\,(-0.092)$ & $0.443\,(-0.122)$ & $4.256\,(+1.209)$ \\
Naive correlation
 & $14/14$ & $0.683\,(+0.001)$ & $0.309\,(+0.005)$ & $0.464\,(+0.003)$ & $2.470\,(-0.195)$ \\
\addlinespace[3pt]
\multicolumn{6}{l}{\emph{Amortized / foundation-model methods (mean imputation, except CDFM (native))}} \\
AVICI
 & $14/14$ & $0.770\,(-0.034)$ & $0.382\,(-0.025)$ & $0.517\,(-0.015)$ & $1.105\,(+0.102)$ \\
SEA
 & $14/14$ & $0.781\,(-0.016)$ & $0.295\,(-0.016)$ & $0.416\,(-0.038)$ & $1.730\,(-0.025)$ \\
TabCausal
 & $14/14$ & $0.803\,(-0.035)$ & $0.442\,(-0.053)$ & $0.586\,(-0.031)$ & $1.112\,(+0.156)$ \\
TabCausal-v2
 & $14/14$ & $0.816\,(-0.015)$ & $\bm{0.513}\,(+0.004)$ & $\bm{0.638}\,(-0.058)$ & $1.075\,(+0.091)$ \\
Arrow
 & $14/14$ & $0.734\,(-0.010)$ & $0.334\,(+0.006)$ & $0.520\,(-0.029)$ & $1.452\,(+0.103)$ \\
CauScale
 & $14/14$ & $0.786\,(-0.014)$ & $0.284\,(-0.081)$ & $0.436\,(-0.088)$ & $1.464\,(+0.273)$ \\
CDFM (mean-imputed)
 & $14/14$ & $0.827\,(-0.029)$ & $0.447\,(-0.022)$ & $0.601\,(-0.030)$ & $1.718\,(+0.137)$ \\
CDFM (native)
 & $14/14$ & $0.848\,(-0.012)$ & $0.464\,(+0.008)$ & $0.605\,(-0.017)$ & $1.689\,(+0.031)$ \\
\addlinespace[3pt]
\multicolumn{6}{l}{\emph{No imputation (missing values kept as missing)}} \\
\textbf{\cdtname (ours)}
 & $14/14$ & $\bm{0.863}\,(-0.020)$ & $0.481\,(-0.050)$ & $0.629\,(-0.080)$ & $\bm{1.005}\,(+0.130)$ \\
\bottomrule
\end{tabular}
\caption{Selection-driven (MAR) missingness at the $10\%$ level: averages over the 14 benchmark datasets, with the change against the same method on the clean tables in parentheses, as in Table~\ref{tab:missing-mcar-10}.}
\label{tab:missing-10}
\end{table}

\begin{table}[h]
\centering
\small
\setlength{\tabcolsep}{4pt}
\renewcommand{\arraystretch}{1.15}
\begin{tabular}{@{}lccccc@{}}
\toprule
\textbf{Method}
 & \textbf{Cov.}
 & \textbf{AUROC $\uparrow$ ($\Delta$)}
 & \textbf{$F_1$ $\uparrow$ ($\Delta$)}
 & \textbf{Skel-$F_1$ $\uparrow$ ($\Delta$)}
 & \textbf{SHD$/d$ $\downarrow$ ($\Delta$)} \\
\midrule
\multicolumn{6}{l}{\emph{Classical baselines (mean imputation)}} \\
PC ($\alpha{=}0.01$)
 & $14/14$ & $0.668\,(-0.045)$ & $0.365\,(-0.084)$ & $0.587\,(-0.078)$ & $1.522\,(+0.213)$ \\
PC ($\alpha{=}0.05$)
 & $14/14$ & $0.660\,(-0.064)$ & $0.341\,(-0.110)$ & $0.575\,(-0.085)$ & $1.693\,(+0.299)$ \\
FCI ($\alpha{=}0.05$)
 & $14/14$ & $0.702\,(-0.040)$ & $0.377\,(-0.086)$ & $0.577\,(-0.087)$ & $1.770\,(+0.307)$ \\
GES
 & $13/14^{\dagger}$ & $0.681\,(-0.063)$ & $0.380\,(-0.101)$ & $0.592\,(-0.073)$ & $1.700\,(+0.228)$ \\
GRaSP
 & $14/14$ & $0.681\,(-0.081)$ & $0.364\,(-0.152)$ & $0.555\,(-0.137)$ & $1.854\,(+0.474)$ \\
DirectLiNGAM
 & $14/14$ & $0.608\,(-0.044)$ & $0.241\,(-0.054)$ & $0.467\,(-0.066)$ & $2.804\,(+0.351)$ \\
ICA-LiNGAM
 & $14/14$ & $0.663\,(-0.003)$ & $0.290\,(-0.037)$ & $0.463\,(-0.091)$ & $2.759\,(+0.359)$ \\
NOTEARS-Linear
 & $14/14$ & $0.574\,(-0.034)$ & $0.226\,(-0.083)$ & $0.433\,(-0.099)$ & $\bm{1.276}\,(+0.086)$ \\
DAGMA (linear)
 & $14/14$ & $0.586\,(-0.029)$ & $0.250\,(-0.059)$ & $0.493\,(-0.107)$ & $1.343\,(+0.075)$ \\
SCORE
 & $14/14$ & $0.591\,(-0.081)$ & $0.199\,(-0.122)$ & $0.391\,(-0.174)$ & $4.672\,(+1.625)$ \\
Naive correlation
 & $14/14$ & $0.678\,(-0.004)$ & $0.309\,(+0.005)$ & $0.458\,(-0.003)$ & $2.293\,(-0.372)$ \\
\addlinespace[3pt]
\multicolumn{6}{l}{\emph{Amortized / foundation-model methods (mean imputation, except CDFM (native))}} \\
AVICI
 & $14/14$ & $0.681\,(-0.123)$ & $0.241\,(-0.166)$ & $0.421\,(-0.111)$ & $1.326\,(+0.323)$ \\
SEA
 & $14/14$ & $0.763\,(-0.034)$ & $0.261\,(-0.050)$ & $0.383\,(-0.071)$ & $1.859\,(+0.104)$ \\
TabCausal
 & $14/14$ & $0.738\,(-0.100)$ & $0.313\,(-0.182)$ & $0.504\,(-0.113)$ & $1.376\,(+0.420)$ \\
TabCausal-v2
 & $14/14$ & $0.715\,(-0.116)$ & $0.364\,(-0.145)$ & $0.551\,(-0.145)$ & $1.378\,(+0.394)$ \\
Arrow
 & $14/14$ & $0.690\,(-0.054)$ & $0.273\,(-0.055)$ & $0.463\,(-0.086)$ & $1.512\,(+0.163)$ \\
CauScale
 & $14/14$ & $0.736\,(-0.064)$ & $0.257\,(-0.108)$ & $0.403\,(-0.121)$ & $1.607\,(+0.416)$ \\
CDFM (mean-imputed)
 & $14/14$ & $0.766\,(-0.090)$ & $0.379\,(-0.090)$ & $0.533\,(-0.098)$ & $1.969\,(+0.388)$ \\
CDFM (native)
 & $14/14$ & $\bm{0.848}\,(-0.012)$ & $\bm{0.462}\,(+0.006)$ & $\bm{0.614}\,(-0.008)$ & $1.558\,(-0.100)$ \\
\addlinespace[3pt]
\multicolumn{6}{l}{\emph{No imputation (missing values kept as missing)}} \\
\textbf{\cdtname (ours)}
 & $14/14$ & $0.797\,(-0.086)$ & $0.341\,(-0.190)$ & $0.472\,(-0.237)$ & $1.295\,(+0.420)$ \\
% dagger: GES did not complete ecoli_like (timeout: stopped at the time limit of 7200 s). Its averages are over the other 13 datasets, and its changes are against its averages over the same 13 datasets of the clean tables.
\bottomrule
\end{tabular}

\vspace{2pt}
{\footnotesize\raggedright $^{\dagger}$GES did not complete \textsc{ecoli\_like} within the limit of two hours per run. Its averages are over the other 13 datasets, and its changes are against its averages over the same 13 datasets of the clean tables.\par}
\caption{Selection-driven (MAR) missingness at the $30\%$ level: averages over the 14 benchmark datasets (GES: 13 datasets, see $^{\dagger}$), with the change against the same method on the clean tables in parentheses, as in Table~\ref{tab:missing-mcar-10}.}
\label{tab:missing-30}
\end{table}

\clearpage
\paragraph{\notationchange{MAR Results.}}
Tables~\ref{tab:missing-10} and~\ref{tab:missing-30} report the averages in the same way.
At both levels, \cdtname changes more on every metric than under purely random missingness.
At the $10\%$ level, it keeps the best AUROC ($0.863$) and $\mathrm{SHD}/d$ ($1.005$) and is second in $F_1$ ($0.481$) and Skel-$F_1$ ($0.629$), behind TabCausal-v2 ($0.513$ and $0.638$).
At the $30\%$ level, its $F_1$ falls by $0.190$ to $0.341$ and its Skel-$F_1$ by $0.237$ to $0.472$, the largest changes of all rows on these two metrics.
Its $F_1$ is ninth of the 20 rows (at the printed precision, it is level with PC at $\alpha{=}0.05$), and its Skel-$F_1$ is eleventh.
It is second in AUROC ($0.797$; CDFM with native handling has $0.848$) and in $\mathrm{SHD}/d$ ($1.295$; NOTEARS-Linear has $1.276$).
CDFM with native handling has the best AUROC, $F_1$ ($0.462$), and Skel-$F_1$ ($0.614$) at this level; on these three metrics, its averages change least among the amortized methods ($-0.012$, $+0.006$, and $-0.008$), and its $\mathrm{SHD}/d$ falls by $0.100$.
GES did not complete one dataset at this level; on the 13 datasets that it completed, \cdtname has AUROC $0.791$, $F_1$ $0.348$, Skel-$F_1$ $0.486$, and $\mathrm{SHD}/d$ $1.292$, so the order of the two methods on each metric is the same as in Table~\ref{tab:missing-30}.

\cdtname accepts incomplete tables without imputation.
Under purely random missingness, it stays first on all four averages at the $10\%$ level; at the $30\%$ level, it keeps the best AUROC and $\mathrm{SHD}/d$ but not the best $F_1$ (GRaSP, CDFM with native handling, and GES are ahead) or Skel-$F_1$.
Under selection-driven missingness, it degrades more: it is second in $F_1$ at the $10\%$ level and ninth at the $30\%$ level (level with the eighth row at the printed precision), where CDFM with native handling changes least among the amortized methods.
The averages of several classical methods are not monotone in the level: for example, under purely random missingness GRaSP has a higher $F_1$ at the $30\%$ level ($0.458$) than at the $10\%$ level ($0.419$).
\notationchange{Each table shows a single realization of the missing cells, and the two levels use different driver and target sets.}

\section{Latent Confounder Recovery: Protocol and Results}
\label{app:confounder}

This appendix details the comparison of FCI, CDFM, and \cdtname in Section~\ref{sec:confounder-eval}.

A pair of observed variables $i$ and $j$ is confounded, $C^*_{ij}=1$, when both are reached from the same hidden node through hidden nodes only, whether or not $i$ and $j$ are also adjacent; this is the rule by which the training generator labels hidden common causes (Appendix~\ref{sec:training-data}).
A confounded pair in which one variable is an ancestor of the other (e.g., $U\to X$, $U\to Y$, and $X\to Y$ with $U$ hidden) is joined by a directed rather than a bidirected edge in the maximal ancestral graph, so a correct FCI output does not mark it with $i\leftrightarrow j$; CDFM is likewise trained to give a direct edge priority over a bidirected one~\citep{qiao2026cdfm}.
We therefore report the $F_1$ of each setting twice (Table~\ref{tab:confounder-eval}): on all pairs, and on the pairs in which neither variable is an ancestor of the other.
One variable is an ancestor of the other in $493$ of the $1{,}039$ confounded pairs of the in-distribution setting, in $330$ of $2{,}355$ in the semi-synthetic setting, and in $144$ of $698$ in the benchmark setting.

\paragraph{\notationchange{Evaluation Settings.}}
\notationchange{\emph{In-Distribution:}} $40$ tasks from the base generator of Appendix~\ref{sec:training-data} at its highest difficulty ($\zeta=1$), with $10\le D\le 30$ observed variables and $N=500$, without the task-level injections and the post-processing steps of the training pipeline, except for what the base generator applies itself: missing values (probability $0.15$ per task at $\zeta=1$) and, on the diverse path, multiplicative measurement noise and zero-inflation of single variables (probability $0.15$ and $0.08$ per variable); the generator's mechanism drift (probability $0.08$ per diverse-path task) was drawn in none of the $40$ tasks, and $8$ of them contain missing values.
The labels are those of the generator, which removes the confounding label of a pair whose absolute correlation in the generated data is below $0.1$; a pair with a constant column keeps its label.

\notationchange{\emph{Semi-Synthetic:}} five of the semi-synthetic networks of Table~\ref{tab:datasets} (\textsc{alarm\_like}, \textsc{child}, \textsc{child\_nonlinear}, \textsc{ecoli\_like}, and \textsc{insurance}) with $\mathrm{round}(0.15D)$ hub variables hidden.
As in the training generator, the hidden variables are drawn without replacement among the nodes with at least two children, with weights equal to the number of children plus one.

\emph{Benchmark:} each of the 14 benchmark datasets with $\mathrm{round}(0.15D)$ variables hidden, drawn uniformly at random, without replacement, from all variables: from $1$ variable for the two \textsc{asia} tables ($D=8$) to $7$ for \textsc{ecoli\_like} ($D=46$).

In the last two settings, we draw $10$ hidden sets per dataset and use each distinct set once.
This gives $42$ hidden sets in the semi-synthetic setting ($10$ per network, except $6$ for each of \textsc{child} and \textsc{child\_nonlinear}, whose graph has only five nodes with at least two children) and $129$ in the benchmark setting ($10$ per dataset, except $6$ for each of the two \textsc{asia} tables and $9$ for each of the three \textsc{sachs} tables).
The variables of a hidden set are removed from the table.
The labels follow the rule above with the removed variables as the hidden nodes (the latent projection of the dataset's graph); they depend on the graph and the hidden set alone, and no correlation threshold is applied.

\paragraph{\notationchange{Methods and Scoring.}}
Each method is used in its default configuration.
\emph{\cdtname:} the confounding matrix $\hat{\bm C}$, averaged over the $K=10$ inference passes (Appendix~\ref{app:IP}); a pair is flagged when $\hat C_{ij}>0.5$, and pairs are ranked by $\hat C_{ij}$.
\emph{FCI}~\citep{spirtes2000causation}: the causal-learn implementation~\citep{zheng2024causallearn} with the Fisher-$z$ test at $\alpha=0.05$ on every table, including the discrete \textsc{child} and \textsc{insurance}; a pair is flagged when it has a bidirected edge $i\leftrightarrow j$, and pairs are ranked by a graded score: $2$ for $i\leftrightarrow j$, $1$ for another edge without a tail mark ($\circ\!\!\rightarrow$, $\circ\!\!-\!\!\circ$), and $0$ otherwise.
\emph{CDFM}: the package, checkpoint, and row limit of Appendix~\ref{app:baselines}; it has no separate confounding output, so a pair is flagged when its adjacency output contains both $i\to j$ and $j\to i$ at the threshold that the package computes for each table, and pairs are ranked by the smaller of the two directed-edge probabilities.
$F_1$ is computed from the flags and AUPRC from the rankings; AUPRC is the average precision over the unordered pairs, and pairs with equal scores share one threshold.

All methods are scored on the same pairs and labels.
FCI is run on the same variables as \cdtname: a column that the inference protocol of \cdtname sets aside (a constant column, or one with fewer than $\max(10,\,0.10N)$ observed values; Appendix~\ref{app:IP}) is also set aside for FCI, and neither method flags a pair with such a column.
In the in-distribution setting, $32$ of the $1{,}039$ confounded pairs involve a constant column.
CDFM receives every column and handles missing values itself.
The other missing values stay missing for \cdtname and are mean-imputed per variable for FCI.
Pairs are pooled over the $40$ tasks of the in-distribution setting and over the hidden sets of a dataset in the other two settings, whose reported values are averages over the datasets.

\begin{table}[ht]
\centering
\small
\setlength{\tabcolsep}{4pt}
\renewcommand{\arraystretch}{1.12}
\resizebox{\textwidth}{!}{%
\begin{tabular}{@{}lrrcccccc@{}}
\toprule
 & & & \multicolumn{3}{c}{$F_1$ $\uparrow$} & \multicolumn{3}{c}{AUPRC $\uparrow$} \\
\cmidrule(lr){4-6}\cmidrule(lr){7-9}
\textbf{Dataset} & \makecell{Hidden\\sets} & \makecell{Confounded\\pairs} & FCI & CDFM & \cdtname & FCI & CDFM & \cdtname \\
\midrule
% columns: Dataset & Hidden sets & Confounded pairs & F1 of: FCI & CDFM & FoundCause & AUPRC of: FCI & CDFM & FoundCause
\textsc{alarm\_like} & $10$ & $184$ of $4650$ & $\bm{0.268}$ & $0.230$ & $0.147$ & $0.143$ & $\bm{0.193}$ & $0.102$ \\
\textsc{child} & $6$ & $202$ of $816$ & $\bm{0.260}$ & $0.050$ & $0.066$ & $0.291$ & $\bm{0.294}$ & $0.253$ \\
\textsc{child\_nonlinear} & $6$ & $202$ of $816$ & $0.164$ & $0.081$ & $\bm{0.422}$ & $0.259$ & $0.342$ & $\bm{0.351}$ \\
\textsc{ecoli\_like} & $10$ & $1105$ of $7410$ & $\bm{0.096}$ & $0.092$ & $0.033$ & $0.203$ & $\bm{0.318}$ & $0.205$ \\
\textsc{insurance} & $10$ & $662$ of $2530$ & $\bm{0.221}$ & $0.176$ & $0.090$ & $\bm{0.318}$ & $0.315$ & $0.309$ \\
\midrule
Average & -- & -- & $\bm{0.202}$ & $0.126$ & $0.152$ & $0.243$ & $\bm{0.292}$ & $0.244$ \\
% totals: 42 hidden sets, 2355 confounded pairs of 16222 pairs
\bottomrule
\end{tabular}%
}
\caption{\textbf{Latent confounder recovery on the five semi-synthetic networks with $15\%$ hub variables hidden} (all pairs, each method in its default configuration).
Confounded pairs (pairs that share a hidden common cause) and pairs are pooled over the hidden sets of a dataset; in total there are $42$ hidden sets and $2{,}355$ confounded pairs among $16{,}222$ pairs.
The last row averages over the five datasets; its $F_1$ values are the semi-synthetic row of Table~\ref{tab:confounder-eval}.
Bold marks the highest value of the three methods for each metric.}
\label{tab:confounder-semisynth}
\end{table}

\begin{table}[ht]
\centering
\small
\setlength{\tabcolsep}{4pt}
\renewcommand{\arraystretch}{1.12}
\resizebox{\textwidth}{!}{%
\begin{tabular}{@{}lrrcccccc@{}}
\toprule
 & & & \multicolumn{3}{c}{$F_1$ $\uparrow$} & \multicolumn{3}{c}{AUPRC $\uparrow$} \\
\cmidrule(lr){4-6}\cmidrule(lr){7-9}
\textbf{Dataset} & \makecell{Hidden\\sets} & \makecell{Confounded\\pairs} & FCI & CDFM & \cdtname & FCI & CDFM & \cdtname \\
\midrule
% columns: Dataset & Hidden sets & Confounded pairs & F1 of: FCI & CDFM & FoundCause & AUPRC of: FCI & CDFM & FoundCause
\textsc{alarm\_like} & $10$ & $34$ of $4650$ & $0.060$ & $\bm{0.123}$ & $0.036$ & $0.027$ & $\bm{0.067}$ & $0.018$ \\
\textsc{asia} & $6$ & $2$ of $126$ & $0.000$ & $\bm{0.077}$ & $0.040$ & $0.069$ & $0.094$ & $\bm{0.268}$ \\
\textsc{asia\_nonlinear} & $6$ & $2$ of $126$ & $0.000$ & $0.000$ & $\bm{0.043}$ & $0.016$ & $\bm{0.049}$ & $0.029$ \\
\textsc{causal\_chambers\_lt} & $10$ & $91$ of $1360$ & $0.153$ & $0.000$ & $\bm{0.416}$ & $0.116$ & $\bm{0.588}$ & $0.399$ \\
\textsc{child} & $10$ & $51$ of $1360$ & $\bm{0.119}$ & $0.000$ & $0.029$ & $\bm{0.048}$ & $0.043$ & $0.035$ \\
\textsc{child\_nonlinear} & $10$ & $51$ of $1360$ & $0.061$ & $0.036$ & $\bm{0.082}$ & $0.038$ & $\bm{0.086}$ & $0.054$ \\
\textsc{ecoli\_like} & $10$ & $151$ of $7410$ & $0.097$ & $\bm{0.136}$ & $0.087$ & $0.056$ & $\bm{0.114}$ & $0.039$ \\
\textsc{insurance} & $10$ & $116$ of $2530$ & $0.058$ & $\bm{0.065}$ & $0.033$ & $0.048$ & $\bm{0.061}$ & $0.051$ \\
\textsc{petshop\_high\_traffic} & $10$ & $13$ of $5280$ & $\bm{0.028}$ & $0.000$ & $0.000$ & $0.004$ & $0.003$ & $\bm{0.005}$ \\
\textsc{petshop\_low\_traffic} & $10$ & $14$ of $5950$ & $\bm{0.025}$ & $0.011$ & $0.007$ & $\bm{0.004}$ & $0.004$ & $0.003$ \\
\textsc{petshop\_temporal} & $10$ & $8$ of $5950$ & $\bm{0.034}$ & $0.000$ & $0.013$ & $\bm{0.009}$ & $0.004$ & $0.005$ \\
\textsc{sachs\_full} & $9$ & $55$ of $324$ & $\bm{0.205}$ & $0.127$ & $0.000$ & $0.175$ & $\bm{0.224}$ & $0.200$ \\
\textsc{sachs\_nonlinear} & $9$ & $55$ of $324$ & $0.051$ & $0.060$ & $\bm{0.432}$ & $0.220$ & $0.250$ & $\bm{0.333}$ \\
\textsc{sachs\_obs} & $9$ & $55$ of $324$ & $0.000$ & $\bm{0.096}$ & $0.000$ & $0.164$ & $\bm{0.211}$ & $0.173$ \\
\midrule
Average & -- & -- & $0.064$ & $0.052$ & $\bm{0.087}$ & $0.071$ & $\bm{0.128}$ & $0.115$ \\
% totals: 129 hidden sets, 698 confounded pairs of 37074 pairs
% petshop_low_traffic, AUPRC: CDFM (0.00374) prints like the highest value (FCI, 0.00400) but is lower; bold is decided on the unrounded values.
\bottomrule
\end{tabular}%
}
\caption{\textbf{Latent confounder recovery on each benchmark dataset with $15\%$ of the variables hidden} (all pairs, each method in its default configuration).
Confounded pairs (pairs that share a hidden common cause) and pairs are pooled over the hidden sets of a dataset; in total there are $129$ hidden sets and $698$ confounded pairs among $37{,}074$ pairs.
The last row averages over the 14 datasets; its $F_1$ values are the benchmark row of Table~\ref{tab:confounder-eval}.
Bold marks the highest value of the three methods for each metric, determined before rounding (on \textsc{petshop\_low\_traffic}, the AUPRC of CDFM is below that of FCI).}
\label{tab:confounder-per-dataset}
\end{table}

\paragraph{\notationchange{Recovery Results.}}
Table~\ref{tab:confounder-eval} gives the $F_1$ in the three settings; Tables~\ref{tab:confounder-semisynth} and~\ref{tab:confounder-per-dataset} give $F_1$ and AUPRC per dataset on all pairs.
In distribution, \cdtname has the highest $F_1$ ($0.299$; FCI $0.197$, CDFM $0.111$); it also has the highest $F_1$ on the pairs without an ancestor relation ($0.334$).

In the semi-synthetic setting, FCI has the highest average $F_1$ ($0.202$; \cdtname $0.152$, CDFM $0.126$) and the highest $F_1$ on four of the five networks; \cdtname has it on \textsc{child\_nonlinear}, and CDFM on none.

In the benchmark setting, all three methods score low; $698$ of the $37{,}074$ pairs ($1.9\%$) are confounded.
\cdtname has the highest average $F_1$ ($0.087$; FCI $0.064$, CDFM $0.052$), but the highest $F_1$ on only four of the 14 datasets, against five each for FCI and CDFM.
Its average is carried by \textsc{causal\_chambers\_lt} ($0.416$) and \textsc{sachs\_nonlinear} ($0.432$): without these two datasets, it has the lowest average $F_1$ of the three ($0.031$; FCI $0.057$, CDFM $0.056$).

On the pairs without an ancestor relation, FCI keeps the highest average $F_1$ in the semi-synthetic setting ($0.228$) and \cdtname in the benchmark setting ($0.158$), where FCI ($0.075$) and CDFM ($0.077$) are almost level.
There, \cdtname has the highest $F_1$ on five of the 14 datasets, CDFM on six, and FCI on three; without \textsc{causal\_chambers\_lt} and \textsc{sachs\_nonlinear}, CDFM has the highest average $F_1$ ($0.090$; \cdtname $0.083$, FCI $0.066$).

On all pairs, \cdtname has the highest recall in every setting ($0.246$, $0.189$, and $0.338$, in the order of Table~\ref{tab:confounder-eval}; FCI $0.133$, $0.143$, and $0.127$; CDFM $0.070$, $0.096$, and $0.098$), and FCI the highest precision ($0.381$, $0.480$, and $0.065$; \cdtname $0.379$, $0.253$, and $0.058$; CDFM $0.269$, $0.267$, and $0.063$).

By AUPRC on all pairs (averaged over the datasets in the semi-synthetic and benchmark settings), \cdtname is highest in distribution ($0.352$; CDFM $0.199$, FCI $0.180$) and CDFM in the other two settings (semi-synthetic: $0.292$, \cdtname $0.244$, FCI $0.243$; benchmark: $0.128$, \cdtname $0.115$, FCI $0.071$).

\paragraph{\notationchange{Control With Observed Common Causes.}}
We also scored the pairs and labels of the benchmark setting with the outputs of \cdtname and FCI on the complete tables, in which the common cause of every confounded pair is observed.
Both methods flag almost the same share of pairs as with the variables hidden, with almost the same recall and $F_1$: \cdtname flags, on average over the datasets, $19.0\%$ of the pairs on the complete tables and $19.2\%$ with the variables hidden (recall $0.319$ and $0.338$, $F_1$ $0.082$ and $0.087$), and FCI $5.0\%$ and $5.4\%$ (recall $0.116$ and $0.127$, $F_1$ $0.059$ and $0.064$).
On average over the benchmark datasets, the flags of either method therefore do not distinguish a hidden common cause from an observed one: a flag indicates at most that the pair shares a common cause, whether or not that cause is observed.

\end{document}